\documentclass[lettersize,journal]{IEEEtran}
\usepackage{amsmath,amsfonts}
\usepackage{algorithmic}
\usepackage{algorithm}
\usepackage{array}
\usepackage[caption=false,font=normalsize,labelfont=sf,textfont=sf]{subfig}
\usepackage{textcomp}
\usepackage{stfloats}
\usepackage{url}
\usepackage{verbatim}
\usepackage{graphicx}
\usepackage{cite}
\usepackage[export]{adjustbox} 
\usepackage{booktabs,multirow}
\newcommand{\best}[1]{\textbf{#1}}
\newcommand{\second}[1]{\underline{#1}}

\usepackage{microtype} 

\usepackage[font=small,labelfont=bf]{caption}
\usepackage{subcaption}
\usepackage{booktabs}
\usepackage{tabularx}
\newcolumntype{L}{>{\raggedright\arraybackslash}X}
\usepackage{needspace}

\usepackage[hidelinks]{hyperref}  
\usepackage[capitalize,noabbrev]{cleveref} 
\usepackage{listings}
\usepackage{color}
\usepackage{caption}

\definecolor{dkgreen}{rgb}{0,0.6,0}
\definecolor{gray}{rgb}{0.5,0.5,0.5}
\definecolor{mauve}{rgb}{0.58,0,0.82}

\begin{document}

\title{VAE with Hyperspherical Coordinates: Improving Anomaly Detection from Hypervolume-Compressed Latent Space}

\author{1\textsuperscript{st} Alejandro Ascarate
\textit{School of Electrical Engineering and Robotics} \\
\textit{Queensland University of Technology}\\
Brisbane, Australia \\
a.ascaratecastro@hdr.qut.edu.au
\and
2\textsuperscript{nd}  Leo Lebrat
\textit{School of Electrical Engineering and Robotics} \\
\textit{Queensland University of Technology}\\
Brisbane, Australia \\
leo.lebrat@qut.edu.au
\and
3\textsuperscript{rd} Rodrigo Santa Cruz
\textit{Data61} \\
\textit{CSIRO}\\
Brisbane, Australia \\
rodrigo.santacruz@csiro.au
\and
4\textsuperscript{th} Clinton Fookes
\textit{School of Electrical Engineering and Robotics} \\
\textit{Queensland University of Technology}\\
Brisbane, Australia \\
c.fookes@qut.edu.au
\and
5\textsuperscript{th} Olivier Salvado
\textit{School of Electrical Engineering and Robotics} \\
\textit{Queensland University of Technology}\\
Brisbane, Australia \\
olivier.salvado@qut.edu.au
}



\maketitle

\begin{abstract}
    Variational autoencoders (VAE) encode data into lower-dimensional latent vectors before decoding those vectors back to data. Once trained, one can hope to detect \textit{out-of-distribution} (OOD) latent vectors (abnormal), but several issues arise when the latent space is high dimensional. This includes an exponential growth of the hypervolume with the dimension, which severely affects the generative capacity of the VAE. In this paper, we draw insights from high dimensional statistics: in these regimes, the latent vectors of a standard VAE are distributed on the `equators' of a hypersphere, challenging the detection of anomalies. We propose to formulate the latent variables of a VAE using hyperspherical coordinates, which allows compressing the latent vectors towards a given direction on the hypersphere, thereby allowing for a more expressive approximate posterior. We show that this improves both the fully unconditional-OOD and conditional-OOD anomaly detection ability of the VAE, achieving the best performance on the datasets we considered, outperforming existing methods. For the unconditional-OOD and conditional-OOD modalities, respectively, these are: i) detecting unusual landscape from the Mars Rover camera and unusual Galaxies from ground based imagery (complex, real world datasets); ii) standard benchmarks like Cifar10 and subsets of ImageNet as the in-distribution (ID) class. 
\end{abstract}


\section{Introduction}
\label{sec:intro}

\begin{figure*}[!h]
    \centering
    \includegraphics[width=0.89\textwidth]{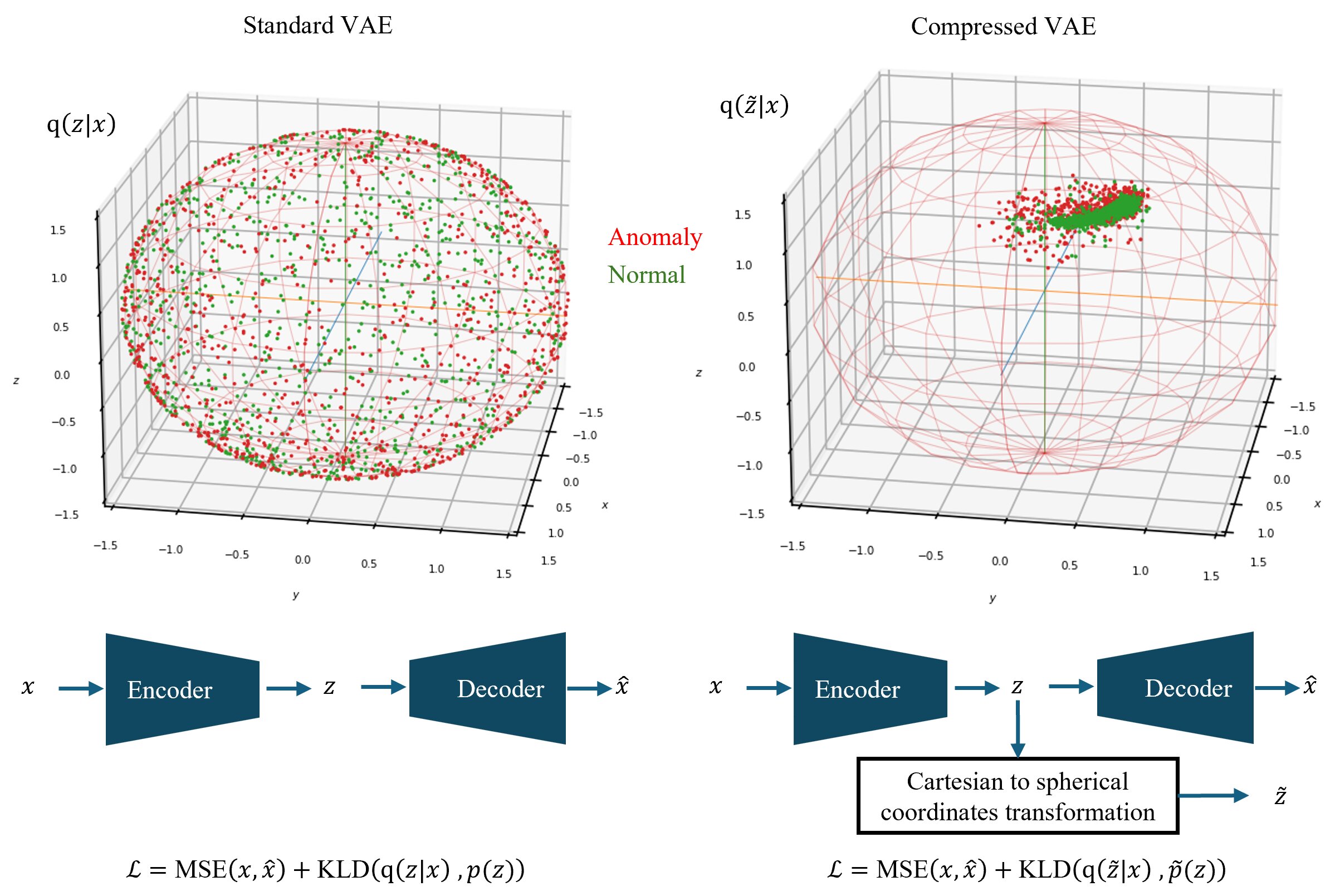}
    \caption{Proposed method for the \textbf{fully unconditional-OOD case}. The standard VAE (left) is modified (right) by converting the latent vectors into hyperspherical coordinates. In our new formulation, the latent vectors from the normal class in green can be moved during training towards a given direction on the hypersphere, forming a dense and compact ``island'', illustrated here by projecting the latent distributions on a 2D sphere (see subsection Results \ref{subsec:results} for more details about how this is done). Anomalies in red are detected by measuring their distance to the island. The figure corresponds to results from the experiment on the Galaxy Zoo dataset (cf. Table \ref{tab:AUC_FPR_table}, third column).}
    \label{fig:Figure-summary}
\end{figure*}

\begin{figure*}[!h]
    \centering
    \includegraphics[width=0.89\textwidth]{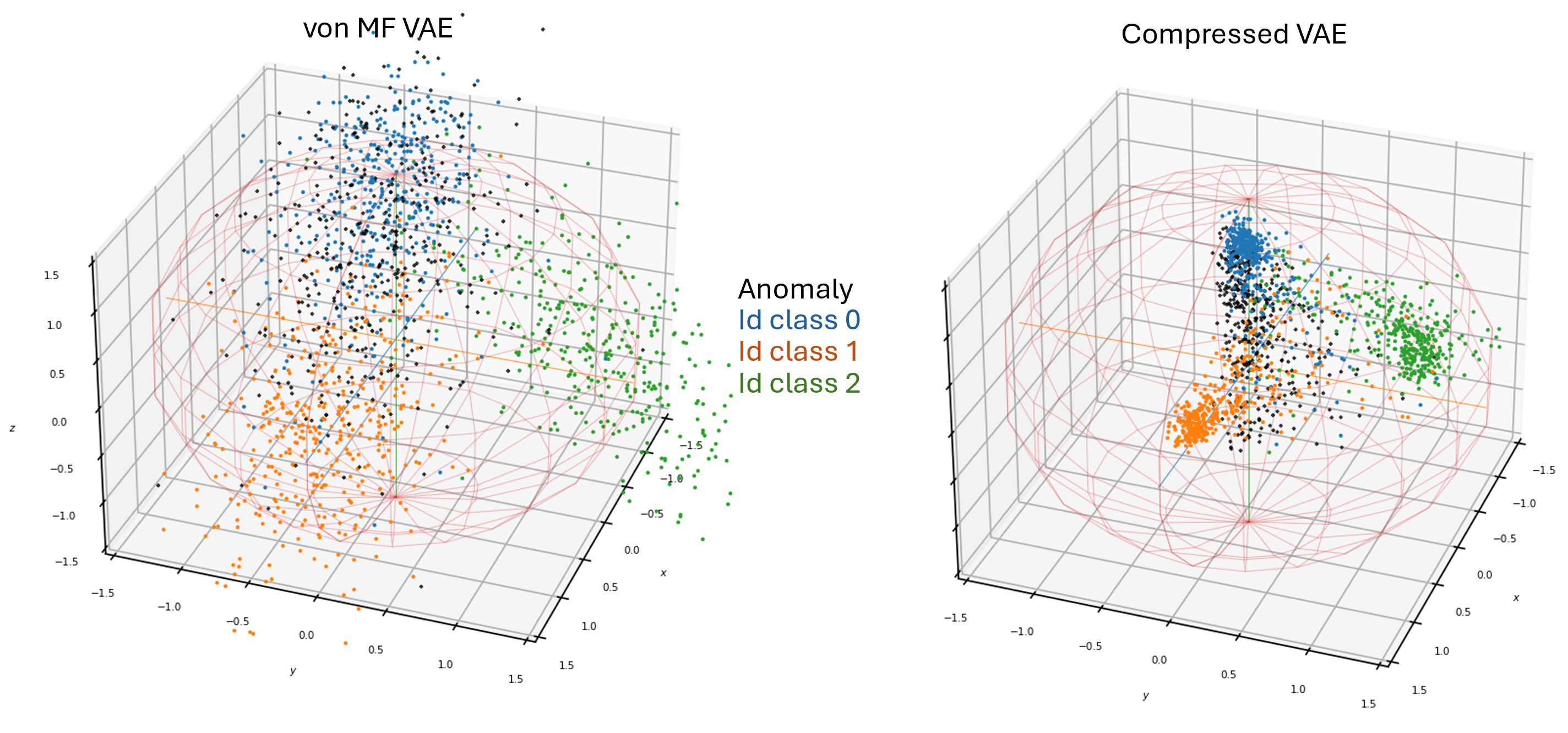}
    \caption{Proposed method for the \textbf{conditional-OOD case}. In this case, a compression of the same type as in the previous figure is done on each of the ID class clusters, simply by re-orienting the full hyperspherical coordinate system such that the first angular coordinate is the angle w.r.t. the Cartesian orthogonal axis whose number is equal to a corresponding ID class label. The von Mises-Fisher-based method (von MF) shows more noisy and dispersed samples because having only one single parameter (the first hyperspherical angle) to compress the volume and thus reduce the sparsity of the HD space is not enough, as we show in the Appendix (App.~\ref{conditional-OOD experiments}--\ref{Imagenette ID Training}), where we also see that t-SNE can be misleading for assessing compression. In contrast, our method compresses all of the hyperspherical angles. The figure corresponds to results from the experiment on Imagenette vs close Imagenet (cf. Table \ref{tab:nearood-imagenette}).}
    \label{fig:Figure-summary_2}
\end{figure*}

Anomaly detection (AD) can be done in a fully unconditional-OOD or conditional-OOD manner. Fully unconditional-OOD anomaly detection, where one \textit{does not have access to sub-class information/labels in the normal/ID data}, is a challenging task, and two main approaches have shown promise. They both rely on an autoencoder (AE), which encodes the data into a lower dimension latent space, before decoding the latent vectors back to data. The main assumption is that the AE, having learned to encode/decode the training dataset, would do poorly in processing a sample outside that distribution.

The detection can either be done by comparing the reconstructed and the original data \cite{ThomasG.Dietterich2021AnomalyDetection,Kerner2020ComparisonMissions}, or by detecting whether latent vectors lie outside the ``normal'' \textit{latent} distribution. A variation of the former uses a diffusion-based generative model to reconstruct data, but the detection is also between the generated and original data \cite{Liu2025ADetection}. 

Many recent works have focused on the conditional-OOD AD case (and often call it just ``OOD''), where one uses sub-class information from the normal training set to help the task, for example by disentangling the normal subclasses with a classifier, e.g., a ResNet \cite{Wang2023LearningDetection,Li2025Out-of-DistributionQuestions}; the ‘good’ embeddings/features from the penultimate layer can then be used for AD via some standard method outlier detection techniques, e.g., k-NN \cite{Sun2022Out-of-DistributionNeighbors}. This greatly simplifies the problem. 

In this work, we explored the performance of our method on both of these two types of AD scenarios.

Our main contribution is a novel way to detect anomalies in latent space. AD is challenging due to the high dimensionality of practical datasets such as the images used in this paper: because of concentration of measure effects, when encoding samples uniformly to a latent hypersphere, they tend to \textit{only} populate its `\textit{equators}'. We proposed to convert the latent vectors from Cartesian coordinates to hyperspherical coordinates, de facto disentangling the dimensions: a point on a hypersphere can be moved on the surface by changing only one hyperspherical angle. This is impossible to do using standard Cartesian coordinates: moving a point on the hypersphere surface involves modifying every Cartesian latent dimension. Because of that disentanglement, we could reformulate the VAE cost function to move all the latent samples towards a given direction on the hypersphere, creating a very dense island, away from the high hypervolume equators. Measuring how far a sample is from that island of normal data becomes easier than in the case of a uniform distribution on the hypersphere with its vast equators, and a simple $k-$nearest neighbors achieves better results for AD than existing methods. The creation of a dense island by this particular method has already been shown to be useful when using the VAE for the generative task \cite{Ascarate2025ImprovingCoordinates}, and here we show its value for AD.

Reviews on the standard VAE can be found in \cite{DiederikPKingma2014Auto-EncodingBayes, Kingma2019AnAutoencoders}. The latent space tends towards a high dimensional independent multivariate Gaussian, which has properties that we briefly review next.

\subsection{High Dimensional Spaces}
\label{subsec:HDS}

Sampling from a multivariate Gaussian in a high-dimensional (HD) Euclidean space of dimension \(n\) exhibits several counterintuitive properties. Although the origin has the highest probability density, the probability of drawing samples near it is nearly zero. Instead, most samples concentrate near the \((n-1)\)-dimensional hypersphere \(\mathbb{S}_{\sqrt{n}}^{n-1}\) of radius \(\sqrt{n}\). The norm of the samples follows a \(\chi(n)\) distribution, which implies that samples lie within a thin shell around the hypersphere. The thickness of this shell relative to its radius \(\sqrt{n}\) shrinks as \(n\) increases.

As \(n\) grows large, the distribution of \(\mathcal{N}(0, I_n)\) approaches the uniform distribution on the hypersphere. Furthermore, any two independent samples from \(\mathcal{N}(0, I_n)\) are always nearly orthogonal to one another, a property called \textit{almost-orthogonality} (see~\cite{Vershynin2018High-DimensionalProbability} for a formal treatment).

These behaviors are deeply tied to how (hyper-)volume behaves in HD spaces. Under the uniform measure on a hypersphere, most of its exponentially growing volume in \(n\) is concentrated in extremely thin \textbf{equatorial} bands relative to \textbf{any randomly chosen “north pole.”} This is a truly remarkable and deep fact. It is formalized in~\cite{Wainwright2019ConcentrationMeasure}, and is part of the broader notion known in mathematics as \textit{concentration of measure}. Standard low-dimensional (2D or 3D) intuitions about spheres break down in HD spaces, and such properties significantly affect anomaly detection in models like VAEs, as we discuss next. We give some simple yet insightful examples in App.~\ref{appendix:measurecon}, as well as some remarks on its connection with the volume, which will be key in our paper.

\subsection{Anomaly Detection in High Dimension}
\label{subsec:HDA}

High resolution/complex images need many latent dimensions to capture all the information they convey. But, as mentioned before, HD spaces often display properties that go against the intuition gained from their low dimensional counterparts, where many of the original methods for AD were developed. 

For example, a common assumption is that anomalies will be located in the tail of the normal data distribution. Then, for an AD method to have good performance, one would need this tail to allow for some concentration of samples, that is, a heavy tail. In HD spaces, the tails of a wide class of functions of distributions, like the norm of a sample from a standard Gaussian, tend to be very short. This is the classic concentration of measure phenomenon. More formally:

\smallskip

\noindent\textbf{Proposition} (Measure Concentration \cite{Vershynin2018High-DimensionalProbability,Wainwright2019ConcentrationMeasure,ChrisAkers2022TheComplexity}): Let $z$ be a Gaussian random vector and $f:\mathbb{R}^{n}\,\longrightarrow\mathbb{R}$ a Lipschitz function with Lipschitz constant $K$. Then, 
\[\mathrm{Pr}\left(\mid f(z)-\mathbb{E}f(z)\mid\geq t\right)\leq2\,\mathrm{exp}\left(-\frac{t^{2}}{4K}\right).\,\square\]

Note that the previous statement \textit{does not} depend on the dimension $n$ (see also \cite{ChrisAkers2022TheComplexity}, Appendix B, for a compact introduction to the general Riemannian case). The effects of HD can be seen when selecting a particular function: e.g., for the previously alluded concentration of the norm of the Gaussian around $\sqrt{n}$, this follows from the general result applied to $f(z)=\parallel z\parallel$, since $\mathbb{E}\parallel z\parallel\sim\sqrt{n}$ and $K=1$.

This can affect the standard VAE both as a generative model \cite{Ascarate2025ImprovingCoordinates} as well as a tool for AD \cite{Tam2025OnModels}, since it assumes a standard Gaussian distribution as prior. \textit{Crucially, the concentration effects will affect the anomaly score itself}, since it is a function from the Gaussian-like latent encodings to the reals, and thus produce short tails in its distribution, which makes the disentangling between the normal and abnormal classes more difficult (see Fig.~\ref{fig:30b} in the Appendix for examples from our experiments).




Our main hypothesis is that a model with a HD latent distribution/representation resembling a HD Gaussian will be very negatively affected by the concentration of measure phenomena for tasks like AD.

\section{Method}
\label{sec:method}

\subsection{VAE with Hyperspherical Coordinates}

Our approach is based on formulating the initial KL divergence term with a prior from the original VAE, which is in Cartesian coordinates, to one in hyperspherical coordinates. See App.~\ref{appendix:hstransform} for the standard conversion formulas between Cartesian and hyperspherical in high dimension. 

In Cartesian coordinates, the KL divergence between the estimated posterior defined by $\mu_k$ and $\sigma_k$, and the prior defined by $\mu_k^p$ and $\sigma_k^p$ has been well documented \cite{DiederikPKingma2014Auto-EncodingBayes}. It can be written as (see App.~\ref{appendix:KLD}), 

\begin{equation}
\begin{aligned}
\mathrm{KLD}_{\mathrm{CartCoords}}^{w/\mathrm{Prior}}
&\simeq \sum_{k=1}^{n} \Big[
\big(\mathbb{E}_{b}[\sigma_{k}] - \sigma_{k}^{p}\big)^{2}
+ \sigma_{b}[\sigma_{k}]^{2} \\
&\hspace{2.6em}+
\big(\mathbb{E}_{b}[\mu_{k}] - \mu_{k}^{p}\big)^{2}
+ \sigma_{b}[\mu_{k}]^{2}
\Big],
\end{aligned}
\end{equation}
where \(\mathbb{E}_b\) and \(\sigma_b\) denote the mini batch statistics of size \(N_b\).

This formulation using Cartesian coordinates includes batch statistics and was partly inspired by the construction in \cite{Bardes2021VICReg:Learning}. It will be useful for our next step. 

We now introduce hyperspherical coordinates in the KLD formulation. We start with the Cartesian coordinates \((\mu_{i},\sigma_{i})\), given by the encoder, and transform these to their hyperspherical counterparts \((\overset{\mu}{r},\overset{\mu}{\varphi_{k}};\overset{\sigma}{r},\overset{\sigma}{\varphi_{k}})\) with $r$ a scalar and $k$ the index of the $n-1$ spherical angles. 

The KLD-like objective becomes for the angles $\varphi_k$,

\begin{align}
\mathrm{KLD}_{\mathrm{HSphCoords}}^{w/\mathrm{Prior}}(\varphi)
&=
\sum_{k=1}^{n-1}\Big[
\alpha_{\sigma,k}\big(\mathbb{E}_{b}[\cos \overset{\sigma}{\varphi_k}] - a_{\sigma,k}\big)^{2}
\nonumber\\
&\quad+
\beta_{\sigma,k}\big(\sigma_{b}[\cos \overset{\sigma}{\varphi_k}] - b_{\sigma,k}\big)^{2}
\nonumber\\
&\quad+
\alpha_{\mu,k}\big(\mathbb{E}_{b}[\cos \overset{\mu}{\varphi_k}] - a_{\mu,k}\big)^{2}
\nonumber\\
&\quad+
\beta_{\mu,k}\big(\sigma_{b}[\cos \overset{\mu}{\varphi_k}] - b_{\mu,k}\big)^{2}
\Big],
\end{align}
and for the norm $r$,

\begin{align}
\mathrm{KLD}_{\mathrm{HSphCoords}}^{w/\mathrm{Prior}}(r)
&=
\alpha_{\sigma,r}\big(\mathbb{E}_{b}[\overset{\sigma}{r}]-a_{\sigma,r}\big)^{2}
\nonumber\\
&\quad+
\beta_{\sigma,r}\big(\sigma_{b}[\overset{\sigma}{r}]-b_{\sigma,r}\big)^{2}
\nonumber\\
&\quad+
\alpha_{\mu,r}\big(\mathbb{E}_{b}[\overset{\mu}{r}]-a_{\mu,r}\big)^{2}
\nonumber\\
&\quad+
\beta_{\mu,r}\big(\sigma_{b}[\overset{\mu}{r}]-b_{\mu,r}\big)^{2},
\end{align}
with the priors for the mean over the batch $a_{i,j}$, the standard deviation over the batch $b_{i,j}$, and the gains for each term $\alpha_{i,j},\,\beta_{i,j}$, for $i\in\{\sigma,\mu\}$ and $j\in\{1,...,n-1,r\}$.

We use the cosines rather than the angles to avoid costly extra computations of the corresponding arccosines. The coordinate transformation is done using a vectorized implementation (code provided in App.~\ref{appendix:hstransformcode}). The reparameterization trick is still done in the Cartesian coordinates representation. The total loss is,

\begin{equation}
\label{eq:HSloss}
\begin{aligned}
\mathcal{L}
&= \mathrm{MSE}(x,\hat{x}_z)
\\
&\quad + \beta\,\mathrm{KLD}_{\mathrm{HSphCoords}}^{w/\mathrm{Prior}}(\varphi_k)
\\
&\quad + \beta\,\mathrm{KLD}_{\mathrm{HSphCoords}}^{w/\mathrm{Prior}}(r).
\end{aligned}
\end{equation}

\subsection{Volume Compression of the Latent Manifold}

We discussed previously that the standard VAE forces the latent samples to be uniformly distributed on the hypersphere, which in high dimensions results in data located within equators of the hypersphere where the volume is the greatest. A benefit of using hyperspherical coordinates is the ability to set a prior for the $\varphi_k$ that forces the latent samples away from the equators of \textit{each} and \textit{all} $(n-k)$-hyperspheres, $\forall k$, contained (or equal, if $k=1$) in the initial one (since they are HD too), thereby escaping these regions. This can be done for each angular coordinate, as all are uncorrelated with each other, by simply setting (north pole),

\begin{equation}
\boxed{a_{\mu,k}=1,\,\forall k}.
\end{equation}

By doing so, the samples can be moved to a zone with much reduced volume. We speculate that this allows for a more expressive approximate posterior, better suited for the AD task than the one obtained from a standard Gaussian as prior. See App.~\ref{appendix:volumeelement} for a more detailed analysis of the behavior of the hypervolume in this situation. Finally, by setting, 

\begin{equation}
a_{\mu,r}=\sqrt{n},
\end{equation}

(and normalizing \(z\), after sampling via the reparameterization trick, to the same radius $\sqrt{n}$) we can force the latent samples to be on the hyperspherical surface of that radius.

For the fully unconditional-OOD case we compress all of the normal samples into a same cluster. For the conditional-OOD case, we re-orient the full hyperspherical coordinate system such that the first angular coordinate is the angle w.r.t. the Cartesian orthogonal axis whose number is equal to a corresponding ID class label; thus, we get a compact cluster for each normal class close to the intersection between the hypersphere and the corresponding labeled axis. This re-orientation is done by applying the roll operation from PyTorch on each latent representation, where the amount of shifting is equal to the class label of the sample (in our implementation we spaced each of the labeled dimensions by 10 unlabeled ones for having a more convenient way to see the stacked histograms in App.~\ref{conditional-OOD experiments}--\ref{Imagenette ID Training}).

\subsection{von Mises-Fisher-based methods are a subset of our general approach} The case when one reduces \textit{only} a single angular coordinate, e.g., $\varphi_1$, corresponds exactly to varying the \textit{single scalar variance parameter} on a von Mises-Fisher distribution on the hypersphere, which is defined as an isotropic (hence the single free parameter rather than a vector or matrix for this variance) Gaussian whose domain is restricted to the hypersphere. This approach has been used recently in the literature when dealing with encodings into the hypersphere for the conditional-OOD modality \cite{Ming2023HowDetection,SuvraGhosal2024HowDetection}, but as we discuss in more detail in App.~\ref{Comparison with von Mises-Fisher-based approaches}, it cannot reduce the hypervolume as fast as our method in case one wishes to do so.; hence, it will always have more internal dispersion compared to our approach.

\subsection{Anomaly Detection}

AD methods can use a distance function \(d:Z\times Z\,\longrightarrow \mathbb{R}^{+}\) on the space of data points $z \in Z$ as the primary tool to define and compute the anomaly score. This is usually a high-dimensional Euclidean space or a hyper-surface; the distance function can be either the Euclidean distance or the corresponding induced distance, respectively.

The simplest of such methods is the \(k-\)NN score, where the anomaly value of a query data point \(z\) is defined as the mean distance to its \(k-\)nearest neighbors, that is,
\[A(z)=\frac{1}{k}\sum_{i=1}^{k}d(z,z_{i}),\]
where the index \(i\) refers to an ordering of the points in \(Z\) such that \(d(z,z_{1})\leq...\leq d(z,z_{n})\).

In our case, after training the VAE with only the nominal data $X$, we encode this data into the latent space $Z$ and consider the corresponding means $\mu_x$, $x\in X$, given by the encoder. During test time, given a query point $x_{test}$, we encode it to obtain the mean $\mu_{x_{test}}$ and compute its anomaly score $A(\mu_{x_{test}})$ via the previous \(k-\)NN score w.r.t. all the mentioned means $\mu_x$ of the training set.

We use the standard Euclidean distance and set $k=3$.

\section{Related Work}

The Isolation Forest \cite{Liu2008IsolationForest,Emmott2013SystematicData} (iForest or IF) method creates a forest of random axis-parallel projection trees. It derives a score based on the observation that points becoming isolated and closer to the root of a tree are easier to separate from the rest of the data and therefore, are more likely to be anomalous.

When working with VAEs, there is a natural method that suggests itself for AD. This relies on the hypothesis that the reconstruction, by the trained network, of data close to the training set should be reasonably good, since it was optimized for that task, while the reconstruction for anomalies should be worse, since these data points are intrinsically different than those in the training set, the network should have more problems reconstructing them \cite{Pang2022DeepDetection,ThomasG.Dietterich2021AnomalyDetection}. A common anomaly score is thus the reconstruction error (MSE): \(A(x)=\parallel x-\hat{x}\parallel.\)

It is hard to prevent a plain autoencoder from learning to be a general image compression algorithm, though. When this occurs, it is not helpful for anomaly detection via reconstruction error, because it does not fail on new images \cite{Bouman2025AutoencoderUnreliable,Gong2019MemAE}. 

Regarding the novel method for AD that we present in the next section, a recent work \cite{Fu2024DenseDetection} shares some similarities. The main differences are: i) volume compression is done radially between all points, which, from our view, misses the key point of the peculiar angular-like, equatorial distribution of volume in the HD regime; ii) the method is specifically designed for AD only, with only an encoder network for feature extraction, while our method uses a VAE, a generative model, which then can be used for other tasks, e.g., for improving generation, as in \cite{Ascarate2025ImprovingCoordinates}; iii) experiments used only simulated anomalies, done with classification datasets like CIFAR10 in which a class is the anomaly to the other nine. We found this latter aspect problematic, since these fictional AD scenarios do not seem adequate for evaluation of fully unconditional-OOD AD methods, as we will argue in more detail in App.~\ref{appendix:cifar10}. The method in this reference, like ours, uses a $k-$NN approach \cite{ThomasG.Dietterich2021AnomalyDetection} for performing AD. Unfortunately, the code is not provided. 

In the conditional-OOD realm, there are a series of recent methods \cite{Sun2022Out-of-DistributionNeighbors, Ming2023HowDetection,SuvraGhosal2024HowDetection} based on performing AD on the feature space of a deep classifier, e.g., a ResNet, via the $k-$NN approach. The work \cite{Ming2023HowDetection}, in particular, uses hyperspherical embeddings via clusters modeled by von Mises-Fisher (vMF) distributions and $k-$NN on that set-up for AD; thus, it is, conceptually, the closest to our approach. We did not found references using the vMF approach for the fully unconditional-OOD case.

For the case of our ImageNet-based set-up (Imagenette vs. close Imagenet), we implemented the idea in \cite{Sun2022Out-of-DistributionNeighbors} from scratch, with the same ResNet as our encoder in the AE and VAE, and is reported as KNN* in Tables \ref{tab:cifar10-ood}, \ref{tab:nearood-cifar10} left, \ref{tab:nearood-imagenette} right (as emphasized in those references, we normalize the test set before performing the KNN). Furthermore, our Comp.VAE method, when restricted to compress only the first hyperspherical angular coordinate (see Method section), is identical to a von Mises-Fisher method. Thus, this provides a straightforward comparison with that idea, since everything else remains identical in the setup w.r.t. the full Comp.VAE. 

\section{Experimental Results}

\subsection{Model and Implementation}

For all our experiments, we use a small, and customized for serving in a VAE, standard ResNet-18-like architecture \cite{He2015DeepRecognition} for both encoder and decoder, for a total of around 0.1 to 1 million parameters. When using the loss in hyperspherical coordinates (\ref{eq:HSloss}), we use an annealing-like schedule \cite{Fu2019CyclicalVanishing} for the gain $\beta$ of the KLD-like loss, which simply increases proportionally with \(\sqrt{\text{epoch}}\) for a total of \(100\) epochs. See App.~\ref{Model Details and Implementation} for more details.

\subsection{Datasets}

\paragraph{Fully unconditional-OOD case}

The Mars Rover Mastcam dataset \cite{Kerner2020ComparisonMissions} for AD comprises multispectral images from the rover-based planetary exploration missions on the planet Mars. The training, all normal, consists in $9124$ images of size $64\times64$, and $6$ channels (i.e., multispectral imaging). See App.~\ref{Mars Rover Mastcam} for more details. 

The Galaxy Zoo dataset \cite{Lintott2008GalaxySurvey,Lintott2011GalaxyGalaxies} for AD covers $61578$ galaxies, each represented by a $400\times400$ sized image with $3$ channels. The galaxies were classified by volunteers using a
series of questions. One of the questions (corresponding to
Class 6.1) ‘‘Is there anything ‘‘odd’’ about the galaxy?’’, can be
used as a ground truth for anomalies. Following \cite{Lochner2021ASTRONOMALY:Data}, we extracted all objects with a Class 6.1
score greater than 0.9, which means at least 90 per cent of the volunteers
labeled the galaxy as odd. This results in $924$ anomalies. Then we randomly selected $924$ images from the remaining ones to build the normal part of the test set. Thus, we get a training set of $59730$ normal images and a test set of $1848$, half of them normal and the other half abnormal. We resized the dataset to $64\times64$ images in order to make the training less expensive to run. See App.~\ref{Galaxy Zoo} for more details.

The popular MVTec \cite{Bergmann2021MVTecAnomalyDetection} dataset for AD is unfortunately too small in our view (less than 5000 training samples), and therefore we did not consider it in our study.

\paragraph{Conditional-OOD case} We follow common practice and use \textbf{CIFAR-10} (10 classes) as in-distribution (ID) dataset \cite{krizhevsky2009learning}. 
For far out-of-distribution (far-conditional-OOD) evaluation we use six widely adopted datasets, all resized to $32\times 32$:
\textbf{Textures} \cite{cimpoi2014describing}, 
\textbf{SVHN} \cite{netzer2011reading}, 
\textbf{LSUN-Crop} and \textbf{LSUN-Resize} \cite{yu2015lsun}, 
\textbf{iSUN} \cite{xu2015turkergaze}, 
and \textbf{Places365} \cite{zhou2017places}.

Following standard protocol, we also treat \textbf{CIFAR-10} as ID and \textbf{CIFAR-100} as near-conditional-OOD (semantically related) to assess detectors under tighter distributional shifts. Near-conditional-OOD is challenging because samples can lie close to the ID support and be mistaken as ID. 

Finally, we made an even more challenging near-conditional-OOD experiment by taking \textbf{Imagenette} \cite{imagenette_fastai} (a subset of ten classes from \textbf{ImageNet} \cite{ILSVRC15}) as ID and, for each class in it, we selected from ImageNet a corresponding near-conditional-OOD class, i.e., semantically close to it (in App.~\ref{ImageNet-based dataset} we detail which specific classes from ImageNet we selected and why/how).

\subsection{Results}
\label{subsec:results}

We report (i) \textbf{FPR95}: false positive rate of the anomalous samples at 95\% true positive rate on ID; (ii) \textbf{AUROC}: area under the ROC curve. For all tables: best in \best{bold}, second best \second{underlined}.

\paragraph{Fully unconditional-OOD case}

The basic and standard fully unconditional-OOD `pure machine learning' baselines we used to compare and ablate our method are IF and $k-$NN on the raw pixel-space data. And then both of these methods again but now on the latent space of a standard AE and VAE; in addition, we also run a MSE method in the latter cases. We tried other standard methods too, but decided to limit the presentation only to these, since they were always the best performing and more consistent.

See Table \ref{tab:AUC_FPR_table} for the results. $k$-NN (pixel-space) outperforms AE+MSE; AE+$k$-NN (latent; see App.~\ref{Standard VAE on the Mars Rover Mastcam dataset}--\ref{Standard VAE on the Galaxy Zoo dataset}) barely improves on this, while our Comp.VAE+$k$-NN (latent; see App.~\ref{Comp.VAE on the Mars Rover Mastcam dataset}--\ref{Comp.VAE on the Galaxy Zoo dataset}) yields a clear gain, also over the vMF version (see App.~\ref{Comp.VAE (vMF) on the Galaxy Zoo dataset}). 

\newcommand{\NA}{\textemdash}
\begin{table*}[h]
  \centering
  \caption{AUROC (↑) and FPR95 (↓) for anomaly detection methods on two datasets (all experiments run by us). Best in \best{bold}, second best \second{underlined}.}
  \label{tab:AUC_FPR_table}
  \resizebox{0.8\linewidth}{!}{
  \begin{tabular}{lcccc}
    \toprule
    \textbf{AD Method} &
      \multicolumn{2}{c}{\textbf{Mars Rover Mastcam}} &
      \multicolumn{2}{c}{\textbf{Galaxy Zoo}} \\
    \cmidrule(lr){2-3}\cmidrule(lr){4-5}
     & AUROC $\uparrow$ & FPR95 $\downarrow$
     & AUROC $\uparrow$ & FPR95 $\downarrow$ \\
    \midrule
    $k$NN (pixel space)                      & 0.669                & \second{0.63} & 0.740                & 0.80 \\
    Isolation Forest (pixel space)           & 0.541                & 0.97 & 0.661                & 0.93 \\
    AE + $k$NN (latent)                      & 0.681       & 0.64 & 0.754       & \textbf{0.72} \\
    AE + IF (latent)                         & 0.591                & 0.93 & 0.712                & 0.82 \\
    AE + MSE                                 & 0.617                & 0.90 & 0.709                & 0.81 \\
    VAE + $k$NN (latent)                     & 0.664                & 0.66 & 0.741                & \second{0.77} \\
    VAE + IF (latent)                        & 0.530                & 0.94 & 0.700                & 0.84 \\
    VAE + MSE                                & 0.653                & 0.88 & 0.730                & 0.80 \\
    Comp.VAE (vMF) + $k$NN (latent)                              & \second{0.712}                & \second{0.63} & \second{0.773}                & 0.82 \\
    \midrule
    \textit{Comp.VAE} + $k$NN (latent) (ours)& \textbf{0.764}      & \textbf{0.62} & \textbf{0.789}      & 0.80 \\
    \bottomrule
  \end{tabular}}
\end{table*}


\paragraph{Conditional-OOD case} We can see in Tables \ref{tab:cifar10-ood}, \ref{tab:nearood-cifar10}, \ref{tab:nearood-imagenette} that our approach offers a systematic and consistent lowering of the FPR95, despite not achieving some of the best AUCs in the far conditional-OOD case, in both far and near conditional-OOD types. In the case of CIFAR-10 (ID) vs CIFAR-100 (see App.~\ref{conditional-OOD experiments}) the gain is considerable and the AUC is comparable to the state-of-the-art results, while in the far conditional-OOD case it also is noticeable (in the cases without contrastive learning, while the methods using contrastive learning beat our result, \cite{Ming2023HowDetection}, but not by as much as w.r.t. the other methods that do not use contrastive learning).

In Table \ref{tab:nearood-imagenette}, the most challenging experiment for near conditional-OOD, full compression of all the hyperspherical coordinates beats the case of compressing only the first one (vMF method) in both FPR95 and AUC (see App.~\ref{Imagenette ID Training}). This is to be expected if our hypothesis is correct, since the compression of more hyperspherical coordinates helps to reduce the sparsity and volume of the HD latent space even faster, as we discuss in more detail in App.~\ref{appendix:volumeelement}--\ref{Comparison with von Mises-Fisher-based approaches}. Thus, the superior results of the similar, vMF-based method CIDER, \cite{Ming2023HowDetection}, w.r.t. to ours in the far conditional-OOD case seems to be mostly caused by the enhancement of the method by the use of the contrastive learning techniques. We were not able to reproduce the reported results with our KNN* implementation. This is likely due to the fact that the mentioned references use other enhancement techniques (besides the contrastive learning) or architectural details. In any case, the backbone we use was the same for our Comp.VAE, which, thus, manages to get close to the state-of-the-art AUC results and give a state-of-the-art FPR95, even after being built on top of such a low-performing baseline (KNN*).

\begin{table*}[t]
\centering
\scriptsize
\setlength{\tabcolsep}{1.5pt} 
\resizebox{\linewidth}{!}{%
\begin{tabular}{
  l
  @{\hspace{10pt}} rr
  @{\hspace{14pt}} rr
  @{\hspace{14pt}} rr
  @{\hspace{14pt}} rr
  @{\hspace{14pt}} rr
  @{\hspace{14pt}} rr
}
\toprule
& \multicolumn{2}{c}{} & \multicolumn{8}{c}{conditional-OOD Dataset} & \multicolumn{2}{c}{Average} \\
\cmidrule(lr){2-3}
\cmidrule(lr){4-5}\cmidrule(lr){6-7}\cmidrule(lr){8-9}\cmidrule(lr){10-11}
\cmidrule(lr){12-13}
\multirow{-2}{*}{Method}
& \multicolumn{2}{c}{SVHN} & \multicolumn{2}{c}{Places365} & \multicolumn{2}{c}{LSUN} & \multicolumn{2}{c}{iSUN} & \multicolumn{2}{c}{Texture} & \multicolumn{2}{c}{} \\
& FPR$\downarrow$ & AUROC$\uparrow$ & FPR$\downarrow$ & AUROC$\uparrow$ & FPR$\downarrow$ & AUROC$\uparrow$ & FPR$\downarrow$ & AUROC$\uparrow$ & FPR$\downarrow$ & AUROC$\uparrow$ & FPR$\downarrow$ & AUROC$\uparrow$ \\
\midrule
\multicolumn{13}{l}{} \\
MSP          & 59.66 & 91.25 & 62.46 & 88.64 & 45.21 & 93.80 & 54.57 & 92.12 & 66.45 & 88.50 & 57.67 & 90.86 \\
Energy       & 54.41 & 91.22 & 42.77 & 91.02 & 10.19 & 98.05 & 27.52 & 95.59 & 55.23 & 89.37 & 38.02 & \second{93.05} \\
ODIN         & 53.78 & 91.30 & 43.40 & 90.98 & 10.93 & 97.93 & 28.44 & 95.51 & 55.59 & 89.47 & 38.43 & 93.04 \\
GODIN        & 18.72 & 96.10 & 55.25 & 85.50 & 11.52 & 97.12 & 30.02 & 94.02 & 33.58 & 92.20 & 29.82 & 92.97 \\
Mahalanobis  &  9.24 & 97.80 & 83.50 & 69.56 & 67.73 & 73.61 &  6.02 & 98.63 & 23.21 & 92.91 & 37.94 & 86.50 \\
KNN          & 27.97 & 95.48 & 18.50 & 96.84 & 24.68 & 95.52 & 26.74 & 94.96 & 47.84 & 89.93 & \second{29.15} & \textbf{94.55} \\
KNN*          & 91.0 & 66.4 & 90.0 & 59.6 & 90.0 & 59.9 & 91.6 & 57.9 & 88.9 & 55.7 & 90.3 & 59.9 \\
Comp.VAE (vMF)          & 51.8 & 81.3 & 52.6 & 79.7 & 46.8 & 81.5 & 50.2 & 80.6 & 72.3 & 76.1 & 54.7 & 79.8\\
\midrule
Comp.VAE (\textit{ours})    &  15.6 & 93.6 & 20.4 & 91.3 & 15.6 & 93.9 & 17.1 & 93.1 & 32.2 & 86.9 & \textbf{20.2} & 91.8 \\
\bottomrule
\end{tabular}%
}
\caption{Results on CIFAR-10 as ID dataset for Far-conditional-OOD (results for the other methods taken from \cite{Ming2023HowDetection}.}
\label{tab:cifar10-ood}
\end{table*}

\begin{table*}[t]
  \centering
  \scriptsize
  \caption{Near-conditional-OOD results. Left: CIFAR-10 (ID) vs CIFAR-100 (results for the other methods taken from \cite{SuvraGhosal2024HowDetection}. Right: Imagenette (ID) vs close ImageNet classes (all run by us).}
  \label{tab:nearood_pair}

  \begin{minipage}[t]{0.35\textwidth}
    \centering
    \label{tab:nearood-cifar10}
    \setlength{\tabcolsep}{6pt}
    \resizebox{\linewidth}{!}{%
    \begin{tabular}{l rr}
      \toprule
      \textbf{Methods} & \multicolumn{2}{c}{\textbf{Near-conditional-OOD}} \\
      \cmidrule(lr){2-3}
      & FPR95$\downarrow$ & AUROC$\uparrow$ \\
      \midrule
      MSP         & 64.66 & 85.28 \\
      ODIN        & 52.32 & 88.90 \\
      GODIN       & 60.69 & 82.37 \\
      Energy score& 58.66 & 86.06 \\
      ReAct       & 53.51 & 88.96 \\
      GradNorm    & 65.44 & 79.31 \\
      LogitNorm   & 55.08 & 88.03 \\
      DICE        & 58.60 & 87.11 \\
      Mahalanobis & 87.71 & 78.93 \\
      KNN         & 58.34 & 87.90 \\
      SNN         & \second{50.10} & \textbf{89.80} \\
      KNN*         & 90.0 & 61.5 \\
      Comp.VAE (vMF)         & 61.1 & 77.4 \\
      \midrule
      Comp.VAE (\second{ours}) & \textbf{23.2} & \second{89.5} \\
      \bottomrule
    \end{tabular}}
  \end{minipage}
  \begin{minipage}[t]{0.35\textwidth}
    \centering
    \label{tab:nearood-imagenette}
    \setlength{\tabcolsep}{6pt}
    \resizebox{\linewidth}{!}{%
    \begin{tabular}{l rr}
      \toprule
      \textbf{Methods} & \multicolumn{2}{c}{\textbf{Near-conditional-OOD}} \\
      \cmidrule(lr){2-3}
      & FPR95$\downarrow$ & AUROC$\uparrow$ \\
      \midrule
      KNN*               & 89.9 & 58.9 \\  
      Comp.VAE (vMF)    & \second{82.3} & \second{66.4} \\
      \midrule
      Comp.VAE (\textit{ours}) & \textbf{78.8} & \textbf{68.3} \\
      \bottomrule
    \end{tabular}}
  \end{minipage}
\end{table*}

\paragraph{3d visualization of the hypersphere} Our method of volume compression allows for a direct $3-$dimensional visualization of the HD latent space. It shows the latent of a standard VAE is uniformly distributed over the sphere and not informative, contrary to our case (Figure \ref{fig:Figure-summary}). This was done by averaging the 256 latent dimensions into three (first 85, second 85, and the remaining 86), and normalizing each of the resulting 3D vectors to the sphere. Each HD latent vector could thus be plotted as a point in 3D and in this way visualize the HD latent space in a rather direct way. Furthermore, the \(3-\)dimensional visualization shows something remarkable in the case of the compressed VAE: the latent samples are compressed towards a small `island' on the hypersphere and away from the equator, but the classes are actually \textit{visible}, unlike the case of the standard VAE. We believe that it is because the samples are now located away from the equator, in a region with a much lower volume, where there are many fewer possibilities to realize this clustering in terms of different possible latent point configurations. 


\section{Discussion and conclusion}

We propose to convert the latent variables of a VAE to hyperspherical coordinates. This allows moving latent vectors on a small island of the hypersphere. We showed that this modification improves AD (in both fully unconditional-OOD and conditional-OOD modalities, for the latter also in the far and near conditional-OOD types) as our results outperform other comparable methods in many cases.

We report state-of-the art results on the FPR95 metric in the conditional-OOD experiments of CIFAR-10 (ID) vs far conditional-OOD standard benchmarks (w.r.t. methods that, like ours, do not use contrastive learning techniques, which by themselves seem to enhance most methods regardless of their inner details), and a very strong one in the case of near conditional-OOD for CIFAR-10 (ID) vs CIFAR-100. We also explored the complex and difficult ImageNet-based near conditional-OOD scenario of Imagenette (ID) vs close ImageNet classes. Our method showed the best results in both metrics, while at the same time providing an important ablation w.r.t. another close type of methods based on vMF distributions, which our own method can actually reproduce as a sub-case (compress only one angular coordinate instead of all of them). Our compression method can be used in \textit{most of the} VAE variations, since it just affects how the KLD term is computed.

The transformation from Cartesian to hyperspherical coordinates adds processing time, despite a vectorized implementation. Computation for an epoch was 32\% more expensive in time using 200 latent dimensions. As the number of dimensions grows, the computation becomes more expensive (see App.~\ref{appendix:trainspeed} for details).

The constants $\alpha_{i,j},\,\beta_{i,j}$ multiplying the elements of the hyperspherical loss are proportional to $1/\sqrt{k+1}$, where \(k\) is the coordinate index. The number of free hyperparameters is thus reduced to  \textit{four scalars} for the angular losses and four scalars for the radial ones. However, we found that only two of them ($\beta$ and $\alpha_{\mu,r}$, the latter only occasionally) need adjustment when changing datasets (at least for the ones used in this paper). Those parameters can be found in the code provided here (to be updated upon acceptance).

This work stemmed from the hypothesis that the latent of a VAE is very sparse, which limits its ability to be used as a generative model because most of the latent is not sampled during training. Recent work showed that compressing the latent using hyperspectral coordinates does indeed improve the generation of new data when sampling the prior \cite{Ascarate2025ImprovingCoordinates}. In this new work we show that, as expected, reducing latent sparsity also helps to detect anomalies. We speculate that controlling and reducing the sparsity of the high dimensional latent manifold should be beneficial for other tasks such as classification, which we aim to explore next. 

\clearpage

\newpage

\clearpage


\appendix
\numberwithin{equation}{subsection}

\counterwithin{figure}{subsection}

\subsection{Hyperspherical
coordinates}\label{Hyperspherical
coordinates}

\subsubsection{Conversion between Cartesian and hyperspherical coordinates}\label{appendix:hstransform}

For reference, we provide the standard formulas for converting between Cartesian and spherical coordinates.

In $n$ dimensions, given a set of Cartesian coordinates $x_k$ with $k\in\{1,\hdots,n\}$, the hyperspherical coordinates are defined by a radius $r$ and $n-1$ angles $\varphi_k$ with $k\in\{1,\hdots,n-1\}$; $\varphi_k \in [0,\hdots,\pi]$ for $k\in\{1,\hdots,n-2\}$ and $\varphi_{n-1} \in [0,\hdots,2\pi)$. 

From hyperspherical to Cartesian conversion:

\begin{equation}
\begin{aligned}
x_1 = & {} r \cos(\varphi_1) \\
x_2 = & {} r \sin(\varphi_1)\cos(\varphi_2) \\
x_2 = & {} r \sin(\varphi_1)\sin(\varphi_2)\cos(\varphi_3) \\
\vdots \\
x_{n-1} = & {} r \sin(\varphi_1)\sin(\varphi_2) \hdots \sin(\varphi_{n-2})\cos(\varphi_{n-1}) \\
x_{n} = & {} r \sin(\varphi_1)\sin(\varphi_2) \hdots \sin(\varphi_{n-2})\sin(\varphi_{n-1}) 
\end{aligned}
\end{equation}

From Cartesian to hyperspherical conversion:

\begin{equation}
\begin{aligned}
r = & {} \sqrt{x_{n}^2+x_{n-1}^2+\hdots+x_2^2+x_1^2} \\
\cos(\varphi_{1}) = & {} \frac{x_1}{ \sqrt{x_{n}^2+x_{n-1}^2+\hdots+x_2^2+x_1^2}}\\
\cos(\varphi_{2}) = & {} \frac{x_2}{ \sqrt{x_{n}^2+x_{n-1}^2+\hdots+x_2^2}}\\
\vdots \\
\cos(\varphi_{n-2}) = & {} \frac{x_{n-2}}{ \sqrt{x_{n}^2+x_{n-1}^2+x_{n-2}^2}}\\
\cos(\varphi_{n-1}) = & {} \frac{x_{n-1}}{ \sqrt{x_{n}^2+x_{n-1}^2}}\\
\end{aligned}
\end{equation}

\subsubsection{Vectorized code for converting between Cartesian and hyperspherical coordinates}\label{appendix:hstransformcode}

This code is accessible here and provided below for reference. It uses vectorized operations that are differentiable by torch. The correct vectorization of this script was key to make our proposal feasible in practice, since the introduction of any Python \textbf{for loop} increased the processing time considerably.

\begin{lstlisting}
import torch

def cart_to_cos_sph (x, device):
    m = x.size(0)
    n = x.size(1)
    mask = torch.triu(torch.ones(n, n)).to(device)
    mask = torch.unsqueeze(mask, dim=0)
    mask = mask.expand(m, n, n)
    X = torch.unsqueeze(x, dim=1).expand(m, n, n)
    X_squared = torch.square(X)
    X_squared_masked = X_squared * mask
    denom = torch.sqrt(torch.sum(X_squared_masked, dim=2)+0.001)
    cos_phi = x / denom
    return cos_phi[:, 0:n-1]
\end{lstlisting}

\subsection{Concentration of measure effects}\label{appendix:measurecon}

\subsubsection{Concentration of Measure: basic idea}

\begin{figure*}[!h]
    \centering
    \includegraphics[width=1\linewidth]{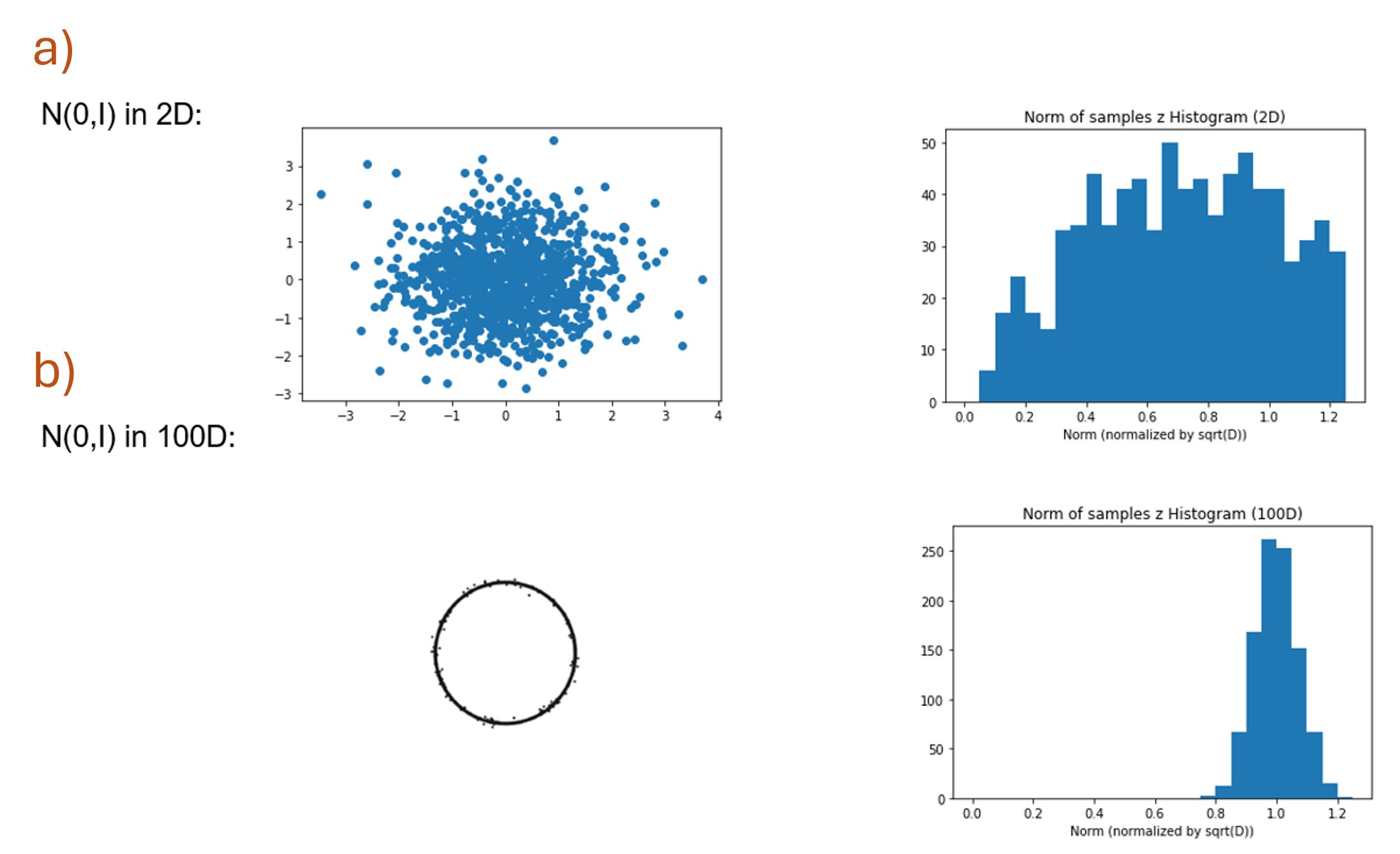}
    \caption{Measure concentration, norm (left image in b), adapted from \cite{Vershynin2018High-DimensionalProbability})}
    \label{fig:4}
\end{figure*}

\begin{figure*}[!h]
    \centering
    \includegraphics[width=1\linewidth]{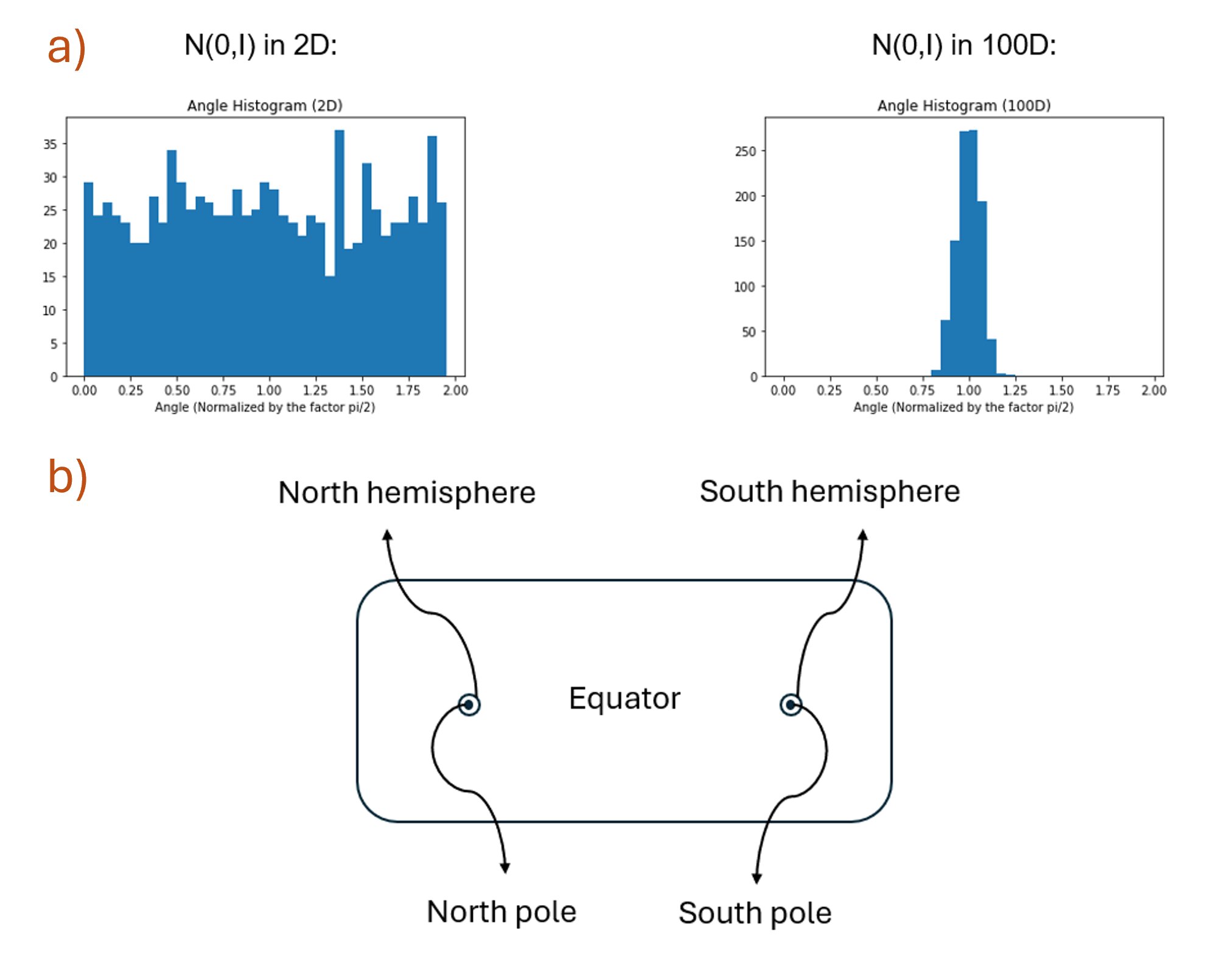}
    \caption{a) Measure concentration, angle; b) Schematic diagram of the mass concentration of the uniform measure of the hypersphere in very high dimensions: most of the volume is in any equator.}
    \label{fig:5}
\end{figure*}

In this appendix, we collect the results of simple experiments that clearly show the concentration of measure effects that occur in high dimensions. In Fig. \ref{fig:4}a), we show the distribution of a simple Normal distribution in $2$ dimensions (left), and the histogram for the norm of the samples (right). In b), the same but for a Normal distribution in $100$ dimensions. In Fig. \ref{fig:5}a), we show the histogram for the angle between two random samples from a Normal distribution in $2$ dimensions (left), and the same but for a Normal distribution in $100$ dimensions (right). In b), we display a schematic diagram of the mass concentration of the uniform measure of the hypersphere in very high dimensions. The intuition in this diagram comes from the more precise result \cite{Wainwright2019ConcentrationMeasure} which states that, for \underline{\textbf{any}} given $y\in\mathbb{R}^{n}$, if we define on the hypersphere an `\textbf{equatorial}' slice of width $\epsilon>0$ as $T_{y}(\epsilon)\doteq\left\{ z\in\mathbb{S}^{n-1}/\mid(z,y)\mid\leq\epsilon/2\right\}$, then its volume according to the uniform measure satisfies the following concentration inequality:
\begin{equation}
\mathbb{P}\left[T_{y}(\epsilon)\right]\geq1-\sqrt{2\pi}\exp(-\frac{n\epsilon^{2}}{2}).
\end{equation}

The previous inequality shows that, in very high dimensions, the equatorial slice $T_{y}(\epsilon)$ occupies a huge portion of the total volume, even for a very small width.

\begin{figure*}[!h]
    \centering
    \includegraphics[width=1\linewidth]{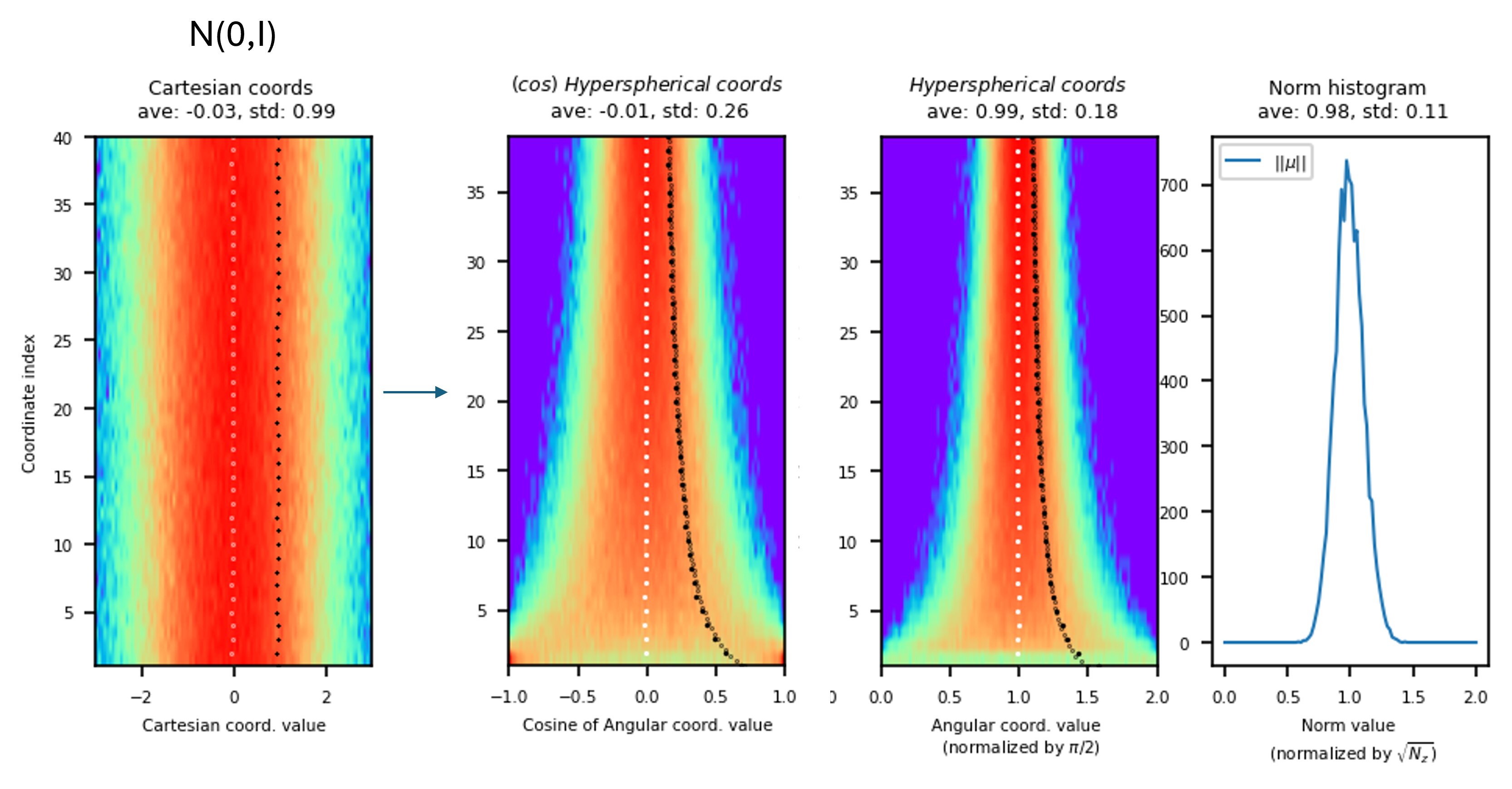}
    \caption{High dimensional Normal distribution in hyperspherical coordinates. For the first three images from the left, each horizontal slice at some vertical index value shows the color coded histogram (red, high density; blue, low density) for the range of the coordinate of that index; the vertical axis stacks all the histograms for all the dimensions (in this example, $40$). The white dots represent the mean and the black dots represent the standard deviation of the corresponding histogram. The numbers on top are the total mean and standard deviation of all these previous values taken together.}
    \label{fig:6}
\end{figure*}

Finally, with this in place, we can understand the peculiar shape that a high dimensional Normal distribution takes when expressed in hyperspherical coordinates (Fig.\ref{fig:6}).

\newpage

\clearpage

\subsubsection{Hypervolume reduction in hyperspherical coordinates}\label{appendix:volumeelement}

The hypervolume element of the hypersphere $\mathbb{S}_{R}^{n-1}$ is given by the following expression when using hyperspherical coordinates\footnote{(see \url{https://en.wikipedia.org/wiki/N-sphere})}:

\begin{equation}
\mathrm{d}V_{\mathbb{S}_{R}^{n-1}}=R^{n-1}\sin^{n-2}(\varphi_{1})\sin^{n-3}(\varphi_{2})\cdots \sin(\varphi_{n-2})\times\nonumber
\end{equation}
\begin{equation}
\mathrm{d}\varphi_{1}\mathrm{d}\varphi_{2}\cdots\mathrm{d}\varphi_{n-1}.
\end{equation}\label{volele}

The volume can be reduced much faster and effectively by reducing the angular coordinates (away from the equators), than by either reducing just the radius of the hypersphere or, equivalently, all of the Cartesian coordinates. The higher the dimension, the more pronounced this difference becomes because each added dimension $k$ adds extra powers of $\sin\varphi_k$ in the hypervolume element (see section Hypervolume reduction in hyperspherical
coordinates). The further the angles from \(\pi/2\), the smaller the infinitesimal hypervolume element becomes as it is multiplied by an increasingly smaller quantity lower than $1$. This is a purely geometric effect.

As mentioned, the hypervolume element of the hypersphere $\mathbb{S}_{R}^{n-1}$ is given by the expression of equation \ref{volele} when using hyperspherical coordinates. In the small angle regime, where $\sin\varphi\approx\varphi$, we can approximately integrate this expression for an angular coordinate hypercube $\left[0,\varphi_{0}\right]^{n-1}$, and the result is proportional to $v_0=R^{n-1}\varphi_{0}^{n(n-1)/2}$. If now we reduce the size of the angular coordinate hypercube by a schedule of the form $\varphi_{t}=\varphi_{0}(1-t),\,t\in\left[0,1\right]$, then we can compare the percentage of hypervolume being reduced from the initial value, while keeping $R$ fixed, to the percentage obtained by reducing the size of the hypersphere by an schedule of the form $R_{t}=R(1-t),\,t\in\left[0,1\right]$, while keeping $\varphi_0$ fixed (this second case is equivalent to reducing all the Cartesian coordinates at once, because $r^2 = x_{n}^2+x_{n-1}^2+\hdots+x_2^2+x_1^2$). Indeed, we get, respectively, $v_t=v_0(1-t)^{n(n-1)/2}$ and $v_t=v_0(1-t)^{n-1}$. 

In Fig.~\ref{fig:6b} we plot the behavior of $v_t/v_0$ in terms of the reduction of the coordinate, given by $(1-t)$, for three, increasing values of dimension $n$. As we can see, already in dimension $20$ (bottom figure in the panel), there is a sharp decrease in volume in the angular case as soon as one decreases the angular coordinates by a minimal amount; in comparison to the radial/Cartesian coordinate case, the abrupt decrease in volume looks almost discontinuous (in Statistical Physics, this type of behavior often occurs when the system undergoes a phase transition \cite{Yeomans1992StatMechPhaseTransitions}). 

Reducing the radius of the ball centered at the prior for the means $\mu$ given by the encoder is how the standard VAE tries to reach this prior: this corresponds to the term of the form $\parallel \mu\parallel^{2}$ in the KLD part in Cartesian coordinates, $\text{KLD} = -\frac{1}{2} \sum^{n}_{k=1} \left(  1+\log\,(\sigma^2_{k}) - \sigma^2_{k} \right)+\frac{1}{2}\parallel \mu\parallel^{2}$ (see \cite{DiederikPKingma2014Auto-EncodingBayes,Kingma2019AnAutoencoders}).

The most general form is $\parallel \mu-\mu_0\parallel^{2}$, where $\mu_0$ is the prior value, the north pole in our case; that is, in both this case and in our method, the means $\mu$ given by the encoder are encouraged to go to the north pole by the losses, but what is really important in high dimensions is \textit{\textbf{how}} we reach it, because of the peculiar `equatorial' presentation of the volume in the hypersphere: we claim that our method using hyperspherical coordinates is better suited for this, since it allows to reduce the volume much faster by directly engaging with moving the samples away from the equators, thing which the radial reduction cannot do by construction.

\subsubsection{Comparison with von Mises-Fisher-based approaches}\label{Comparison with von Mises-Fisher-based approaches}

\begin{figure*}[!h]
    \centering
    \includegraphics[width=1\linewidth]{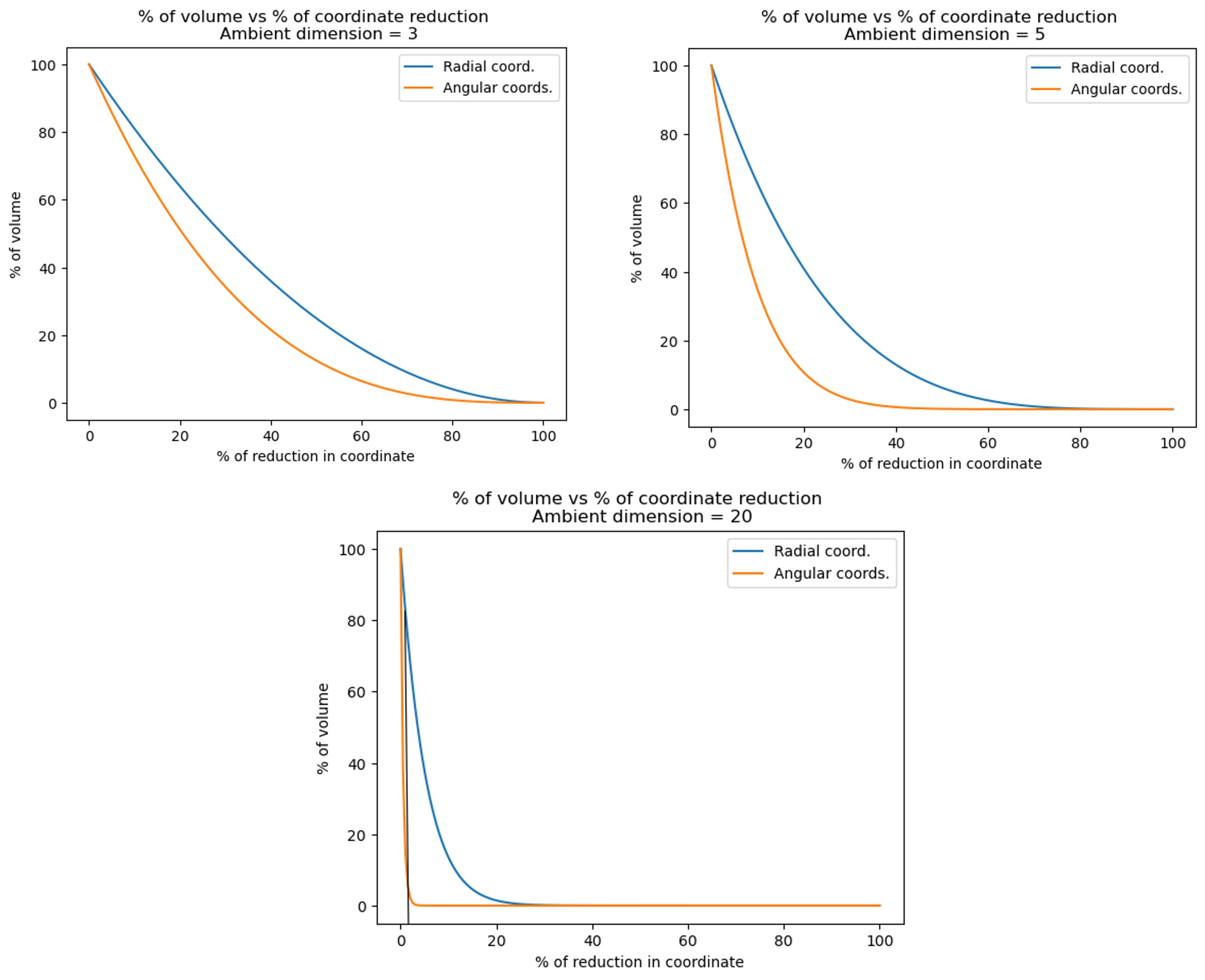}
    \caption{Hypervolume element reduction comparison: $(1-t)^{n-1}$ vs. $(1-t)^{n(n-1)/2}$.}
    \label{fig:6b}
\end{figure*}

The case when one reduces \textit{only} a single angular coordinate, e.g., $\varphi_1$, is exactly like the radial one, since in equation \ref{volele} the exponent for this angle is $n-2$ which upon integration becomes $n-1$, like for the radial coordinate. Thus, the tendencies in the analogous of Fig.~\ref{fig:6b} look exactly the same. This case corresponds to varying the single scalar variance parameter on a von Mises-Fisher distribution on the hypersphere, which is defined as an isotropic (hence the single free parameter rather than a vector or matrix for this variance) Gaussian whose domain is restricted to the hypersphere. This approach has been used recently in the literature when dealing with encodings into the hypersphere \cite{Ming2023HowDetection,SuvraGhosal2024HowDetection}, but as we mentioned, it cannot reduce the hypervolume as fast as our method in case one wishes to do so.

\onecolumn
 
\newpage

\clearpage

\twocolumn

\subsection{Re-writing of the KLD term}\label{appendix:KLD}

In this appendix, we make explicit the steps to go from the standard form of the KLD term in the VAE to the one we used as a starting point, in equation (1) of the main paper, for our own KLD in hyperspherical coordinates.

In Cartesian coordinates, the KLD divergence between the estimated posterior defined by $\mu_k$ and $\sigma_k$ and the prior defined by $\mu_k^p$ and $\sigma_k^p$: 

\begin{equation}
\text{KLD}_{\text{CartCoords}}^{w/Prior}=\nonumber
\end{equation}
\begin{equation}
\frac{1}{2}\sum_{k=1}^{n}\left[\left(\frac{\sigma_{k}}{\sigma_{k}^{p}}\right)^{2}-\log\,\left(\frac{\sigma_{k}}{\sigma_{k}^{p}}\right)^{2}-1+\frac{\left(\mu_{k}-\mu_{k}^{p}\right)^{2}}{\left(\sigma_{k}^{p}\right)^{2}}\right]\nonumber
\end{equation}

A Taylor approximation (up to second order) of the part for sigma around its prior yields for some constants \(\gamma_k\) and \(\widetilde{\gamma}_k\):

\begin{equation}
\text{KLD}_{\text{CartCoords}}^{w/Prior}\approx\sum_{k=1}^{n}\left[\gamma_{k}\left(\sigma_{k}-\sigma_{k}^{p}\right)^{2}+\widetilde{\gamma}_{k}\left(\mu_{k}-\mu_{k}^{p}\right)^{2}\right]\nonumber
\end{equation}

In practice, the optimization is performed over mini batches of data (of size \(N_b\)), using the objective below:

\begin{equation}
\text{KLD}_{\text{CartCoords}}^{w/Prior}\approx\nonumber
\end{equation}
\begin{equation}
\frac{1}{N_b}\sum_{l=1}^{N_b}\sum_{k=1}^{n}\left(\gamma_{k}\left(\sigma_{k,l}-\sigma_{k}^{p}\right)^{2}+\widetilde{\gamma}_{k}\left(\mu_{k,l}-\mu_{k}^{p}\right)^{2}\right)\nonumber
\end{equation}

If we denote the corresponding batch statistics as \(\mathbb{E}_b\) and \(\sigma_b\), then, by using the basic formula,

\begin{equation}
\mathbb{E}_b[X^{2}]=\mathbb{E}_b[X]^{2}+\sigma_b[X]^{2},\nonumber
\end{equation}

we can write this objective as (we omit the constants for ease of reading)

\begin{equation}
\text{KLD}_{\text{CartCoords}}^{w/Prior}\approx\nonumber
\end{equation}
\begin{equation}
\sum_{k=1}^{n} \left( \left( \mathbb{E}_{b}[\sigma_{k}]-\sigma_{k}^{p}\right)^{2}+\sigma_{b}[\sigma_{k}]^{2}+\left(\mathbb{E}_{b}[\mu_{k}]-\mu_{k}^{p}\right)^{2}+\sigma_{b}[\mu_{k}]^{2} \right) \nonumber
\end{equation}

\newpage

\clearpage

\onecolumn

\subsection{Datasets}\label{appendix:Datasets}
$\boxed{\text{Used in the unconditional-OOD experiments}}$
\subsubsection{Mars Rover Mastcam}\label{Mars Rover Mastcam}

\begin{figure*}[!htbp]
    \centering
    \includegraphics[width=.8\linewidth,height=.6\textheight]{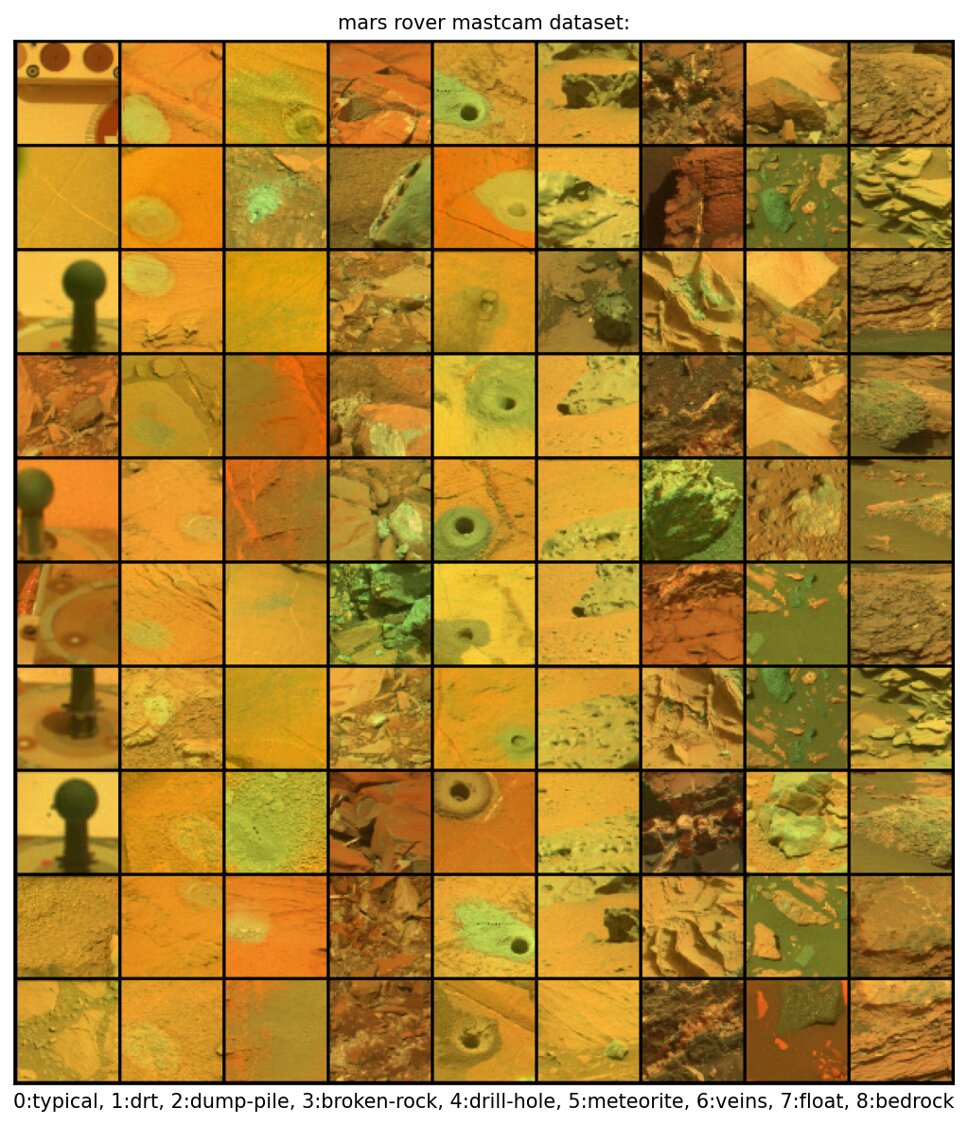}
    \caption{Examples from the test set of the Mars Rover Mastcam dataset. Images on the same column are in the same class.}
    \label{fig:7}
\end{figure*}

\begin{figure*}[!h]
    \centering
    \includegraphics[width=1\linewidth]{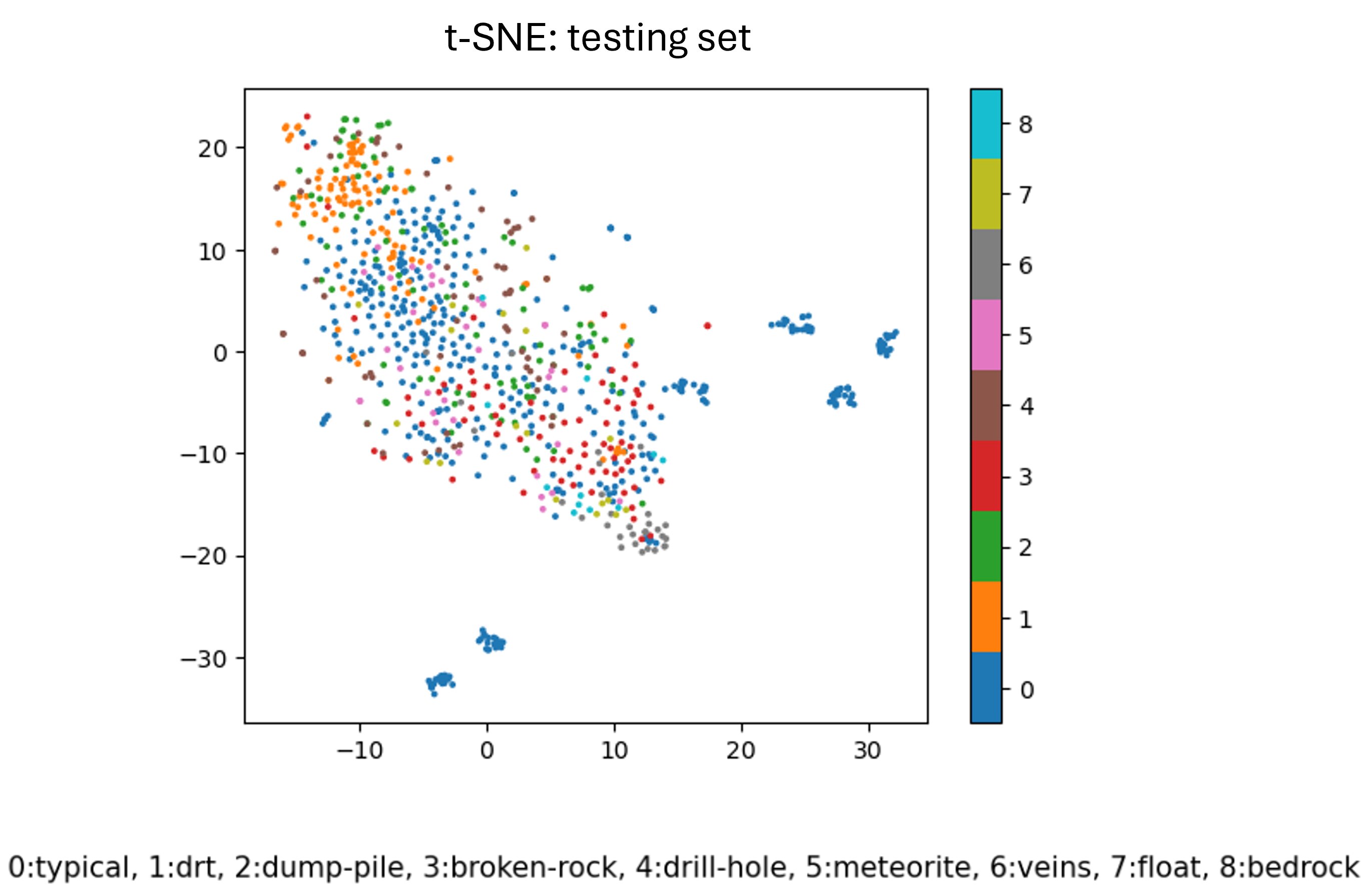}
    \caption{t-SNE of the test set of the Mars Rover Mastcam dataset (pixel-space).}
    \label{fig:8}
\end{figure*}

\begin{figure*}[!h]
    \centering
    \includegraphics[width=1\linewidth]{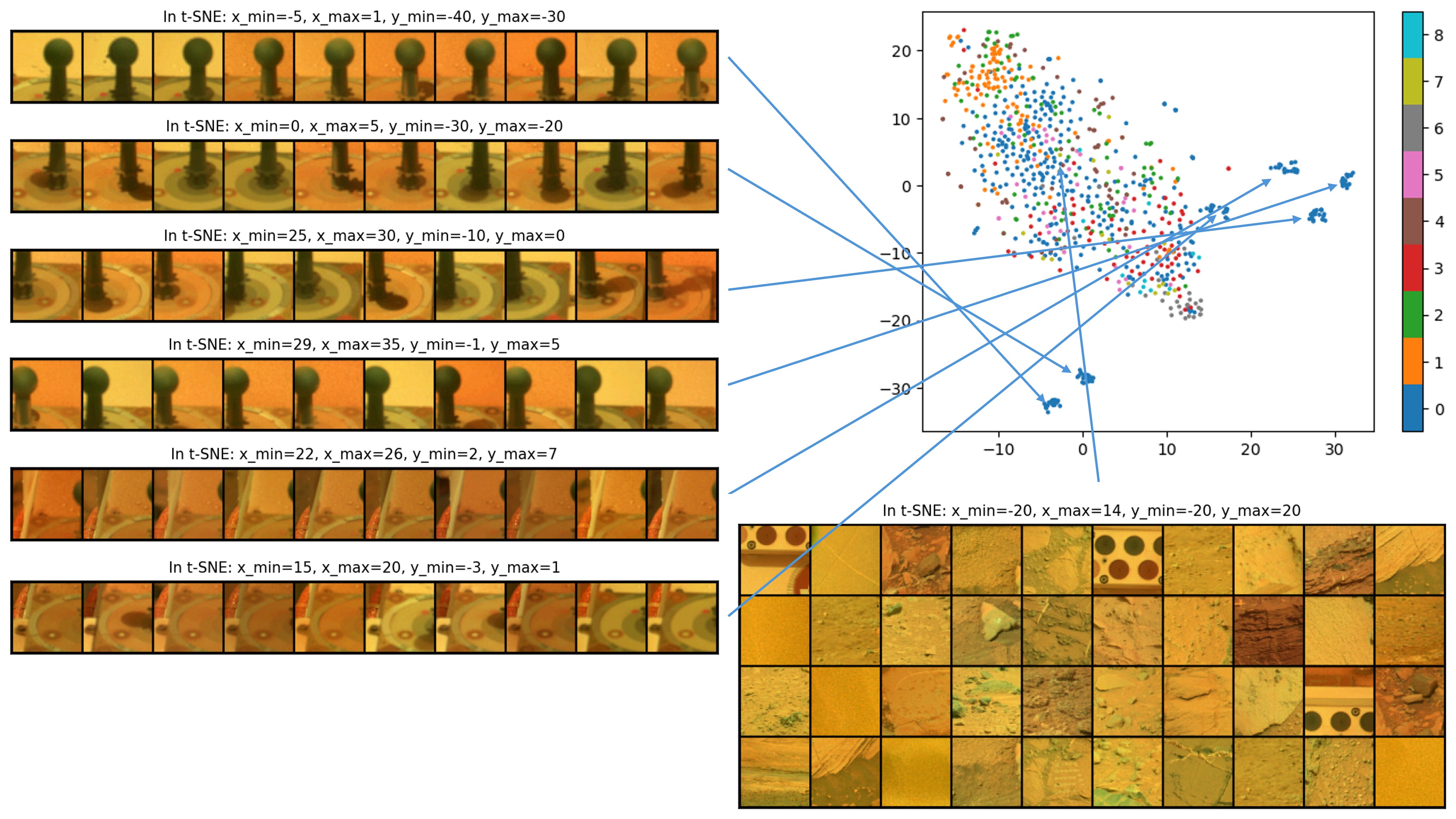}
    \caption{The different sub-classes of the normal class (blue) of the test set of the Mars Rover Mastcam dataset.}
    \label{fig:9}
\end{figure*}

\newpage

\clearpage

\subsubsection{Galaxy Zoo}\label{Galaxy Zoo}

\begin{figure*}[!h]
    \centering
    \includegraphics[width=1\linewidth]{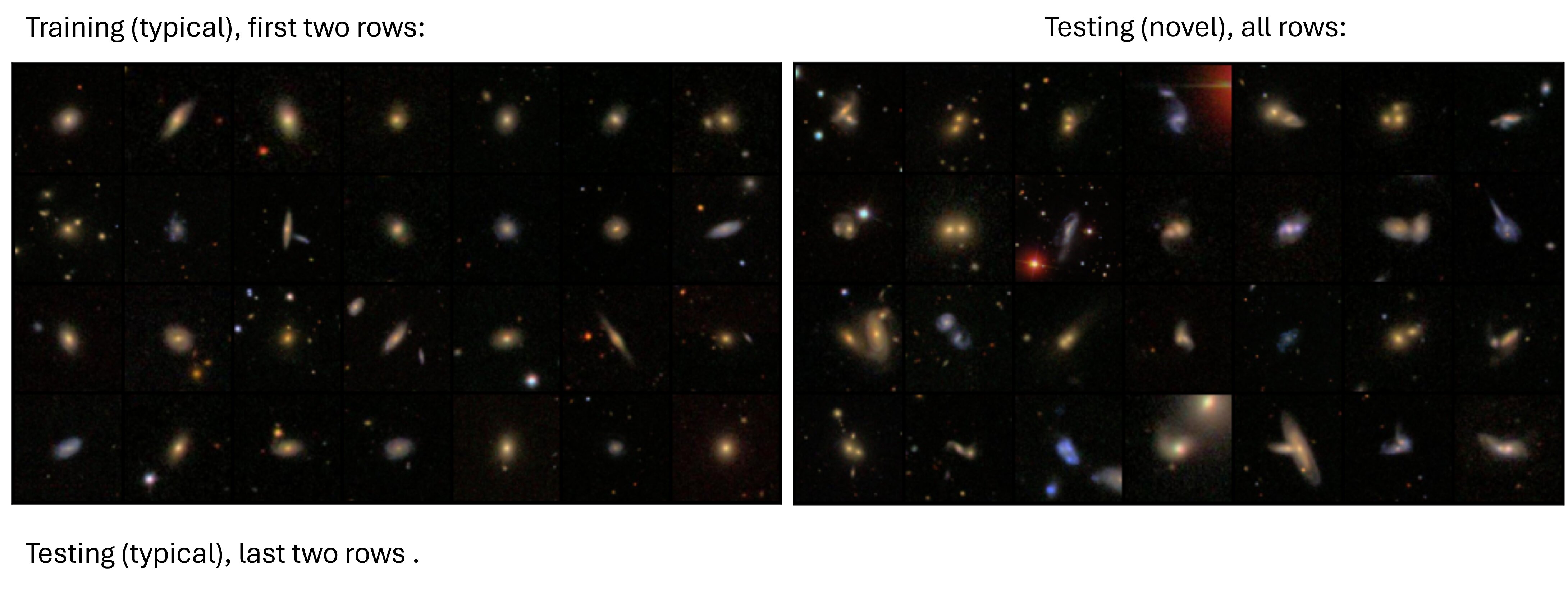}
    \caption{Examples from the Galaxy Zoo dataset. Left: normal/typical. Right: novel.}
    \label{fig:10}
\end{figure*}

\begin{figure*}[!h]
    \centering
    \includegraphics[width=1\linewidth]{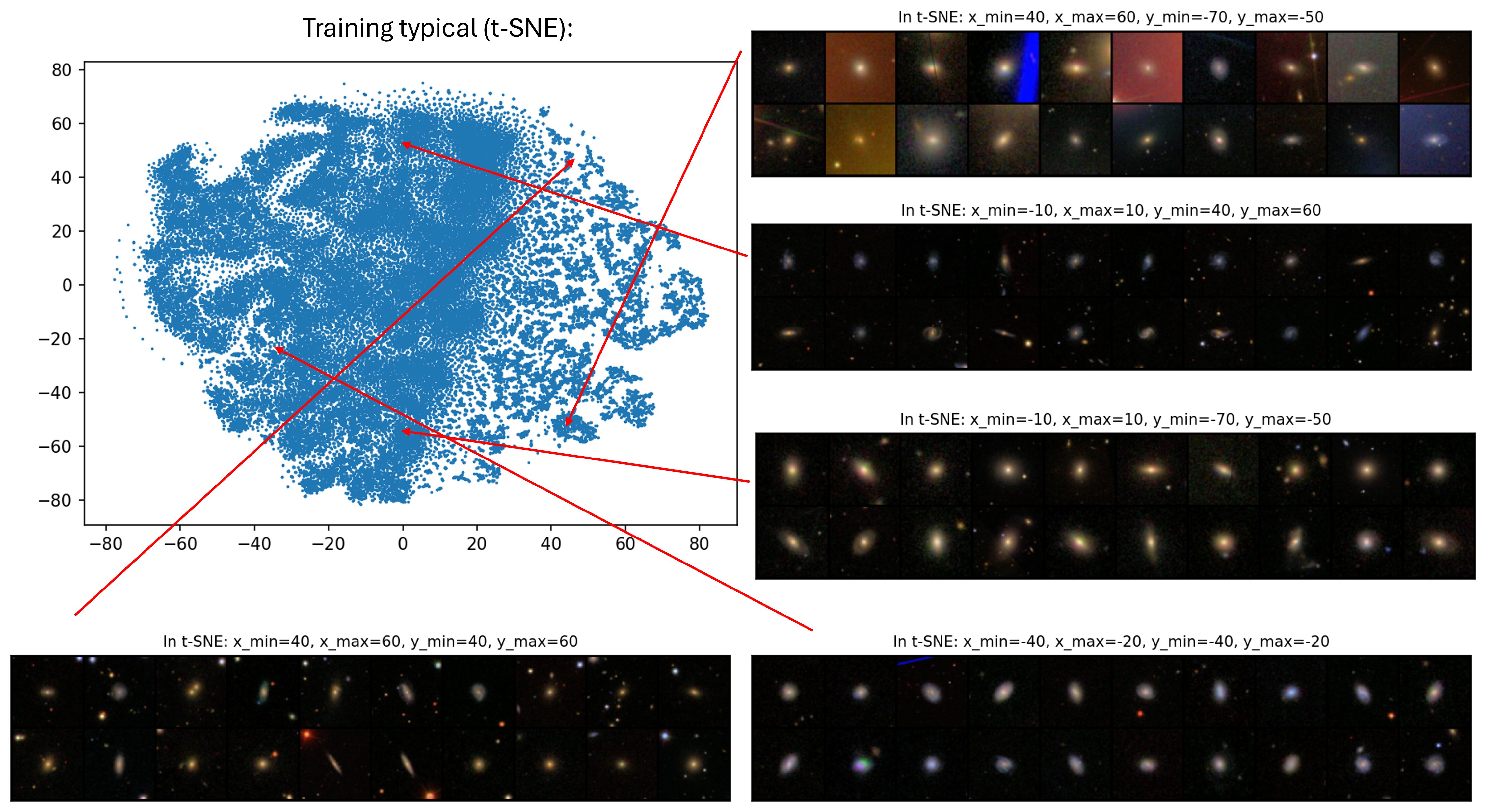}
    \caption{The different sub-classes of the normal class (blue) of the train set of the Galaxy Zoo dataset (t-SNE in pixel-space).}
    \label{fig:11}
\end{figure*}

\newpage

\clearpage

\subsubsection{ImageNet-based dataset used in the conditional-OOD experiments}\label{ImageNet-based dataset}

The selection criterion for the near-conditional-OOD ImageNet classes was based on a simple semantic similarity search on the text corresponding to the classes' description.

\begin{table*}[!h]
  \centering
  \small
  \caption{Imagenette $\rightarrow$ Near-conditional-OOD ImageNet mapping. The third column lists the WordNet/ILSVRC synset ID.}
  \label{tab:imagenette_near_ood_map}
  \setlength{\tabcolsep}{10pt}
  \begin{tabularx}{\textwidth}{L L l}
    \toprule
    \textbf{Imagenette class} & \textbf{Near-conditional-OOD ImageNet class} & \textbf{Near-conditional-OOD synset ID} \\
    \midrule
    tench            & goldfish               & n01443537 \\
    English springer & Welsh springer spaniel & n02102177 \\
    cassette player  & CD player              & n02988304 \\
    chain saw        & hand blower            & n03483316 \\
    church           & mosque                 & n03788195 \\
    French horn      & cornet                 & n03110669 \\
    garbage truck    & moving van             & n03796401 \\
    gas pump         & cash machine           & n02977058 \\
    golf ball        & tennis ball            & n04409515 \\
    parachute        & kite                   & n01608432 \\
    \bottomrule
  \end{tabularx}
\end{table*}

\newpage

\clearpage

\subsection{Results}\label{appendix:Results}

Note: what we call the `replica angle' (dashed lines) is the angle between the testing samples and the mean for all the normal test set (this angle value should give a rough idea about the angular size of the island). For more detailed information, we plot the full histograms of the hyperspherical coordinates.

$\boxed{\text{Fully unconditional-OOD experiments}}$

\subsubsection{Standard VAE on the Mars Rover Mastcam dataset}\label{Standard VAE on the Mars Rover Mastcam dataset}

\begin{figure*}[!h]
    \centering
    \includegraphics[width=0.93\linewidth]{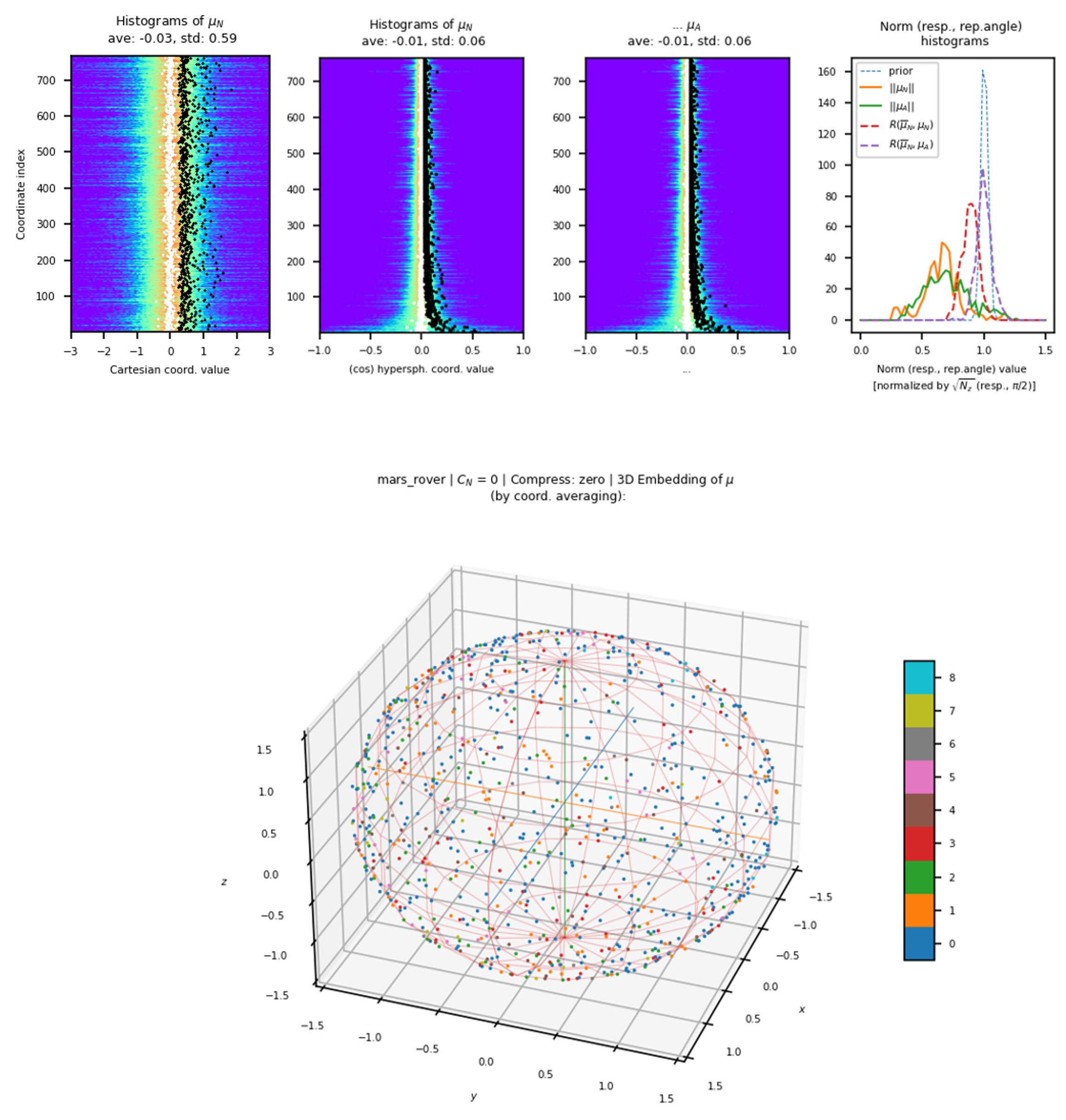}
    \caption{Results of the standard VAE training, on the test set of the Mars Rover Mastcam dataset. In the upper panel we show, from left to right, the histograms of $\mu$, using the same conventions as in Fig.\ref{fig:6}. The third histogram in this panel shows the norm histograms of $\mu$ and the replica angles.}
    \label{fig:12}
\end{figure*}

\begin{figure*}[!h]
    \centering
    \includegraphics[width=1\linewidth]{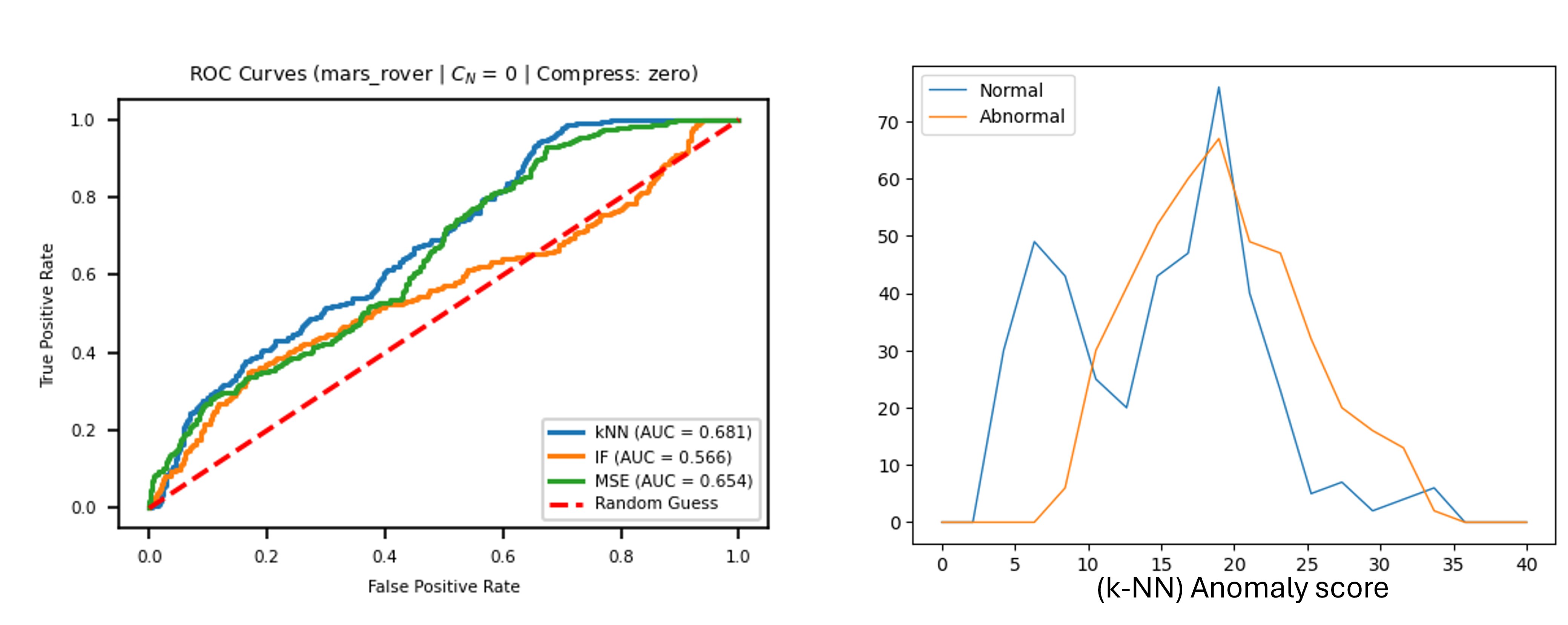}
    \caption{ROC curves and anomaly score histograms, results of the standard VAE training, on the test set of the Mars Rover Mastcam dataset. AUCs: Only values up to the third decimal place are reported, as results beyond that fluctuate across different runs of the same experiment.}
    \label{fig:14}
\end{figure*}

\begin{figure*}[!h]
    \centering
    \includegraphics[width=1\linewidth]{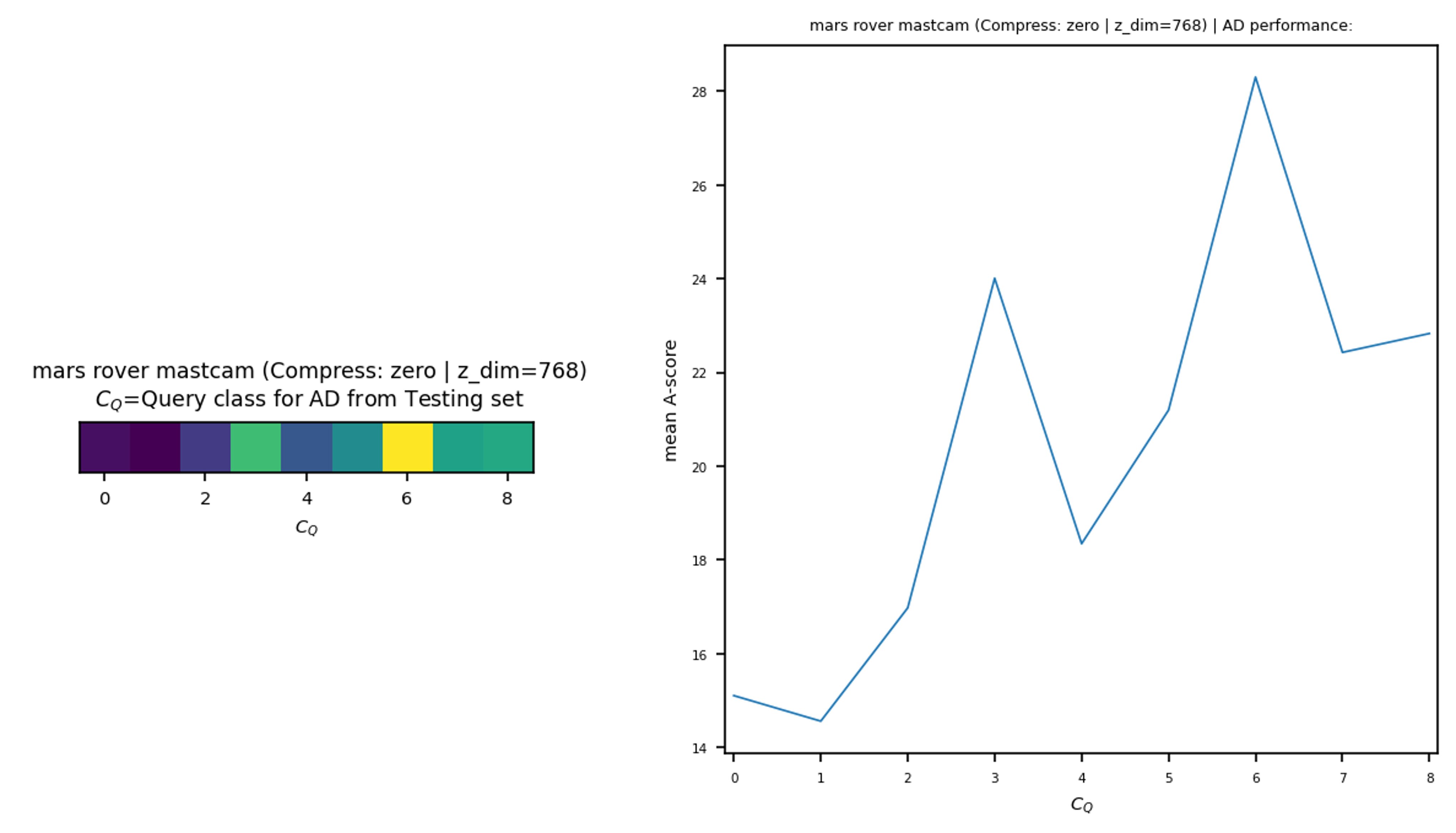}
    \caption{Mean $k-$NN anomaly score values for each class in the test set (0 is the typical/normal class), results of the standard VAE training, on the test set of the Mars Rover Mastcam dataset (blue means low, yellow is high).}
    \label{fig:15}
\end{figure*}

\begin{figure*}[!h]
    \centering
    \includegraphics[width=1\linewidth]{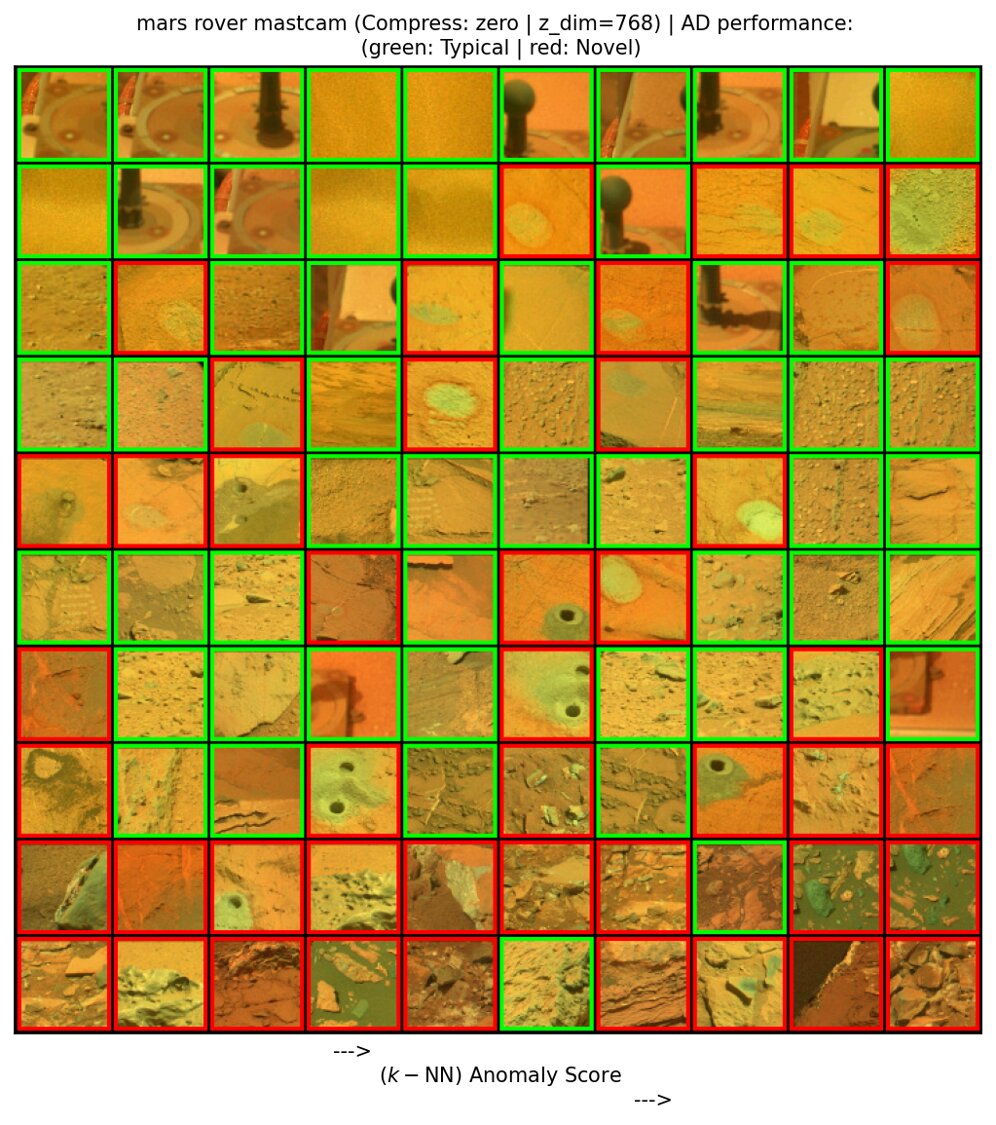}
    \caption{Results of the standard VAE training, on the test set of the Mars Rover Mastcam dataset. Showing the images from the test set but with the index sorted by the anomaly score (from low at the top to high at the bottom). Only displaying one every 8 images, starting from index 0. There are 420 typical images and 435 novel images in the testing set, for a total of 855. A perfect detection would show the top half of the total panel as normal (green framing) and the anomalies (red framing) at the bottom half.}
    \label{fig:16}
\end{figure*}

\begin{figure*}[!h]
    \centering
    \includegraphics[width=1\linewidth]{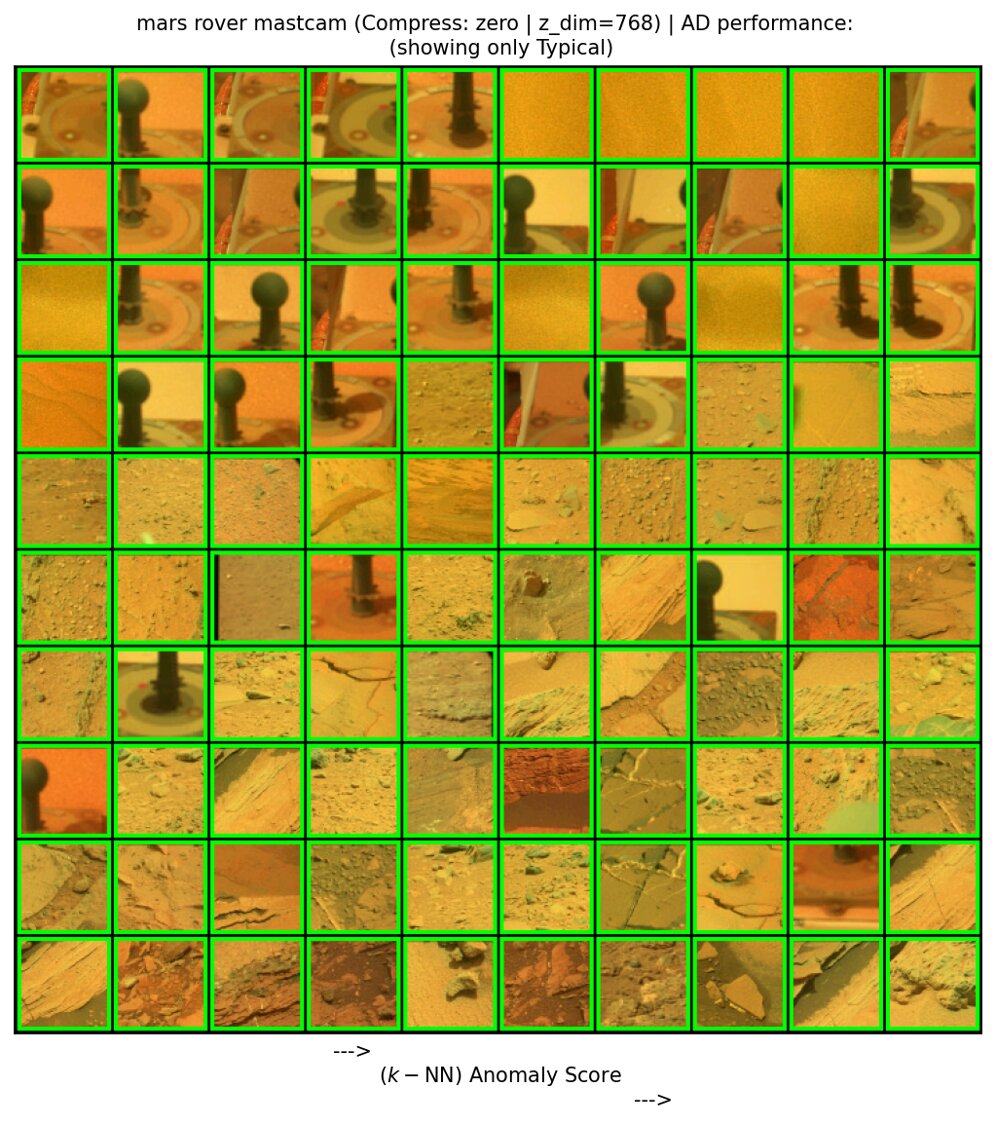}
    \caption{Results of the standard VAE training, on the test set of the Mars Rover Mastcam dataset. Showing the images from the Typical test set only but with the index sorted by the anomaly score (from low to high). Only displaying one every 4 images, starting from index 0.}
    \label{fig:17}
\end{figure*}

\begin{figure*}[!h]
    \centering
    \includegraphics[width=1\linewidth]{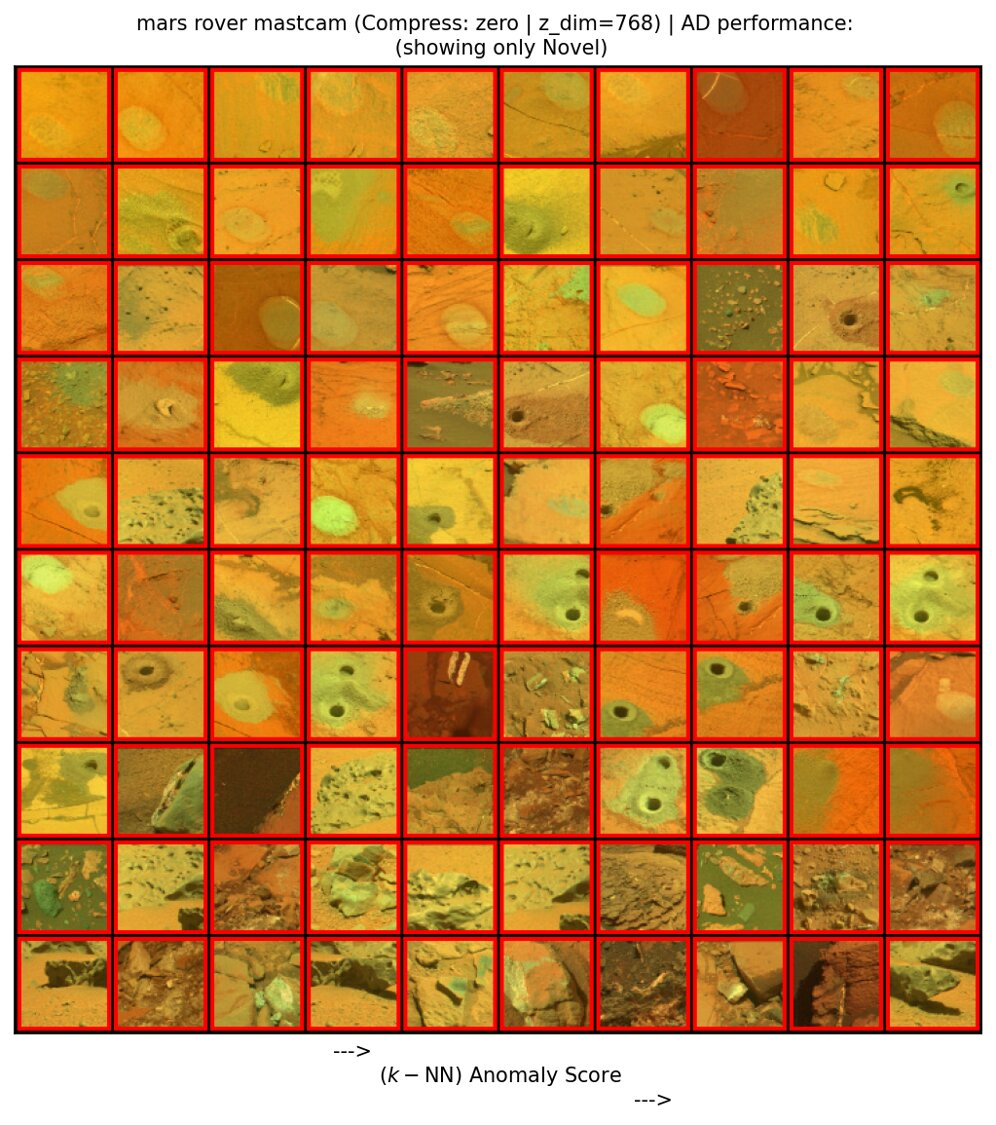}
    \caption{Results of the standard VAE training, on the test set of the Mars Rover Mastcam dataset. Showing the images from the Novel test set only but with the index sorted by the anomaly score (from low to high). Only displaying one every 4 images, starting from index 0.}
    \label{fig:18}
\end{figure*}

\newpage

\clearpage

\subsubsection{Comp.VAE on the Mars Rover Mastcam dataset}\label{Comp.VAE on the Mars Rover Mastcam dataset}

\begin{figure*}[!h]
    \centering
    \includegraphics[width=1\linewidth]{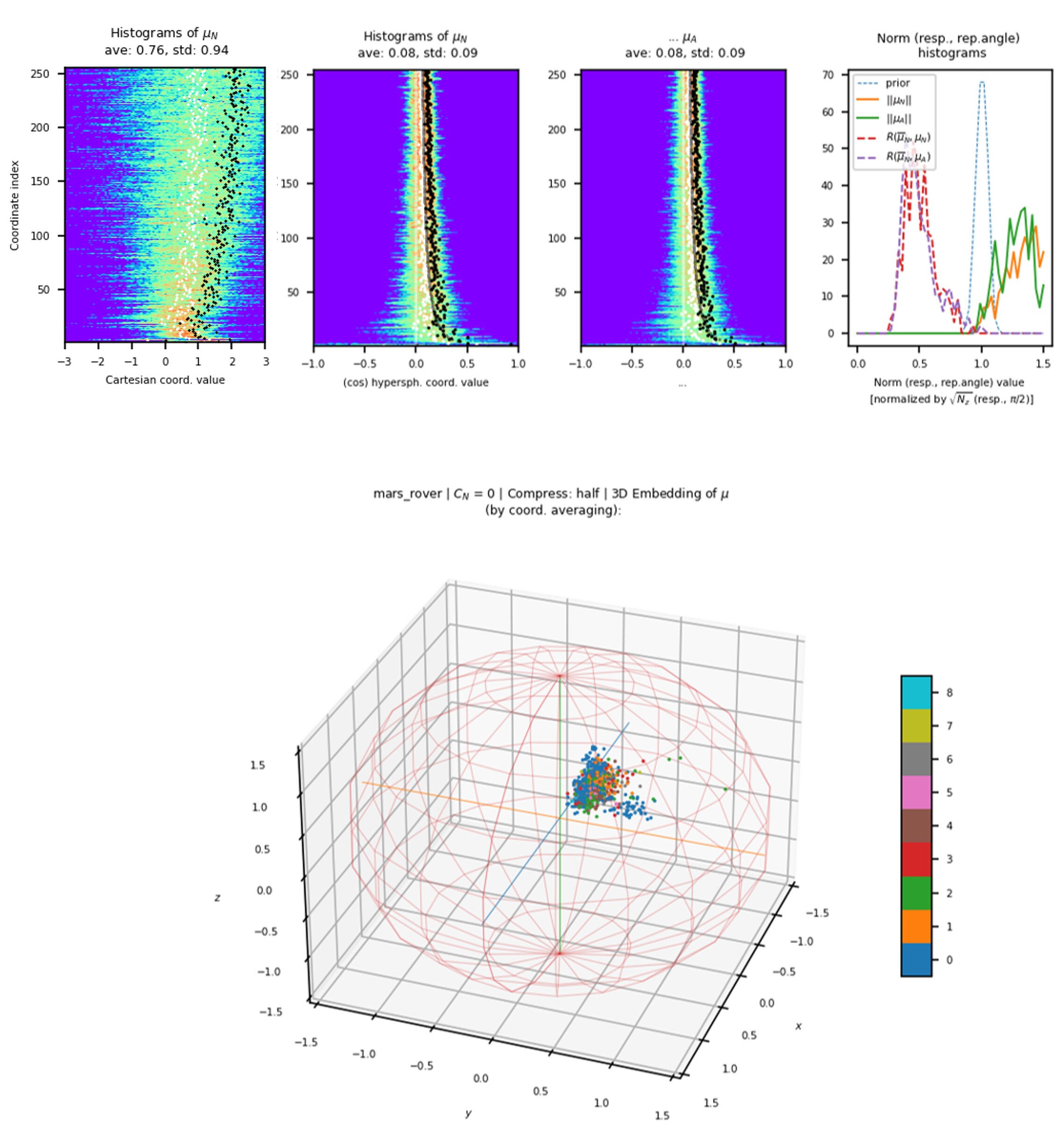}
    \caption{Results of the compressed VAE training, on the test set of the Mars Rover Mastcam dataset. Note how the means for the cosines of the hyperspherical angles are all shifted towards 1.0, but without collapsing the distribution. The compression direction in this case is given by the vector whose Cartesian coordinates are $(1,1,...,1)$ (displayed as faint black dots in the cosine hyperspherical coordinates histograms for reference).}
    \label{fig:13}
\end{figure*}

\begin{figure*}[!h]
    \centering
    \includegraphics[width=1\linewidth]{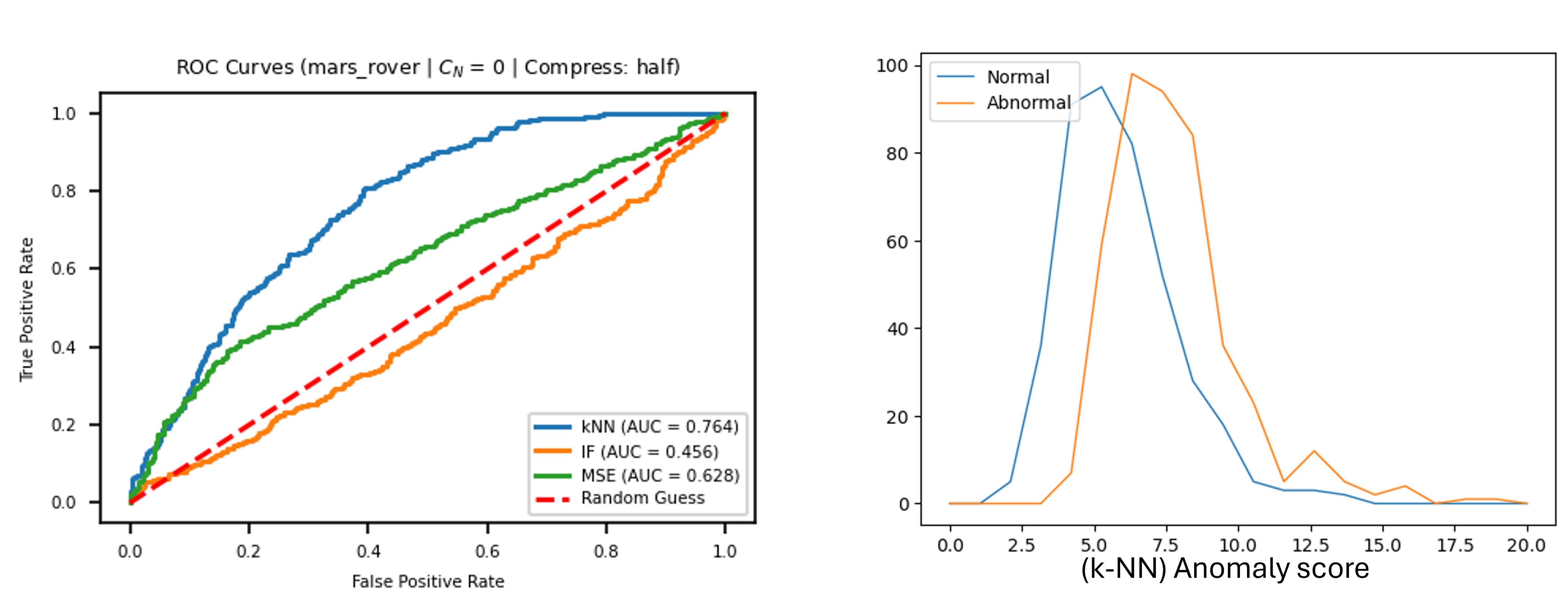}
    \caption{ROC curves and anomaly score histograms, results of the compressed VAE training, on the test set of the Mars Rover Mastcam dataset. Note the compression of the multiple peaks in the normal distribution of Figure \ref{fig:14} into a single one in the compressed version, and how the abnormal class is pushed more to the right.}
    \label{fig:19}
\end{figure*}

\begin{figure*}[!h]
    \centering
    \includegraphics[width=1\linewidth]{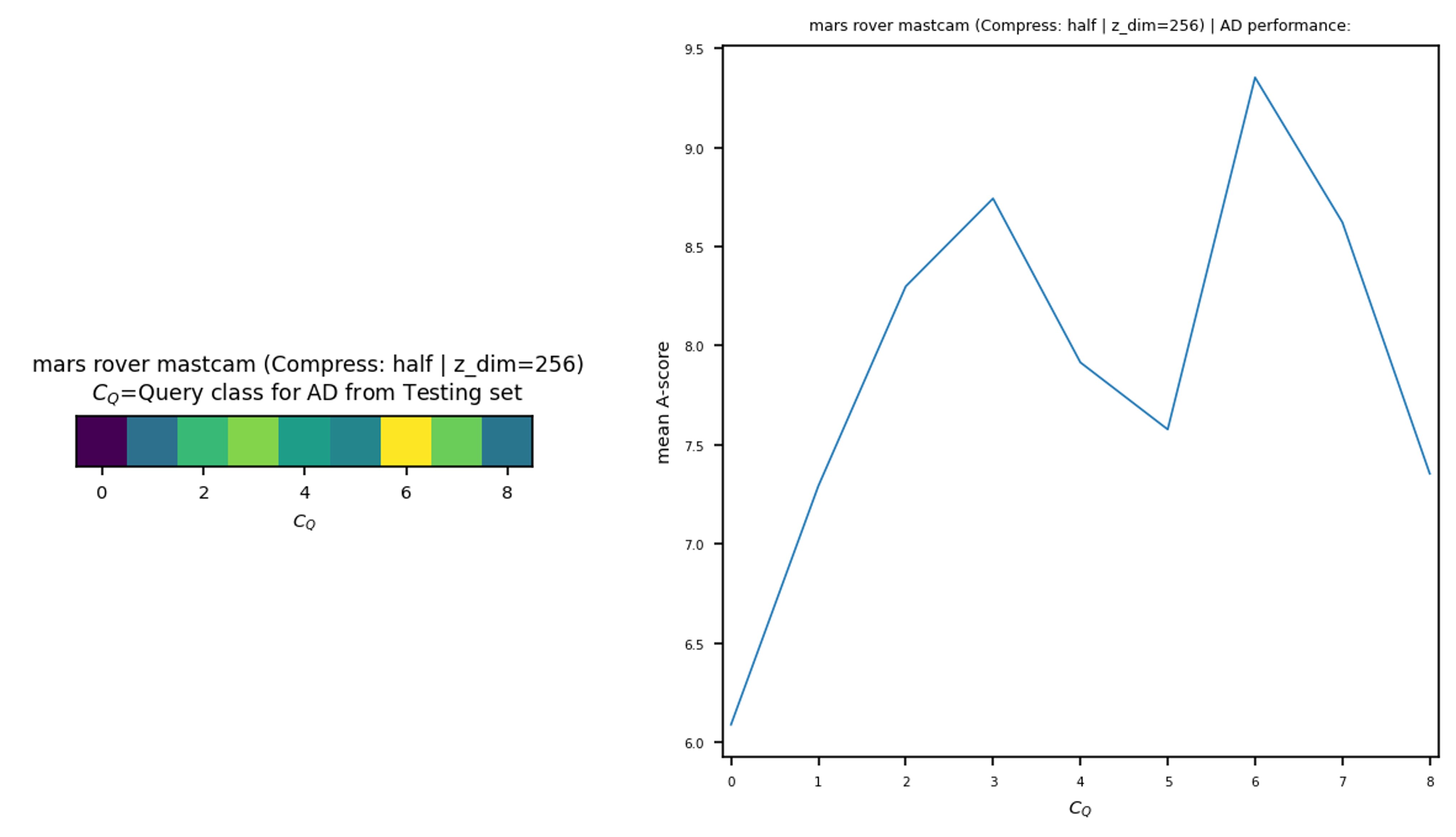}
    \caption{Mean $k-$NN anomaly score values for each class in the test set (0 is the typical/normal class), results of the compressed VAE training, on the test set of the Mars Rover Mastcam dataset.}
    \label{fig:20}
\end{figure*}

\begin{figure*}[!h]
    \centering
    \includegraphics[width=1\linewidth]{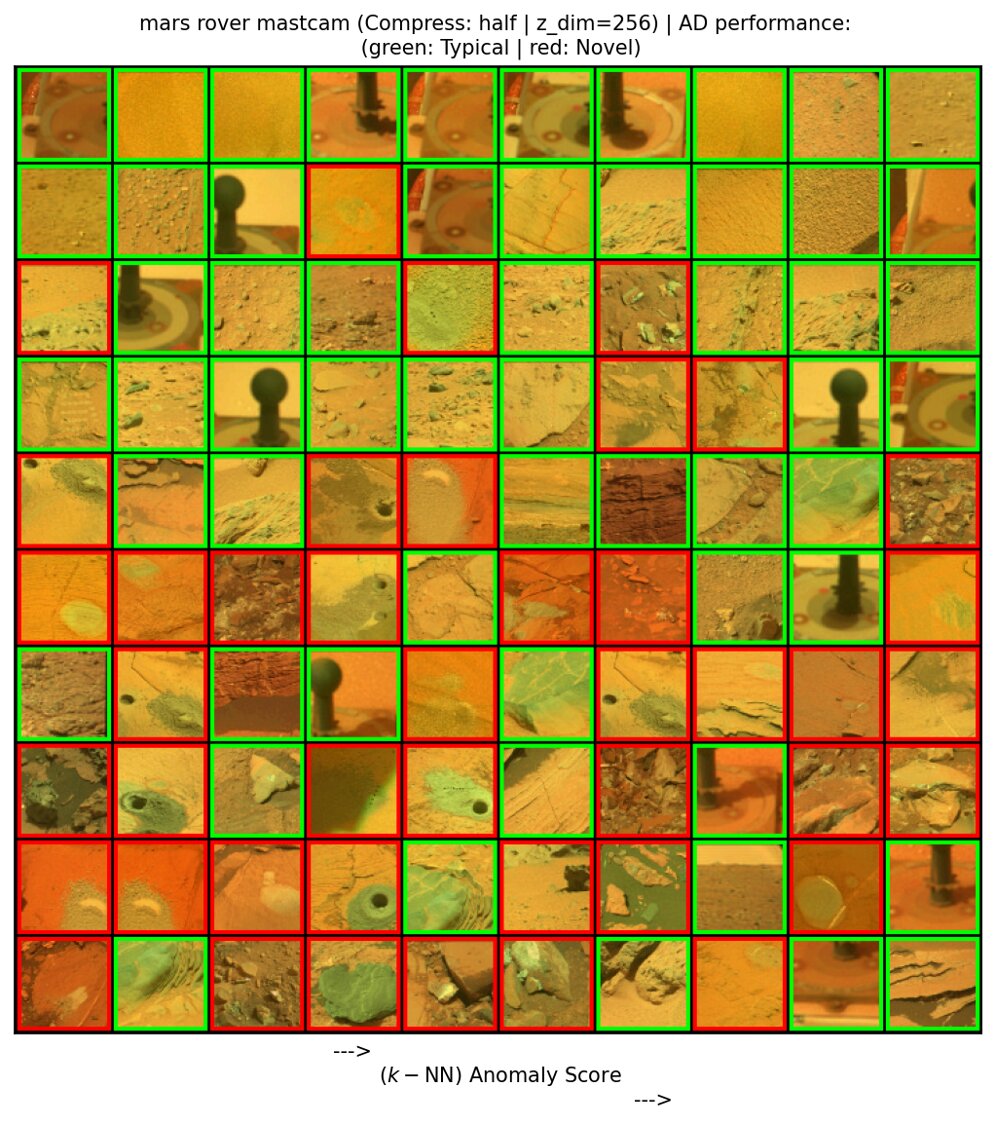}
    \caption{Results of the compressed VAE training, on the test set of the Mars Rover Mastcam dataset. Showing the images from the test set but with the index sorted by the anomaly score (from low at the top to high at the bottom). Only displaying one every 8 images, starting from index 0. There are 420 typical images and 435 novel images in the testing set, for a total of 855. A perfect detection would show the top half of the total panel as normal (green framing) and the anomalies (red framing) at the bottom half.}
    \label{fig:21}
\end{figure*}

\begin{figure*}[!h]
    \centering
    \includegraphics[width=1\linewidth]{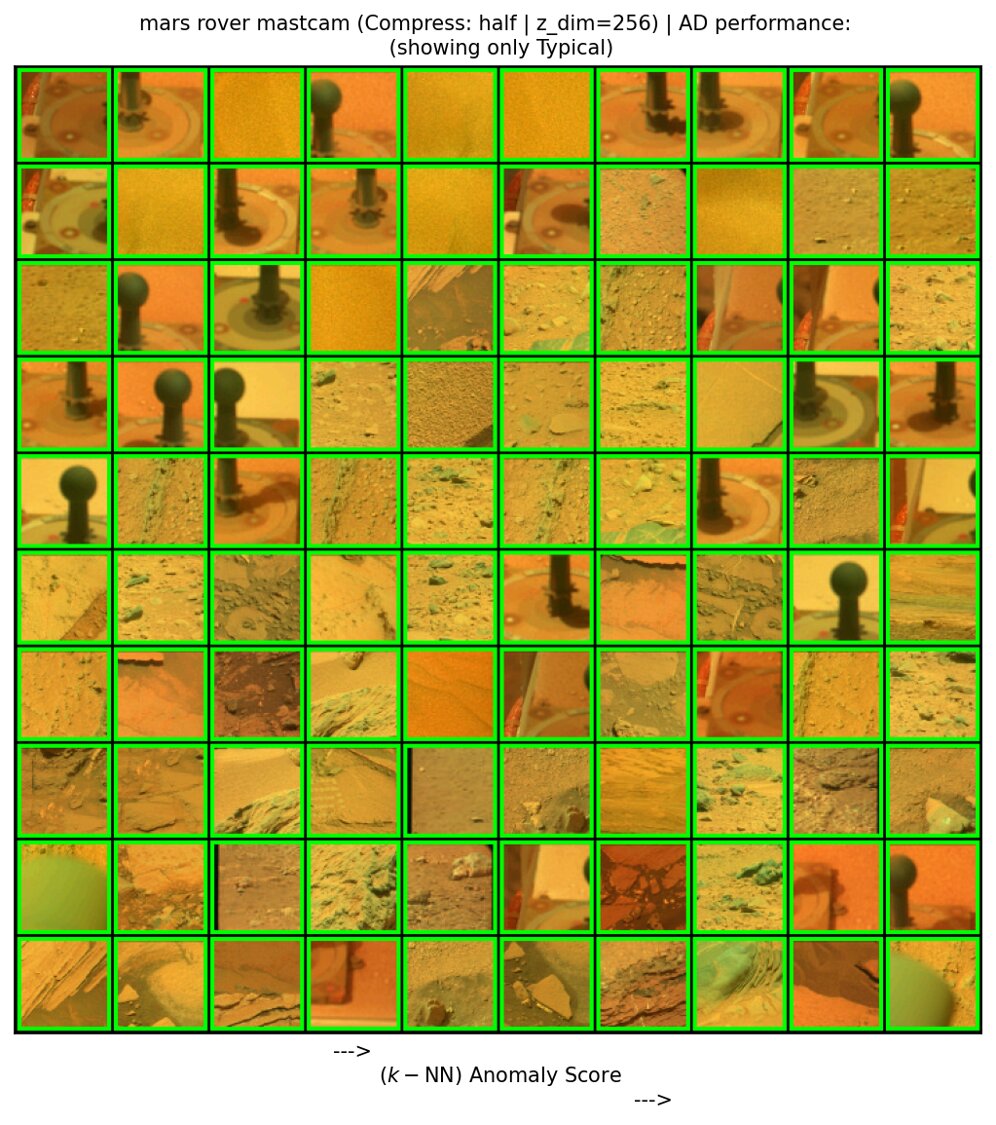}
    \caption{Results of the compressed VAE training, on the test set of the Mars Rover Mastcam dataset. Showing the images from the Typical test set only but with the index sorted by the anomaly score (from low to high). Only displaying one every 4 images, starting from index 0.}
    \label{fig:22}
\end{figure*}

\begin{figure*}[!h]
    \centering
    \includegraphics[width=1\linewidth]{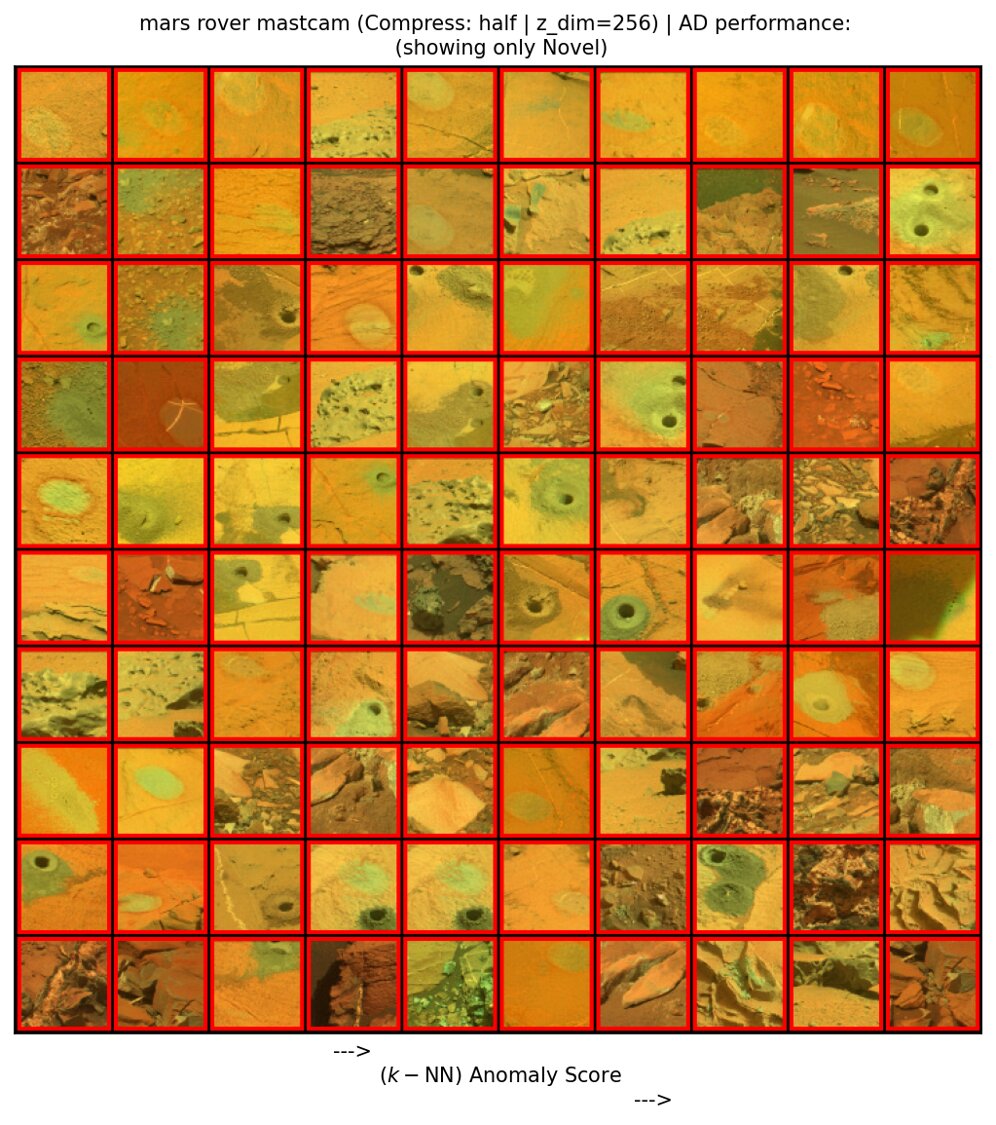}
    \caption{Results of the compressed VAE training, on the test set of the Mars Rover Mastcam dataset. Showing the images from the Novel test set only but with the index sorted by the anomaly score (from low to high). Only displaying one every 4 images, starting from index 0. Note how, in contrast to Figure \ref{fig:18}, the samples with the lowest scores are not exclusively populated by plain textures like the `drt' anomaly type, but also by more rocky textures, thing which makes the `drt' anomaly type more difficult to conflate with similar plain textures in the normal class (and this is indeed reflected in Figure \ref{fig:20}).}
    \label{fig:23}
\end{figure*}

\newpage

\clearpage

\subsubsection{Standard VAE on the Galaxy Zoo dataset}\label{Standard VAE on the Galaxy Zoo dataset}

\begin{figure*}[!h]
    \centering
    \includegraphics[width=1\linewidth]{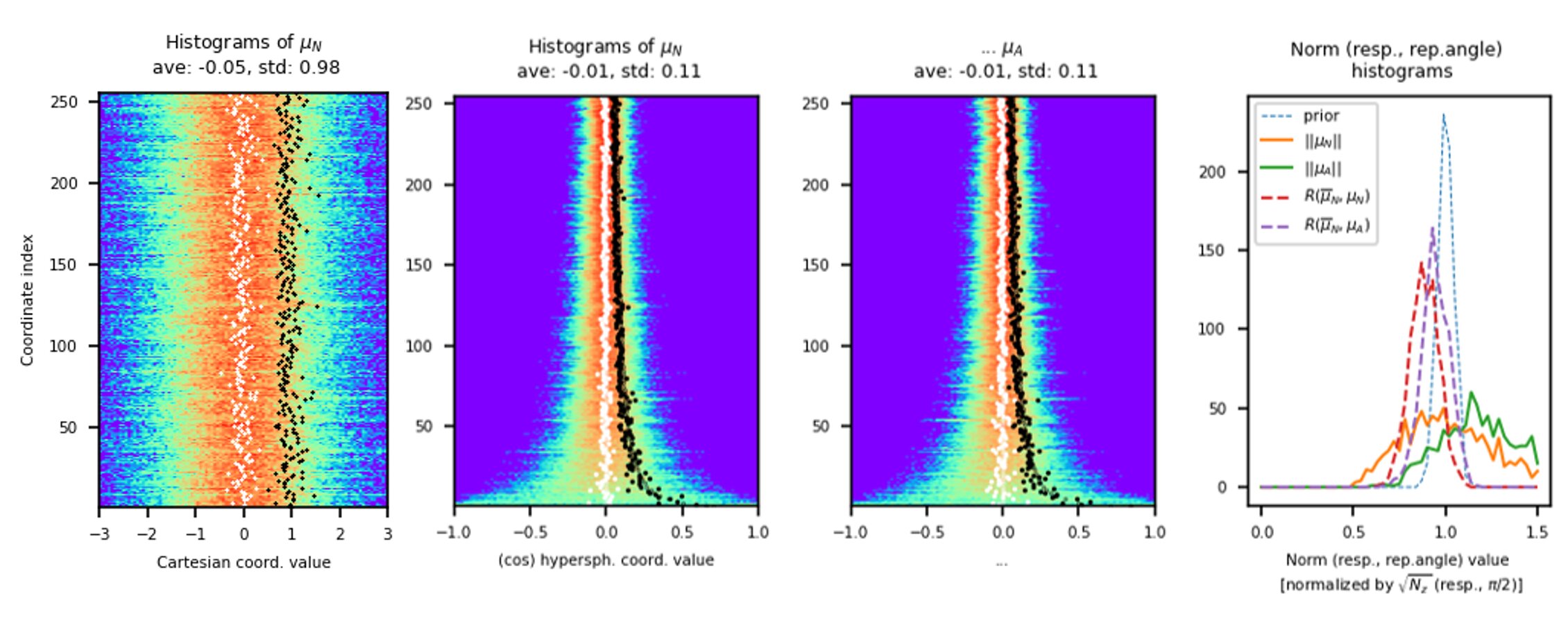}
    \caption{Results of the standard VAE training, on the test set of the Galaxy Zoo dataset.}
    \label{fig:24}
\end{figure*}

\begin{figure*}[!h]
    \centering
    \includegraphics[width=1\linewidth]{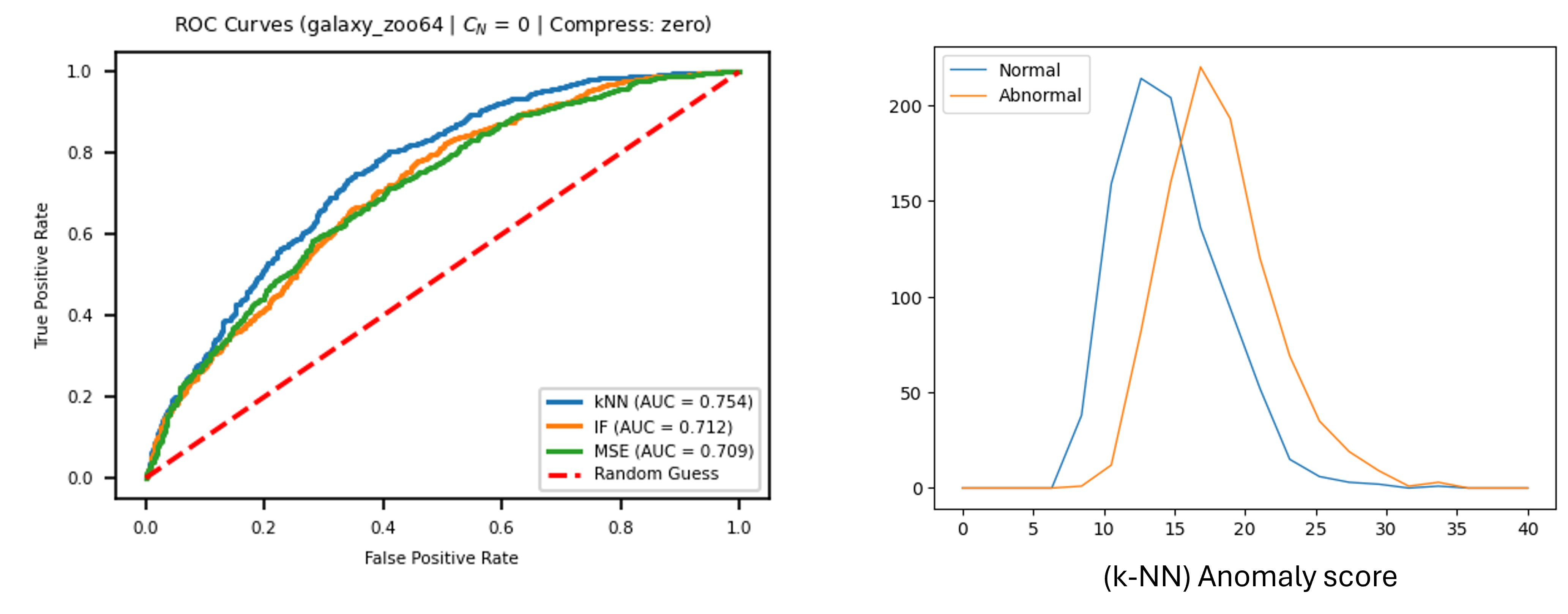}
    \caption{ROC curves and anomaly score histograms, results of the standard VAE training, on the test set of the Galaxy Zoo dataset.}
    \label{fig:26}
\end{figure*}

\begin{figure*}[!h]
    \centering
    \includegraphics[width=1\linewidth]{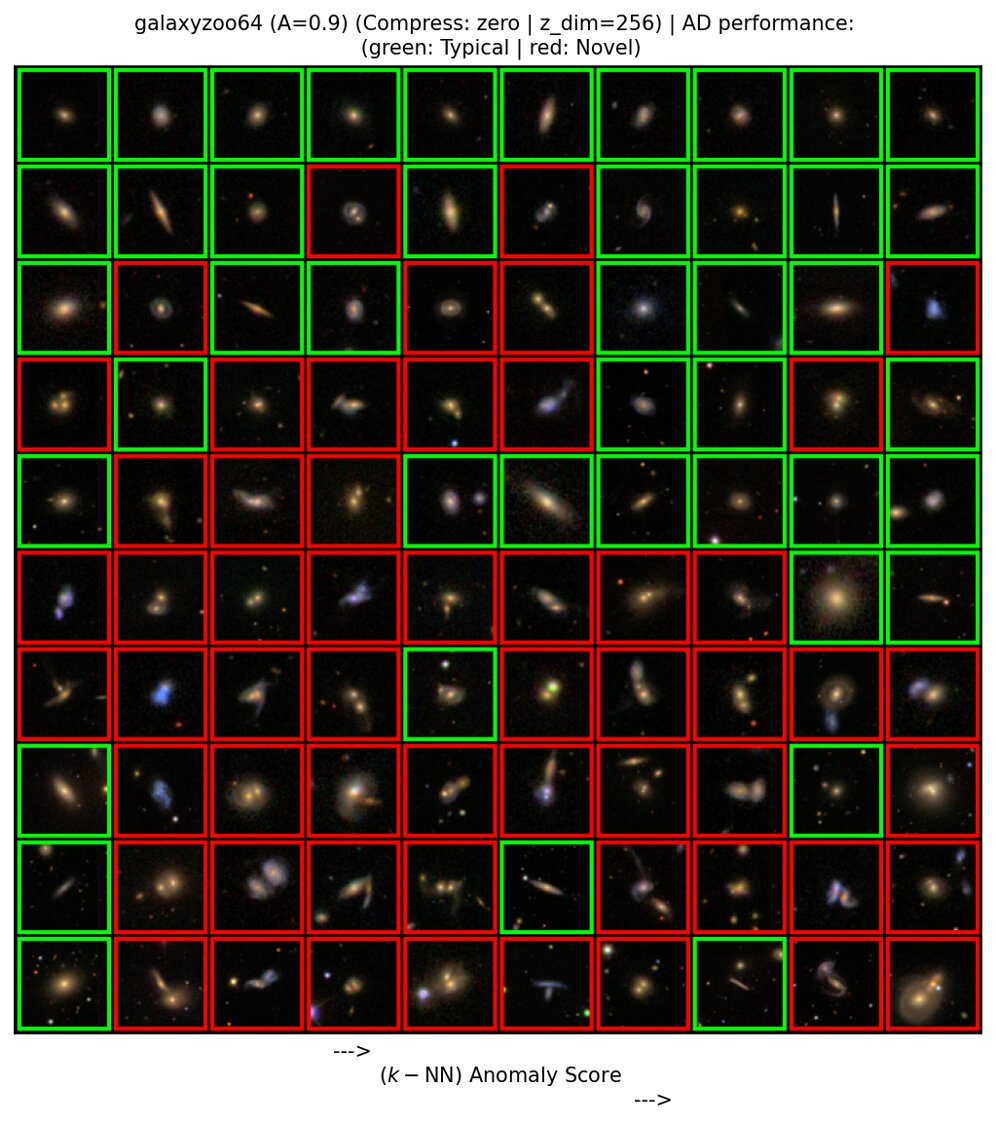}
    \caption{Results of the standard VAE training, on the test set of the Galaxy Zoo dataset. Showing the images from the test set but with the index sorted by the anomaly score (from low at the top to high at the bottom). Only displaying one every 20 images, starting from index 0. There are 1000 typical images and 1000 novel images in the testing set, for a total of 2000. A perfect detection would show the top half of the total panel as normal (green framing) and the anomalies (red framing) at the bottom half.}
    \label{fig:27}
\end{figure*}

\begin{figure*}[!h]
    \centering
    \includegraphics[width=1\linewidth]{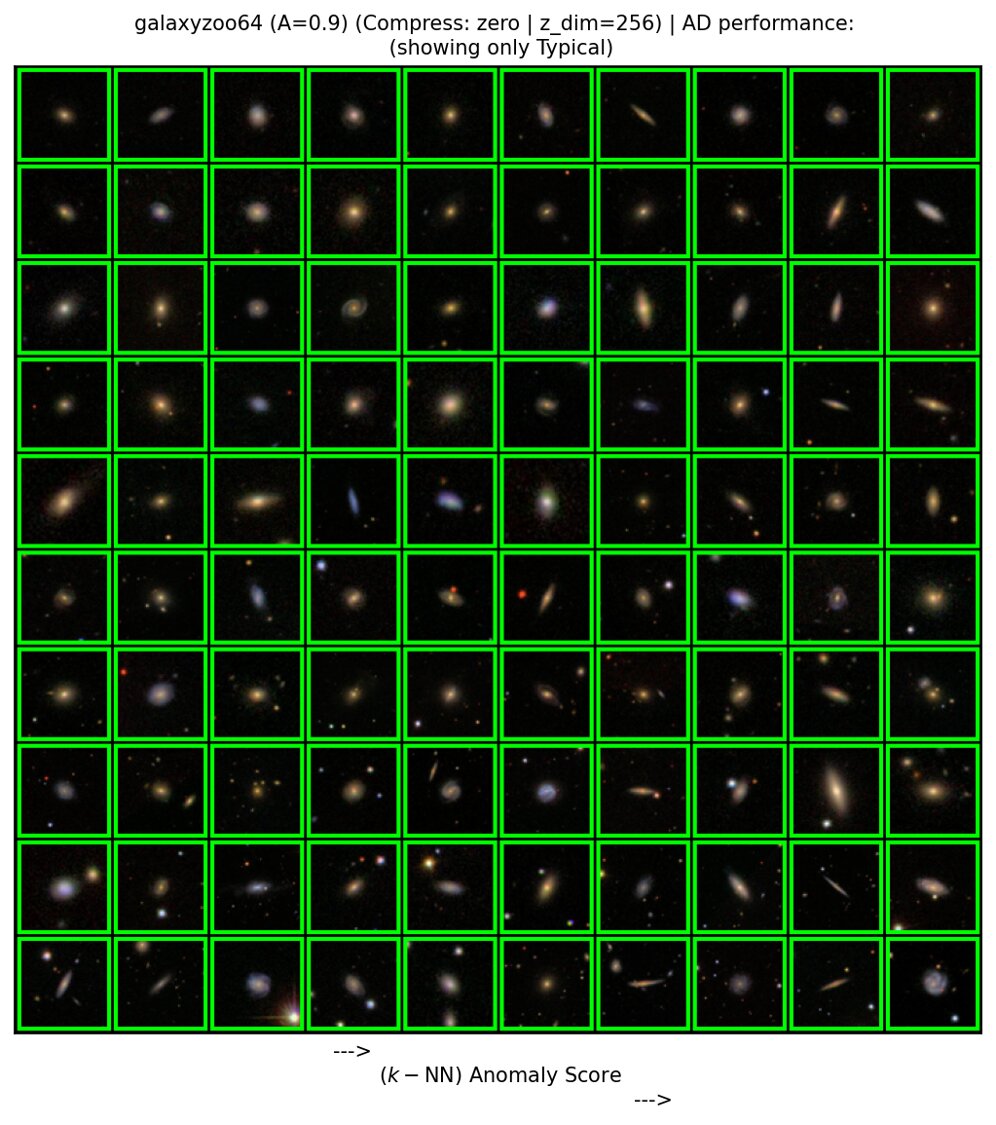}
    \caption{Results of the standard VAE training, on the test set of the Galaxy Zoo dataset. Showing the images from the Typical test set only but with the index sorted by the anomaly score (from low to high). Only displaying one every 10 images, starting from index 0.}
    \label{fig:28}
\end{figure*}

\begin{figure*}[!h]
    \centering
    \includegraphics[width=1\linewidth]{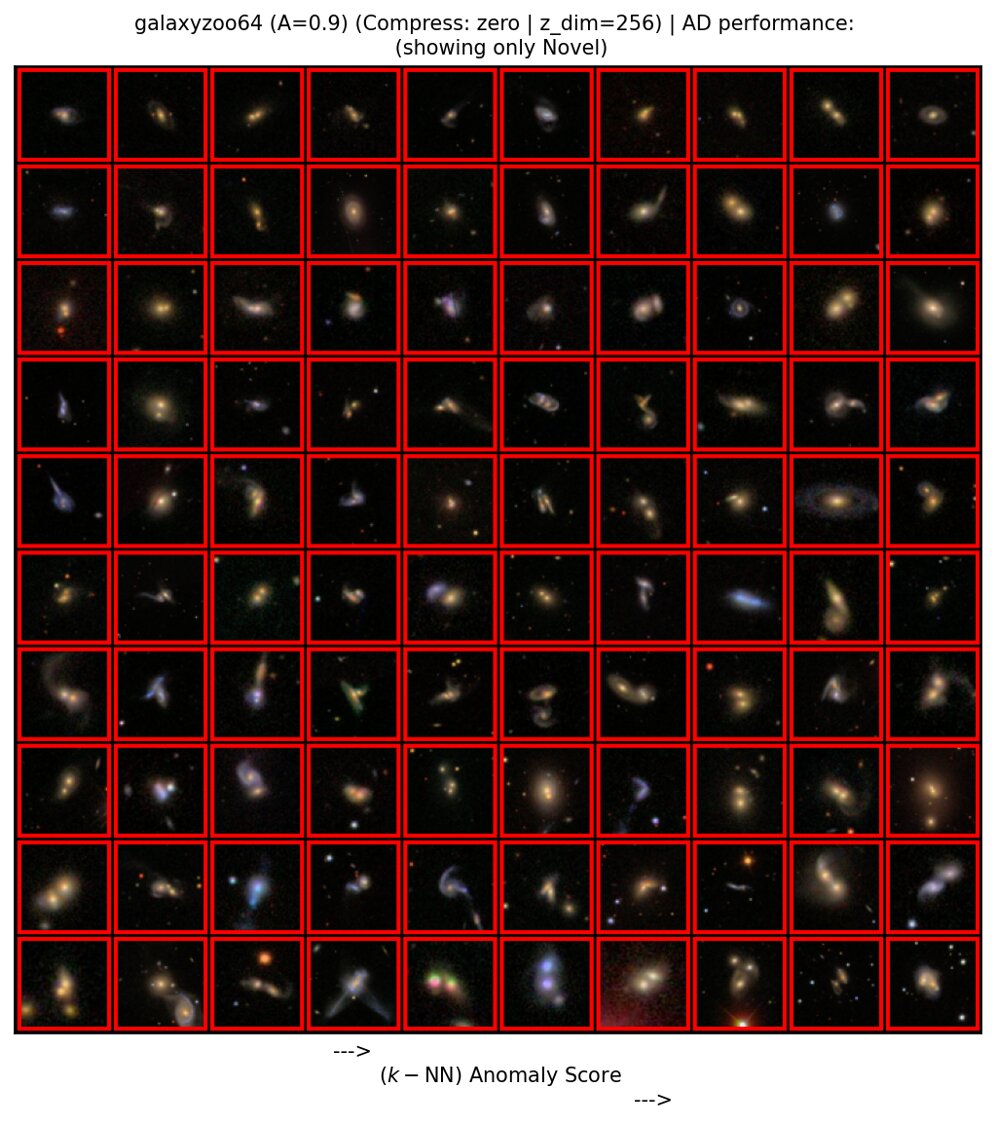}
    \caption{Results of the standard VAE training, on the test set of the Galaxy Zoo dataset. Showing the images from the Novel test set only but with the index sorted by the anomaly score (from low to high). Only displaying one every 10 images, starting from index 0.}
    \label{fig:29}
\end{figure*}

\newpage

\clearpage

\subsubsection{Comp.VAE on the Galaxy Zoo dataset}\label{Comp.VAE on the Galaxy Zoo dataset}

\begin{figure*}[!h]
    \centering
    \includegraphics[width=1\linewidth]{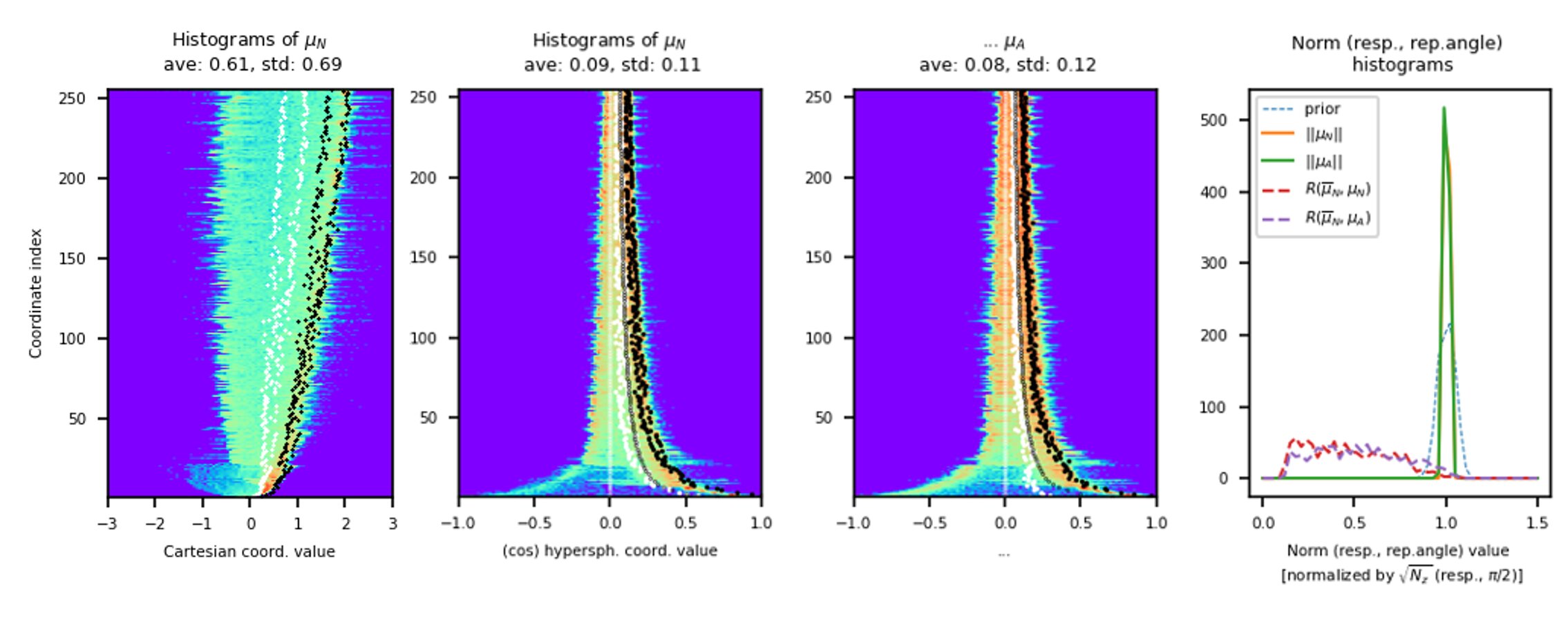}
    \caption{Results of the compressed VAE training, on the test set of the Galaxy Zoo dataset. Note how the means for the cosines of the hyperspherical angles are all shifted towards 1.0, but without collapsing the distribution. The compression direction in this case is given by the vector whose Cartesian coordinates are $(1,1,...,1)$ (displayed as faint black dots in the cosine hyperspherical coordinates histograms for reference).}
    \label{fig:25}
\end{figure*}

\begin{figure*}[!h]
    \centering
    \includegraphics[width=1\linewidth]{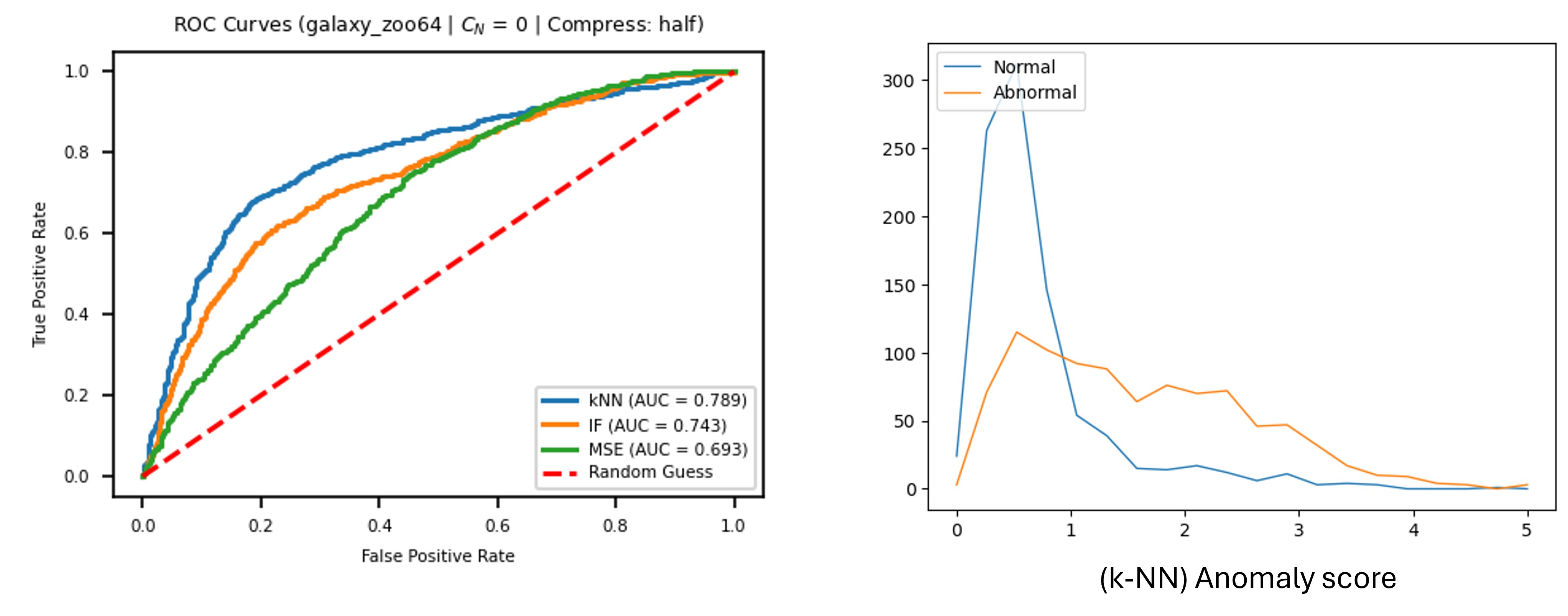}
    \caption{ROC curves and anomaly score histograms, results of the compressed VAE training, on the test set of the Galaxy Zoo dataset.}
    \label{fig:30a}
\end{figure*}

\begin{figure*}[!h]
    \centering
    \includegraphics[width=1\linewidth]{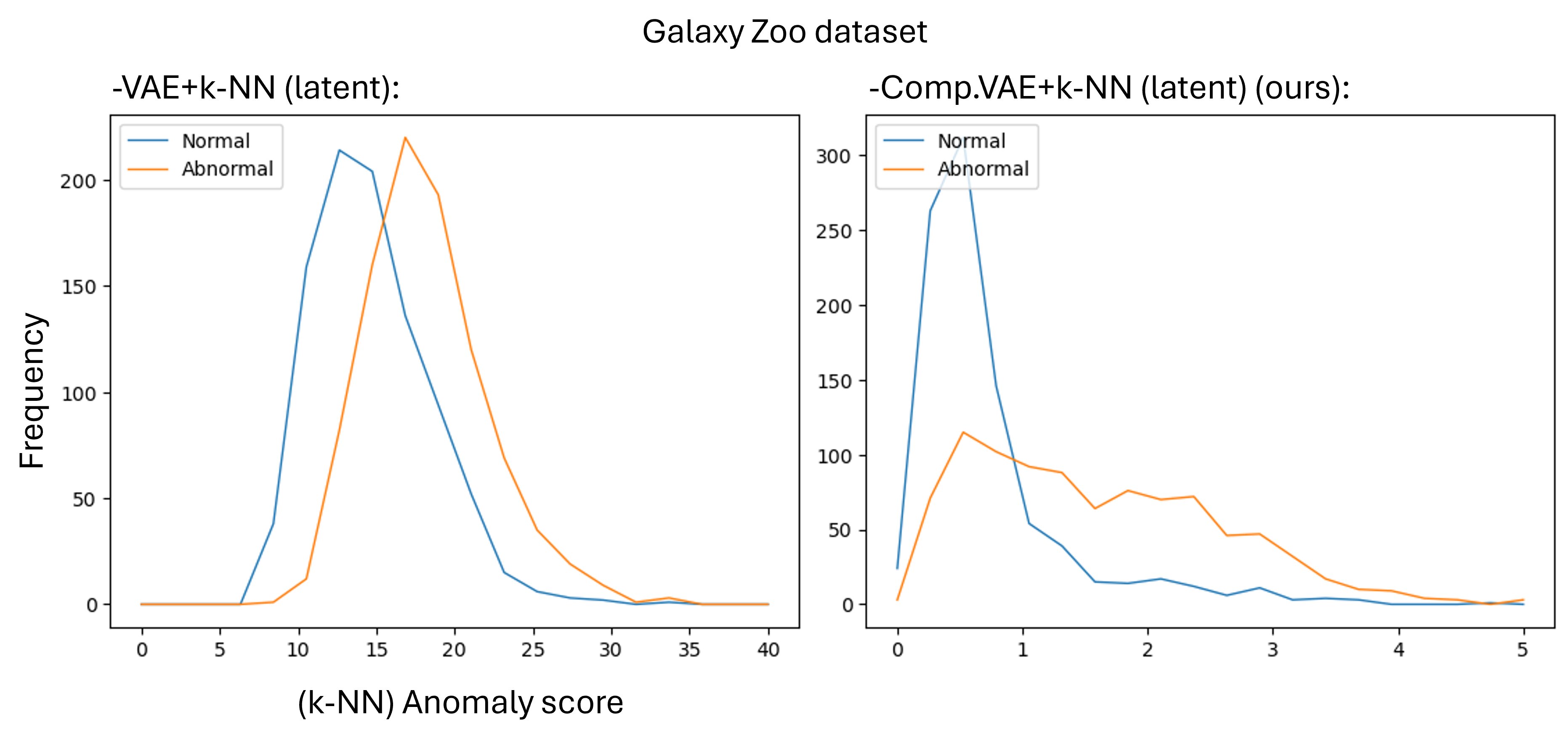}
    \caption{Note the impact of the volume compression of the normal class on the frequency of the $k-$NN anomaly score in the Galaxy Zoo dataset experiment: note the reduced value range in the right figure and the dense and compact population of the values close to 0 by only the normal class, while the anomalies are pushed to an elongated long tail (which can also be seen in Figure 1 from the paper). This results in better AUROC curve values (cf. Table 1, third column, from the paper). Note also the concentration of measure effects on the anomaly score for the standard AE (left figure): the normal class is concentrated around a mode which is far away from zero. There is a basic minimal distance, of around 5, under which no pair of normal samples can be found. This is because the vast volume of the equators, where these samples are located: to fall closer to each other is simply very unlikely, given that there is some much additional space to fall into.}
    \label{fig:30b}
\end{figure*}

\begin{figure*}[!h]
    \centering
    \includegraphics[width=1\linewidth]{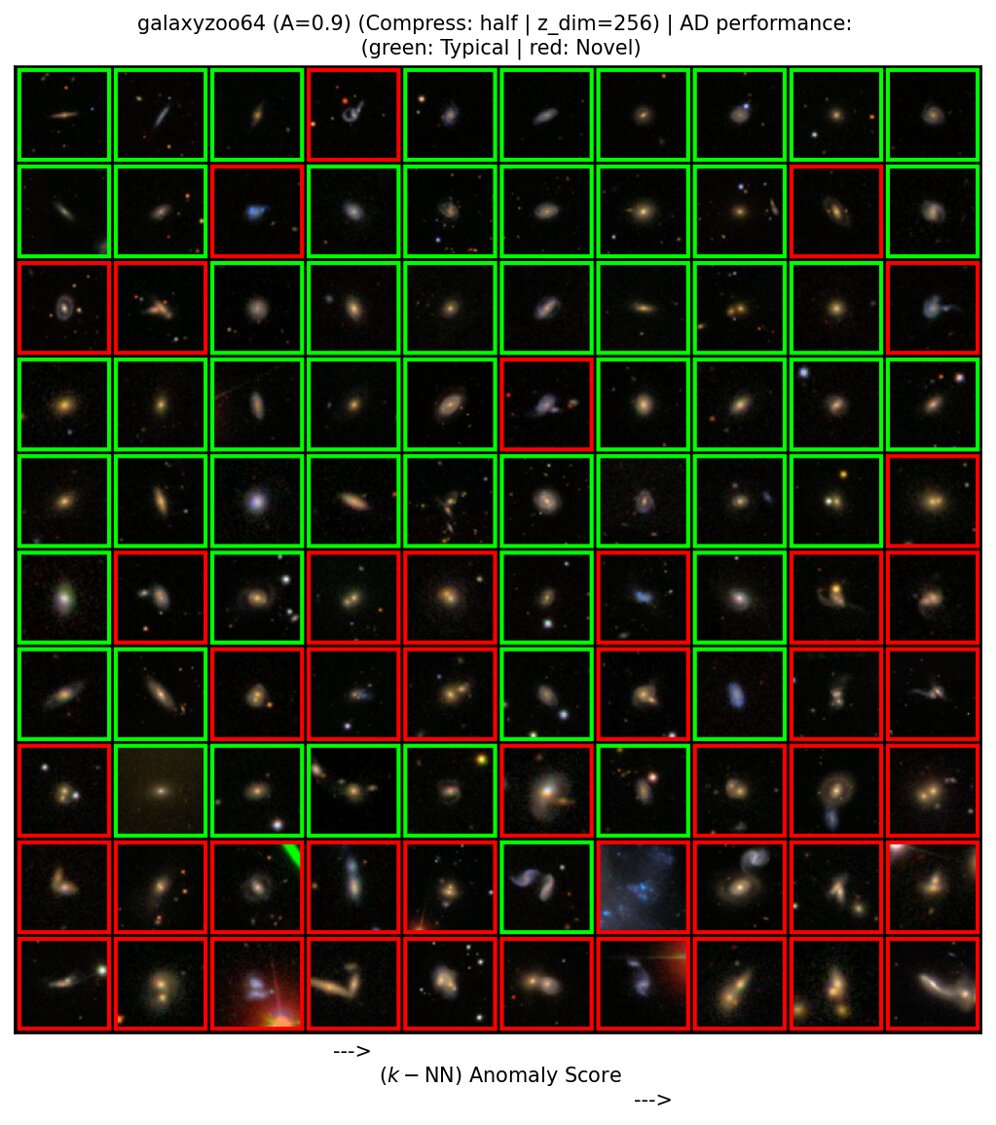}
    \caption{Results of the compressed VAE training, on the test set of the Galaxy Zoo dataset. Showing the images from the test set but with the index sorted by the anomaly score (from low at the top to high at the bottom). Only displaying one every 20 images, starting from index 0. There are 1000 typical images and 1000 novel images in the testing set, for a total of 2000. A perfect detection would show the top half of the total panel as normal (green framing) and the anomalies (red framing) at the bottom half.}
    \label{fig:31}
\end{figure*}

\begin{figure*}[!h]
    \centering
    \includegraphics[width=1\linewidth]{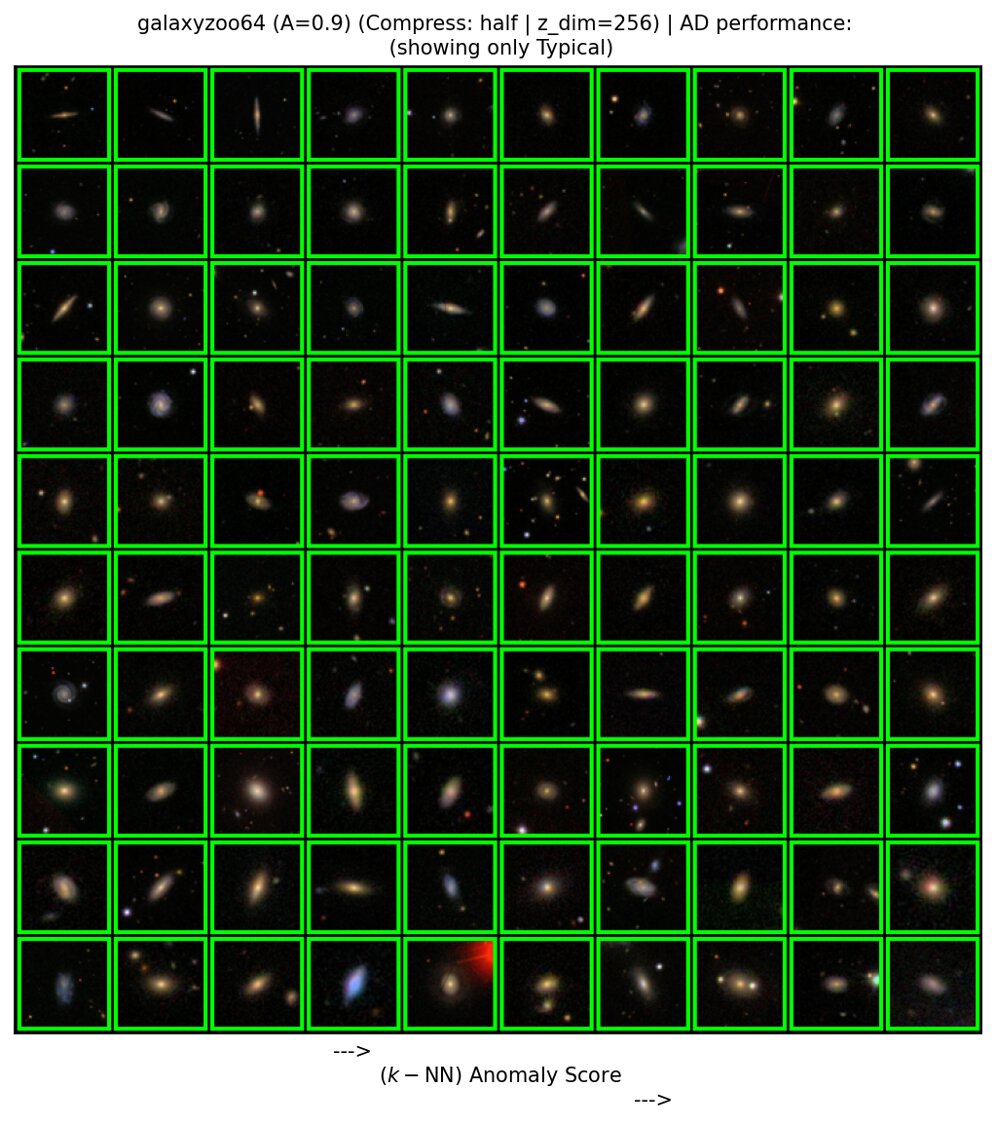}
    \caption{Results of the compressed VAE training, on the test set of the Galaxy Zoo dataset. Showing the images from the Typical test set only but with the index sorted by the anomaly score (from low to high). Only displaying one every 10 images, starting from index 0.}
    \label{fig:32}
\end{figure*}

\begin{figure*}[!h]
    \centering
    \includegraphics[width=1\linewidth]{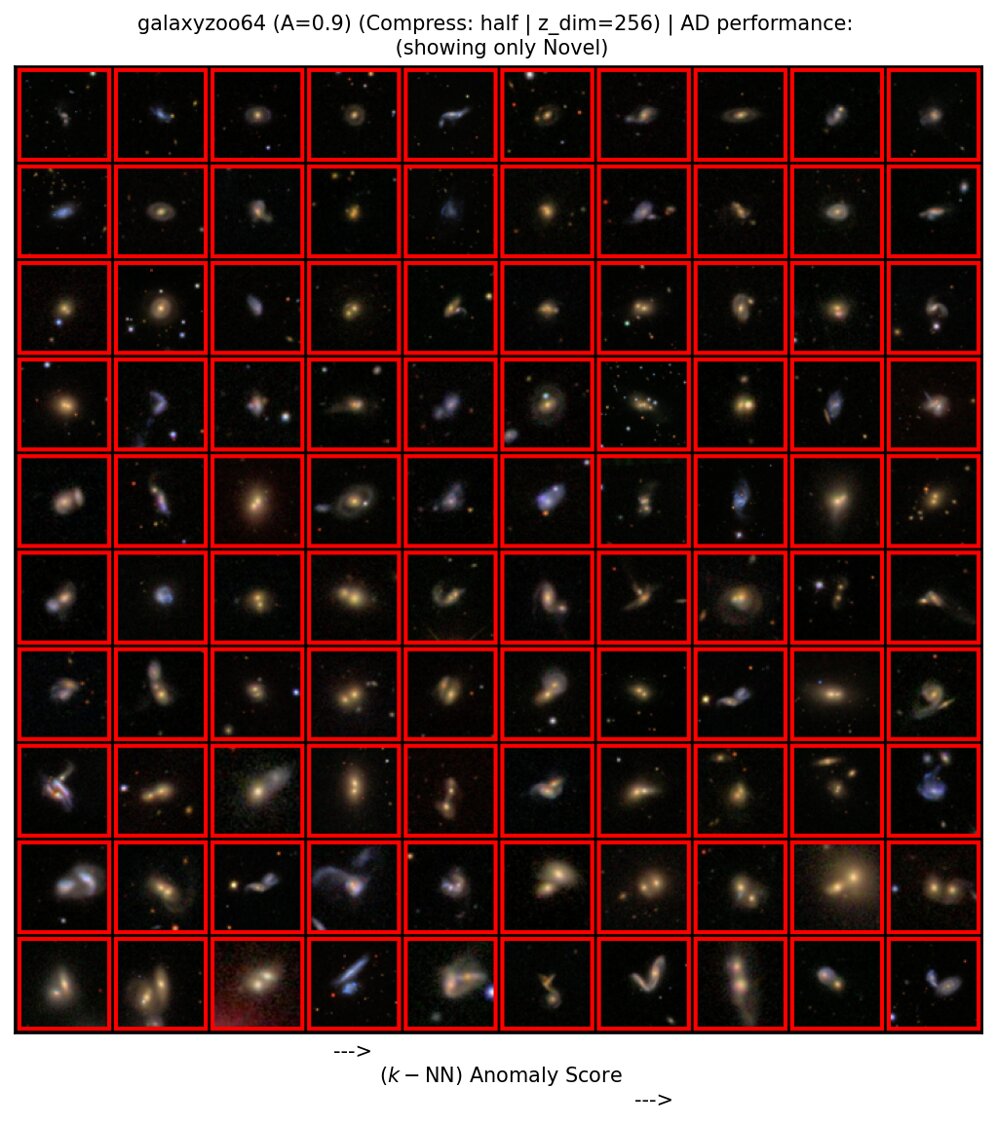}
    \caption{Results of the compressed VAE training, on the test set of the Galaxy Zoo dataset. Showing the images from the Novel test set only but with the index sorted by the anomaly score (from low to high). Only displaying one every 10 images, starting from index 0.}
    \label{fig:33}
\end{figure*}

\newpage

\clearpage

\subsubsection{Comp.VAE (vMF) on the Galaxy Zoo dataset}\label{Comp.VAE (vMF) on the Galaxy Zoo dataset}

\begin{figure*}[!h]
    \centering
    \includegraphics[width=1\linewidth]{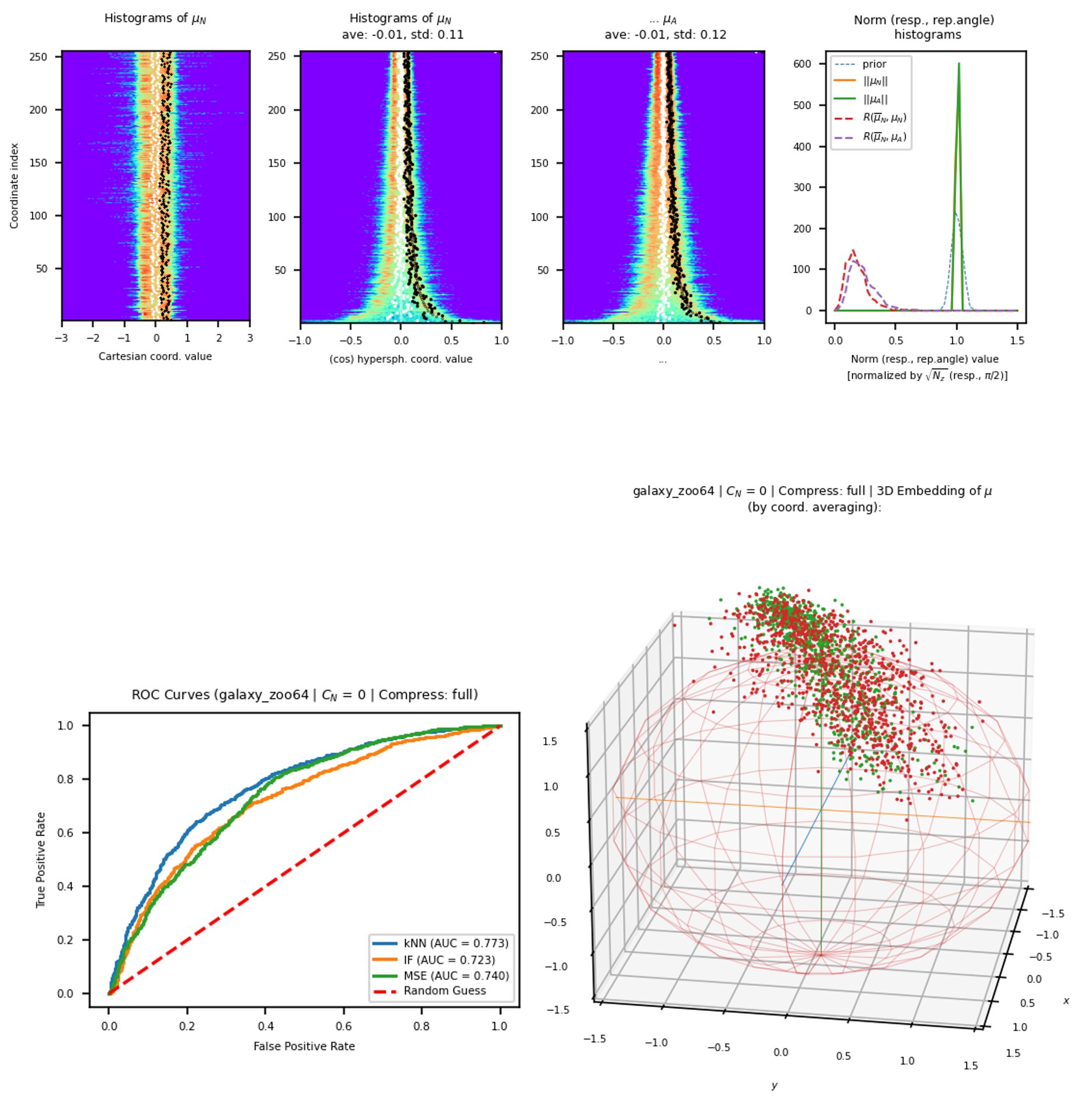}
    \caption{c.f. Figs.\ref{fig:25}, \ref{fig:30a}, and Fig.1 in the main paper.}
    \label{fig:xxy}
\end{figure*}

\newpage

\clearpage

$\boxed{\text{Conditional-OOD experiments}}$

\subsubsection{CIFAR-10 (ID) vs CIFAR-100}\label{conditional-OOD experiments}

\newcommand{\parhead}[1]{%
  \par\smallskip
  \needspace{3\baselineskip}
  \noindent\textbf{#1}\quad
  \nobreak\ignorespaces 
}

\parhead{CIFAR-10 ID Training: Comp.VAE-vMF model}

\begin{figure*}[!h]
    \centering
    \includegraphics[width=1\linewidth]{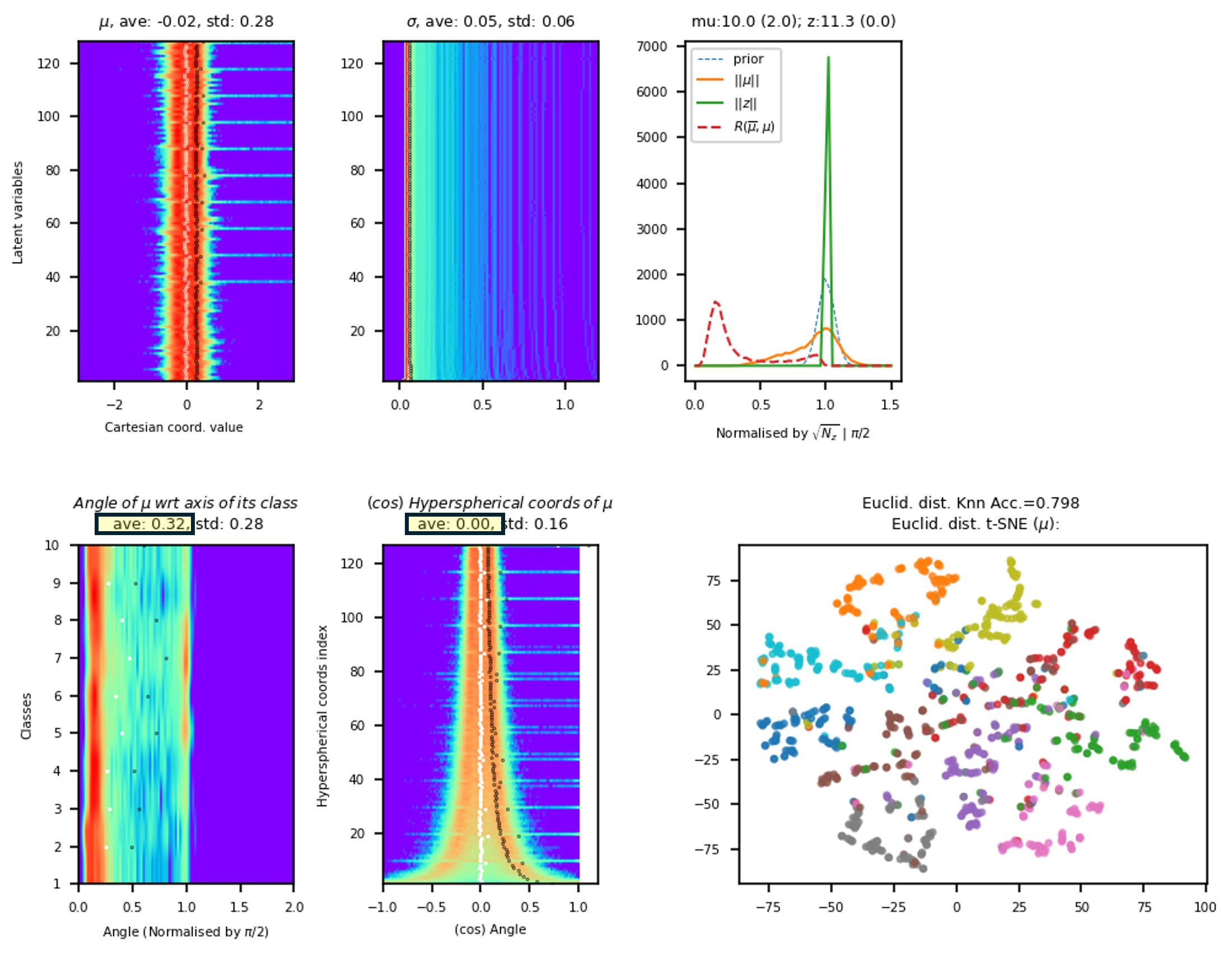}
    \caption{Upper panel, Cartesian coordinates histograms of $\mu$, $\sigma$, and the norm of $\mu$ and replica angle. Bottom panel, histograms for the first hyperspherical coordinate w.r.t. to the Cartesian axis corresponding to each sample's label, then the rest of the hyperspherical coordinates, and a t-SNE of the ID testing set latent embeddings. We highlight the average of the first angle (in the vMF approach, only this angle is compressed) and the average of the remaining hyperspherical coordinates (in our approach, all angles are compressed).}
    \label{fig:33b}
\end{figure*}
\clearpage

\parhead{CIFAR-10 ID Training: Comp.VAE-full compression model}

\begin{figure*}[!h]
    \centering
    \includegraphics[width=1\linewidth]{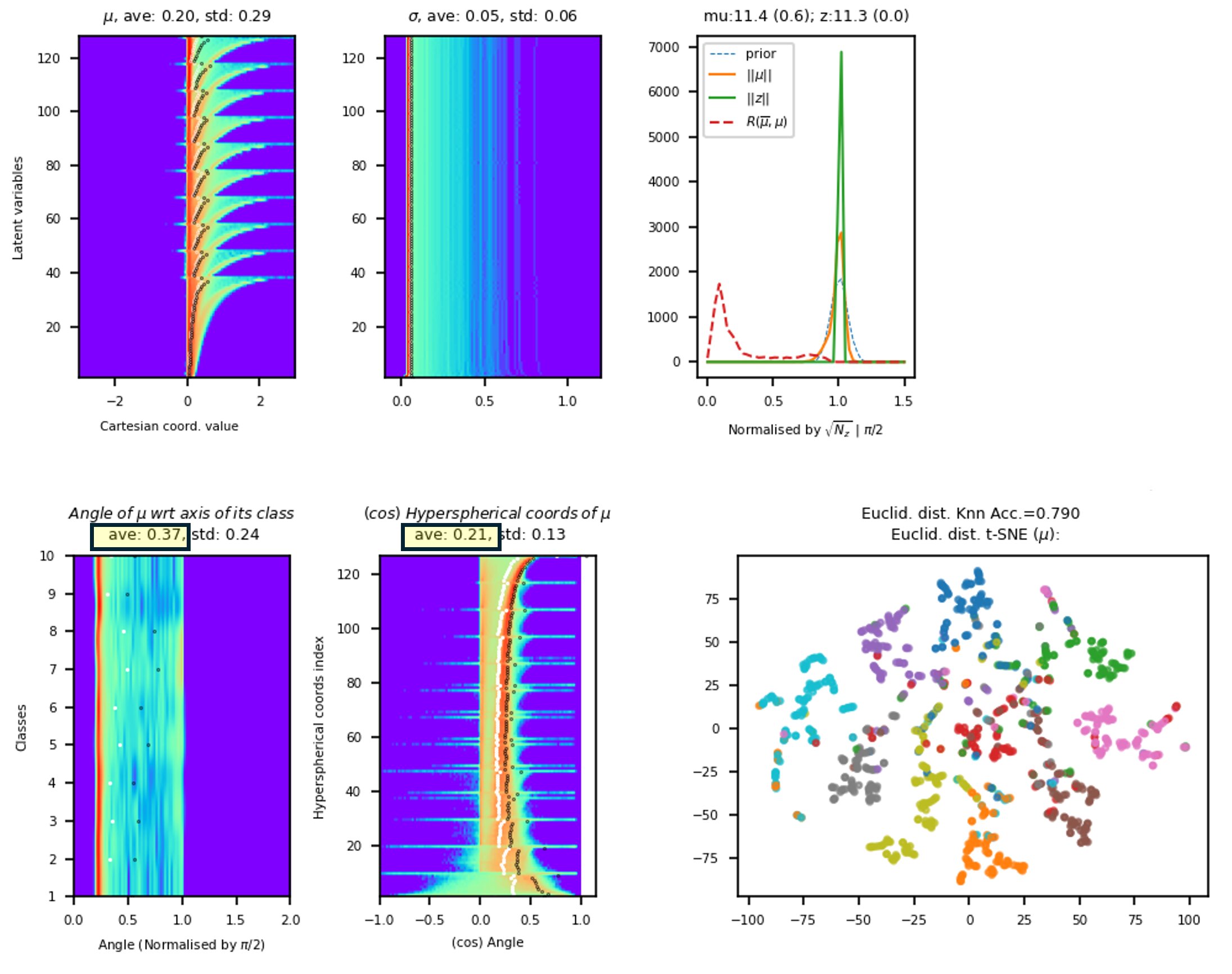}
    \caption{c.f. previous figure.}
    \label{fig:33c}
\end{figure*}
\clearpage

\parhead{CIFAR-100 near-conditional-OOD: comparison}

\begin{figure*}[!h]
    \centering
    \includegraphics[width=1\linewidth]{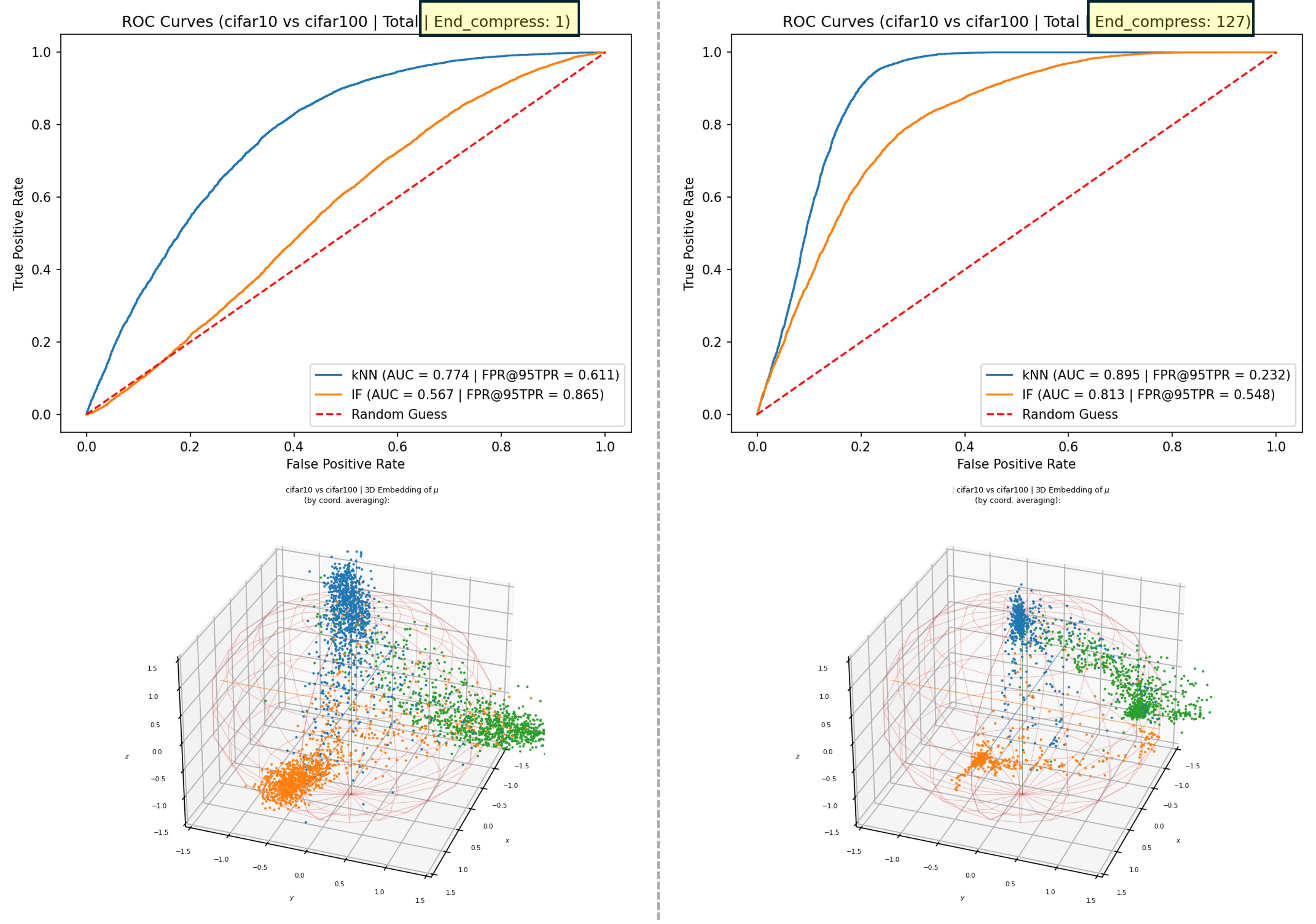}
    \caption{ROC curves and 3d visualizations. Left: Comp.VAE-vMF. Right: Comp.VAE full compression.}
    \label{fig:33d}
\end{figure*}

\newpage

\clearpage

\subsubsection{Imagenette (ID) vs close ImageNet classes}\label{Imagenette ID Training}

\parhead{Imagenette ID Training: Comp.VAE-vMF model}

\begin{figure*}[!h]
    \centering
    \includegraphics[width=1\linewidth]{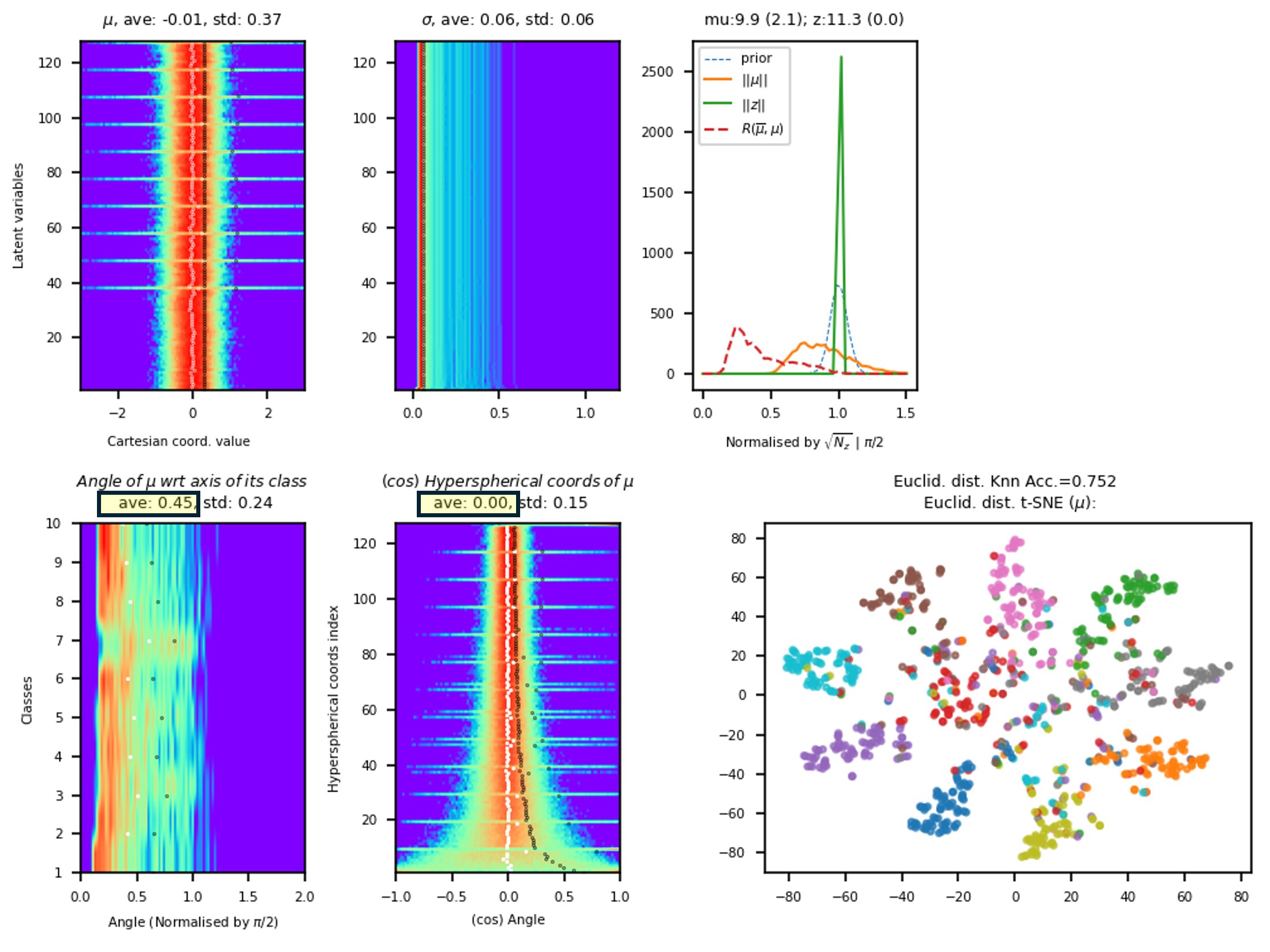}
    \caption{Upper panel, Cartesian coordinates histograms of $\mu$, $\sigma$, and the norm of $\mu$ and replica angle. Bottom panel, histograms for the first hyperspherical coordinate w.r.t. to the Cartesian axis corresponding to each sample's label, then the rest of the hyperspherical coordinates, and a t-SNE of the ID testing set latent embeddings. We highlight the average of the first angle (in the vMF approach, only this angle is compressed) and the average of the remaining hyperspherical coordinates (in our approach, all angles are compressed).}
    \label{fig:33e}
\end{figure*}
\clearpage

\parhead{Imagenette ID Training: Comp.VAE-full compression model}

\begin{figure*}[!h]
    \centering
    \includegraphics[width=1\linewidth]{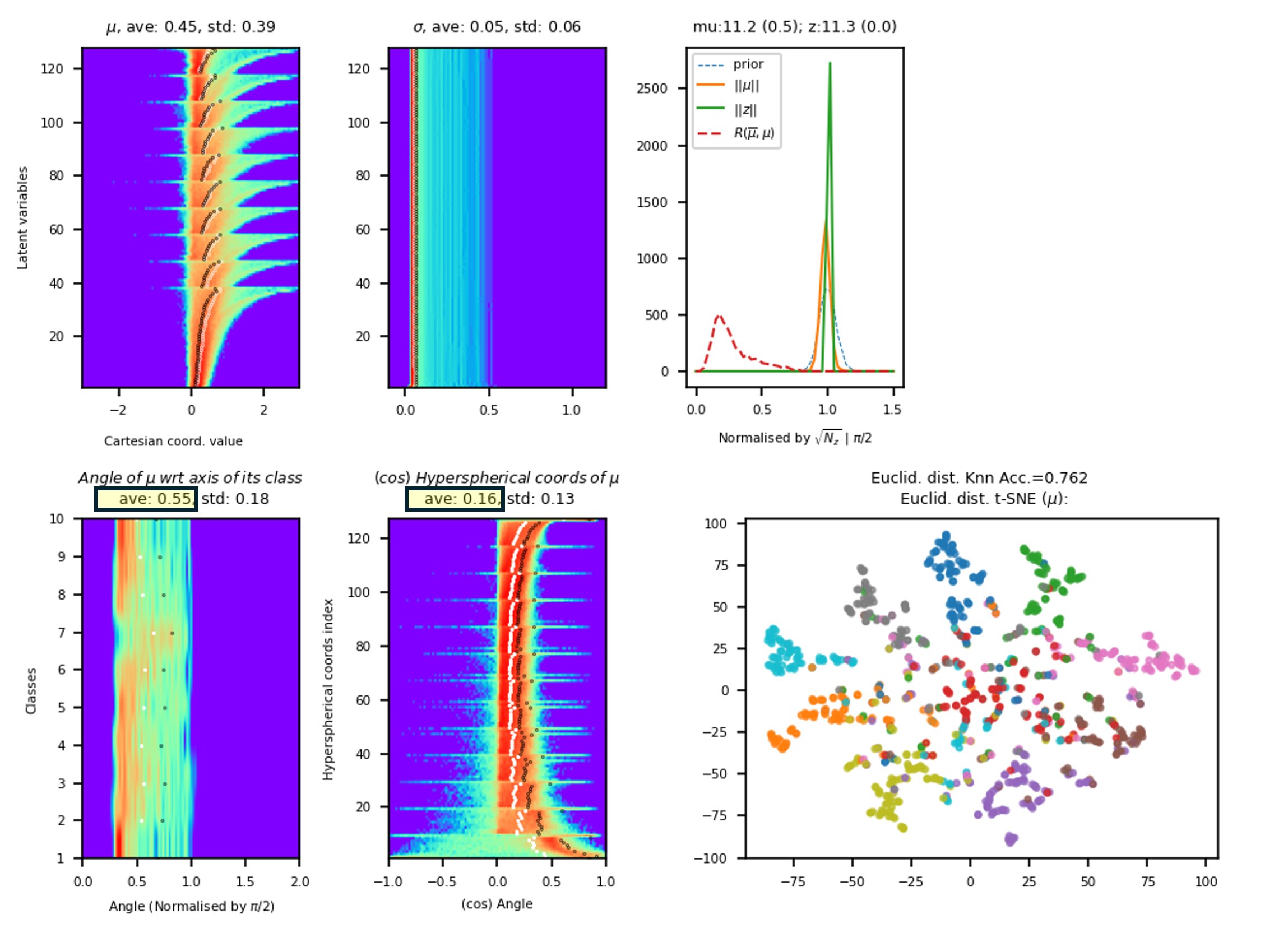}
    \caption{c.f. previous figure.}
    \label{fig:33f}
\end{figure*}
\clearpage

\parhead{close ImageNet near-conditional-OOD: comparison}

\begin{figure*}[!h]
    \centering
    \includegraphics[width=1\linewidth]{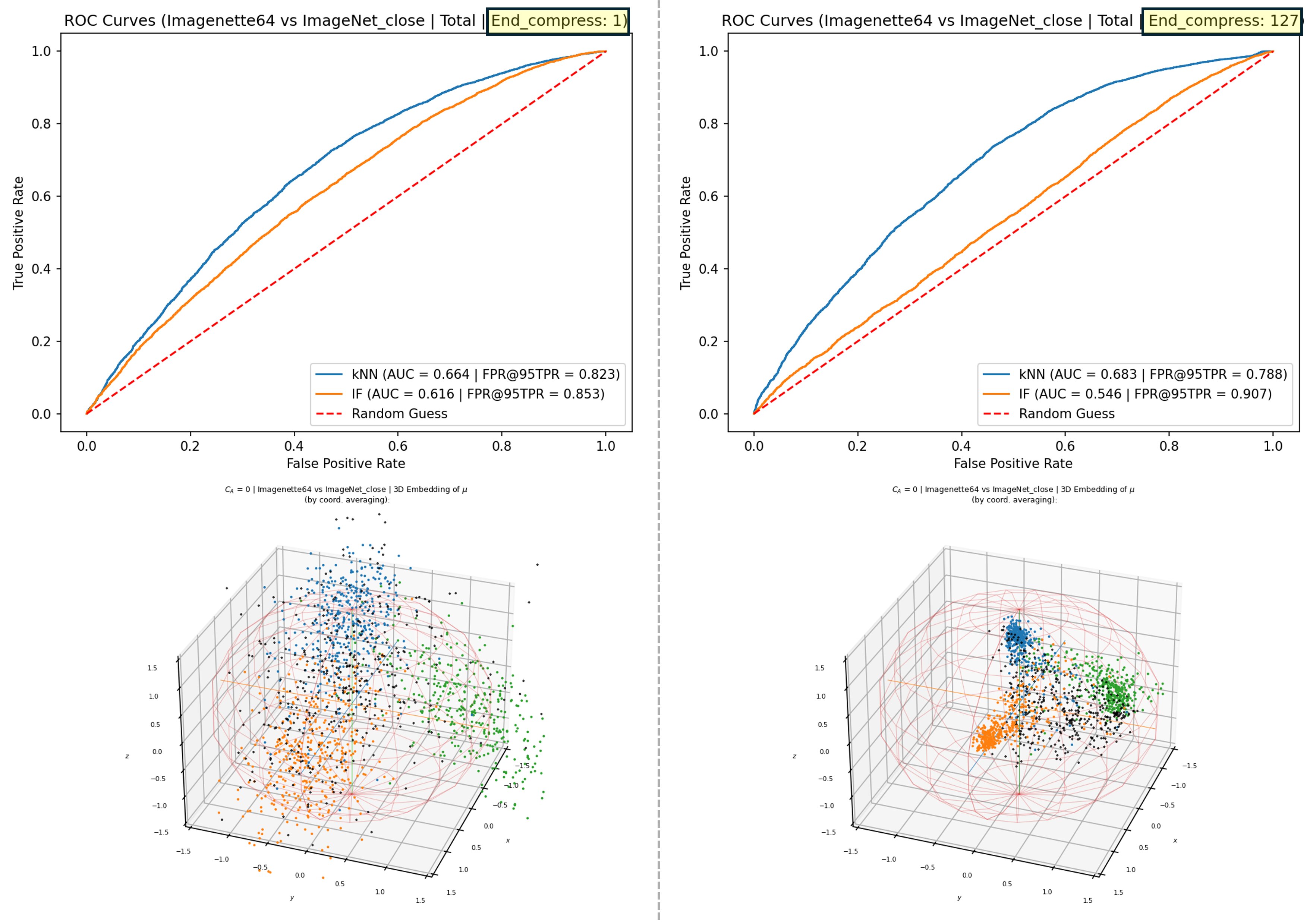}
    \caption{ROC curves and 3d visualizations. Left: Comp.VAE-vMF. Right: Comp.VAE full compression.}
    \label{fig:33g}
\end{figure*}

\newpage

\clearpage

\twocolumn

\subsection{Results on CIFAR10 9-to-1 fully unconditional-OOD AD}\label{appendix:cifar10}

Another issue we encountered in our investigation on AD is the use of simulated anomalies to measure performance in the existing literature \cite{Golan2018DeepTransformations,Han2022ADBench:Benchmark,You2022ADetection}. A typical experiment consists of using standard classification benchmark datasets (e.g., CIFAR10 or SVHN), and using one as the training normal dataset while another is the anomaly to be detected. Most methods perform very well (almost perfect detection) because, together with the supervision for the ID sub-classes (conditional-OOD type of AD), the abnormal samples are very different from the normal ones. An alternative is to use only one dataset, but label some classes as normal, while the remaining classes are the anomaly. This suffers from the opposite problem: in the fully unconditional-OOD case there is no difference between normal and abnormal, and often many normal samples are more abnormal than abnormal classes. We provide evidence of these issues below, and advise against such experimental methods in the fully unconditional-OOD case (the conditional-OOD case performs well in the literature, cf. previous references). 

When one class is used as normal (1-to-9), many of the datasets in the AD literature that we reviewed are too small (e.g., CIFAR10 has 5000 training samples per class) to properly train a model. Correcting the prevalence by using 9 classes as normal (9-to-1) results, in the fully unconditional-OOD case, in that the abnormal data distribution overlaps with the normal, giving very low performance (AUROC curve value $\sim50\%$ in average across all the classes at best). In fact, when looking at the performance per class, the AUC is often below 50\%, revealing an inversion of the classification. In other words, the abnormal class becomes closer to the normal one than most of the normal samples, probably due to the complex topology and geometry of the class manifold, likely related to the randomly changing backgrounds in most of the images, which is not the case in true anomaly detection, when the normal class can be somewhat complex, but should be predictable. Once this is in place, one could study the effect of randomly changing backgrounds, but that is an addition, a posteriori task to basic AD.

We observed this phenomenon whether the anomaly scores are computed from the original data or their latent (via an AE or VAE), in all the methods that we tested. 

Here we include the results of our 9-to-1 AD experiments on pixel-space for CIFAR10 (that is, 9 classes are selected as normal while the remaining one is taken as the source of anomalies during testing). Note that what we performed are the fully unconditional-OOD version of these experiments, that is, the subclass information for the normal distribution (composed of 9 subclasses, as mentioned) is not used at any point, contrary to many experiments in the literature which do take this information into account, thing which is used to disentangle the normal subclasses with a classifier (e.g., a ResNet) and then the obtained `good' embeddings are used to perform the AD, this greatly simplifies the problem.

As can be seen in the figures below, the results can be quite inconsistent when varying the anomalous class, since, for some of them, one can observe pathological behaviors, like the inversion of the ROC curves (which means that the testing anomalous class is closer to the training normal than the testing normal class itself, as the histograms for the anomaly scores confirm).

\onecolumn

\begin{figure*}[!h]
    \centering
    \includegraphics[width=1\linewidth]{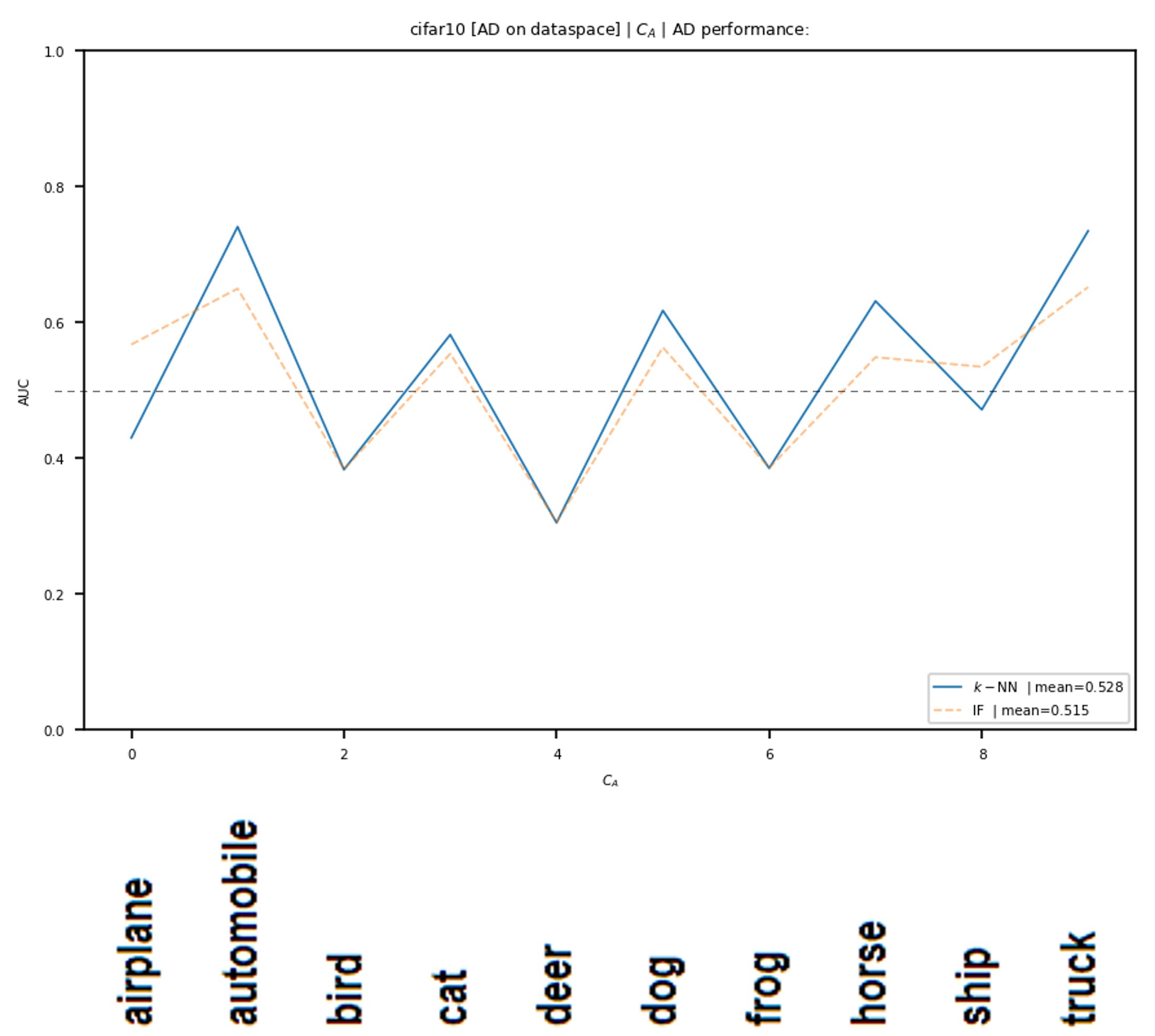}
    \caption{AUROC curve values for 9-to-1 AD experiments on pixel-space for CIFAR10. $C_A$ in the horizontal axis indicates the class being taken as anomalous. The grey horizontal dashed line corresponds to the AUC value of 0.50.}
    \label{fig:34}
\end{figure*}

\begin{figure*}[!h]
    \centering
    \includegraphics[width=1\linewidth]{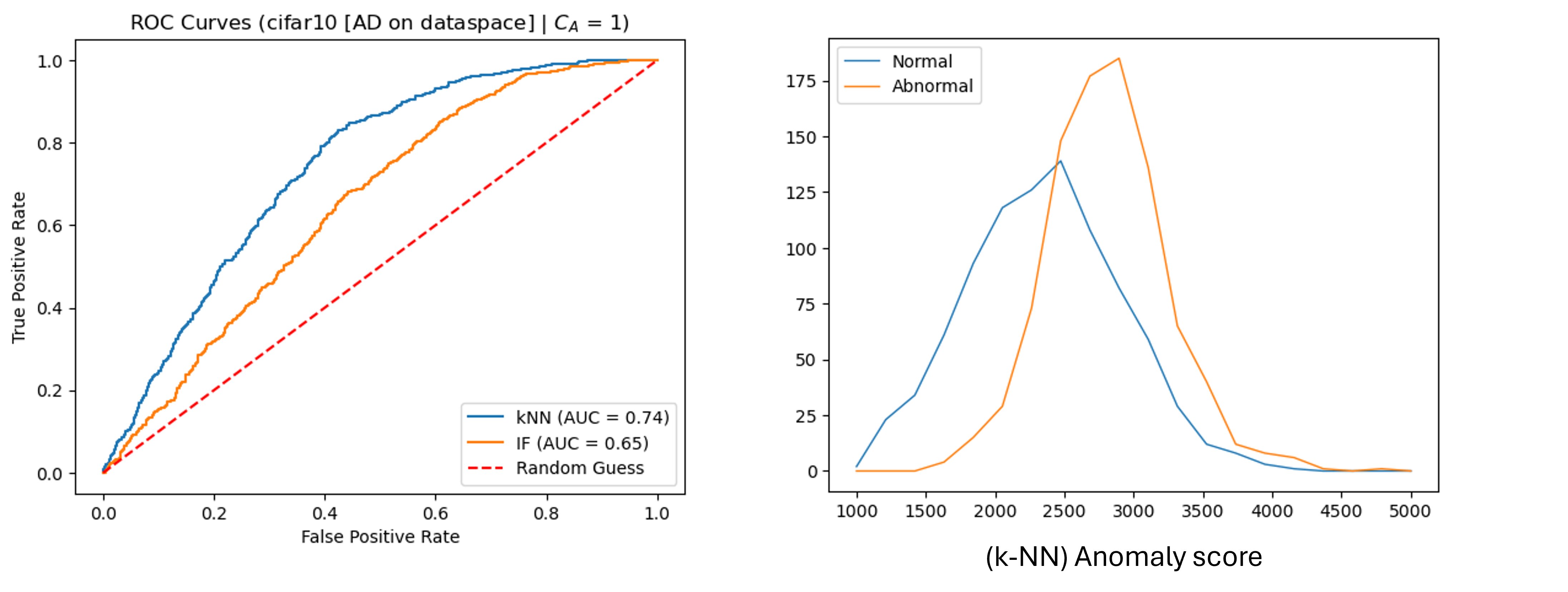}
    \caption{ROC curves and anomaly score histograms, results for the $C_A=automobile$ experiment.}
    \label{fig:35}
\end{figure*}

\begin{figure*}[!h]
    \centering
    \includegraphics[width=1\linewidth]{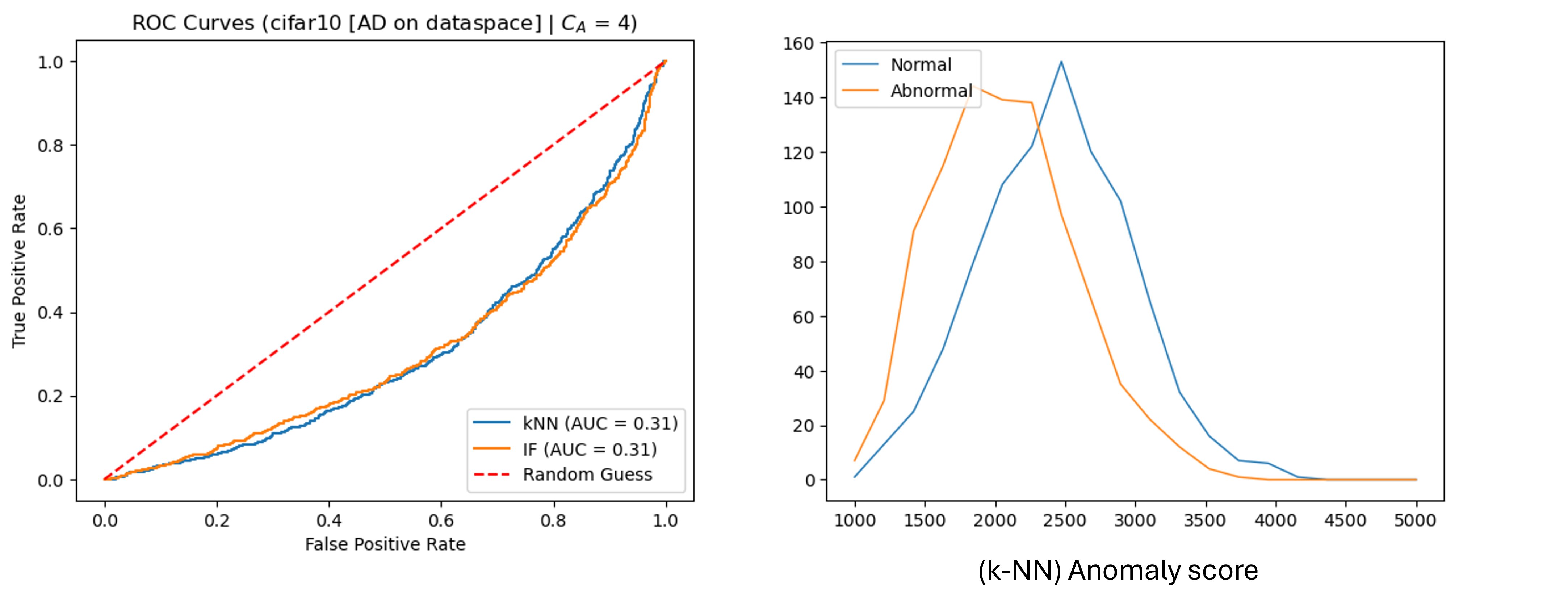}
    \caption{ROC curves and anomaly score histograms, results for the $C_A=deer$ experiment.}
    \label{fig:36}
\end{figure*}

\begin{figure*}[!h]
    \centering
    \includegraphics[width=0.5\linewidth]{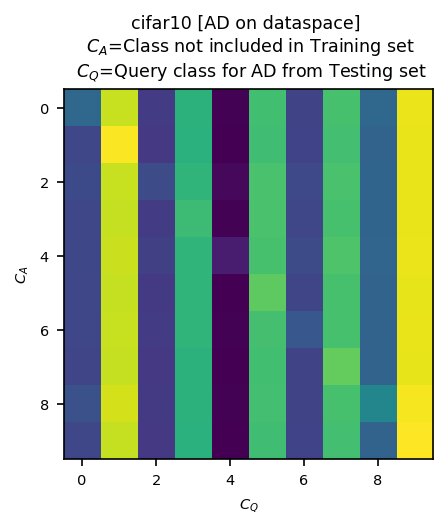}
    \caption{Matrix with all the k-NN anomaly score values (blue means low, yellow is high). The horizontal direction is one single 9-to-1 experiment. In a better behaved case, one would expect the diagonal to be yellow while the rest of the matrix to be blue.}
    \label{fig:37}
\end{figure*}

\newpage

\clearpage

\twocolumn

Even if one employs a new model or method capable of re-inverting scores from baseline methods to recover standard AUC behavior in those test anomaly classes, this does not resolve the underlying issue: the dataset itself becomes unreliable due to this inversion effect. This unreliability undermines confidence in the evaluation process, especially when encountering \textit{new or previously unseen anomaly types/classes}. In such cases, one cannot determine whether a low anomaly score corresponds to an actual anomaly or a misclassified normal sample, since \textit{the inversion depends on the anomaly type/class being considered} (as Fig.\ref{fig:34} shows), and the resolution of the inversion problem is only confirmed for the known types. Of course, this could also happen for a dataset which did not have the inversion problem in the first place for the known classes, since we cannot known what the behavior will be for new anomaly types. But we simply cannot do anything to remedy that, while we can do something to remedy the former inversion case described here. 

If a newly constructed dataset from a real world application yields inverted AUC values for standard baseline methods, this should be treated as a red flag. It suggests that the dataset may not be structurally sound for reliable anomaly detection, and should be reconsidered or refined to ensure that the scoring behavior is consistent and interpretable across all test sample types. One needs solid and stable baselines and datasets to properly perform fully unconditional-OOD AD.

The bottom line is that the main problem with the use datasets which are known to give inverted AUC metrics (i.e., where lower anomaly scores correspond to more anomalous samples) for basic baselines is the ambiguity it introduces in interpreting individual results in new methods. For a new test sample whose anomaly score is lower than those of most normal samples, it is unclear whether the sample is a true anomaly (under an inverted scoring convention), or simply a normal instance that lies in the left tail of the normal distribution.


\subsection{Model Details and Implementation}\label{Model Details and Implementation}
\subsubsection{Model}
\begin{table*}[h]
\centering
\caption{Architecture of the proposed Variational Autoencoder (VAE).}
\label{tab:vae_architecture}
\begin{tabular}{ll}
\hline
\textbf{Component} & \textbf{Layer Details} \\
\hline
\textbf{Encoder} & \\
Input & $3 \times 64 \times 64$ RGB image \\
Conv Block & Conv(3,16,5,pad=2) $\rightarrow$ BN $\rightarrow$ LeakyReLU $\rightarrow$ AvgPool(2) \\
Residual Block @32 & ResidualBlock(16$\rightarrow$32) \\
Downsample & AvgPool(2) \\
Residual Block @16 & ResidualBlock(32$\rightarrow$64) \\
Downsample & AvgPool(2) \\
Residual Block @8 & ResidualBlock(64$\rightarrow$64) \\
Flatten + FC & Linear(4096$\rightarrow$512) \\
\textbf{Latent Space} & $z \in \mathbb{R}^{256}$ (via reparameterization) \\
\textbf{Decoder} & \\
FC + Activation & Linear(256$\rightarrow$4096) $\rightarrow$ BN $\rightarrow$ LeakyReLU \\
Residual Block @4 & ResidualBlock(64$\rightarrow$64) \\
Upsample & $\times 2$ \\
Residual Block @8 & ResidualBlock(64$\rightarrow$32) \\
Upsample & $\times 2$ \\
Residual Block @16 & ResidualBlock(32$\rightarrow$16) \\
Upsample & $\times 2$ \\
Residual Block @32 & ResidualBlock(16$\rightarrow$16) \\
Output & Conv(16,3,5,pad=2) $\rightarrow$ Sigmoid \\
\hline
\textbf{Total Parameters} & 3,422,979 \\
\hline
\end{tabular}
\end{table*}


See Table.\ref{tab:vae_architecture}. Each residual block contains two Conv+BN+LeakyReLU layers, with optional expansion layers.

\subsubsection{Choosing the gain for each loss term}

The constants $\alpha_{i,j},\,\beta_{i,j}$ multiplying the elements of the hyperspherical loss are proportional to $1/\sqrt{k+1}$, where \(k\) is the coordinate index. This was necessary because, unlike the Cartesian coordinates, the hyperspherical coordinates are asymmetric and vary with $k$. This can be seen in the transformation formulas, where a product of an increasing amount of sine functions is necessary as the coordinate index increases. We chose $1/\sqrt{k+1}$, guided by the fact that the vector whose Cartesian coordinates are $(1,1,...,1)$ has a cosine of its spherical angles equal to $1/\sqrt{k+1}$ as the coordinate index k varies, and because it gave the best results experimentally.

In this way, we were able to avoid lengthy calculations to obtain the mathematically exact formulas for both these constants and the KLD in hyperspherical coordinates, which we do not believe to be important for the goals of this work.

Finally, there are single scalar hyperparameters/gains multiplying each loss part after summing up the corresponding index $k$ in each of them, and a global single scalar hyperparameter/gain $\beta$ multiplying the total KLD-like loss in hyperspherical coordinates. The optimal value for these hyperparameters was found via a simple grid search and are provided in the full code release.

\subsubsection{Differences in training speed}\label{appendix:trainspeed}

We provide here data regarding the differences in the training speeds between the standard VAE and our compression VAE via hyperspherical coordinates. The origin of this difference mainly lies in the extra calculations needed for the coordinate transformations in \ref{appendix:hstransform}, which are implemented via the script in \ref{appendix:hstransformcode}.

The measurements were done during typical trainings in a NVIDIA H100 GPU. In Fig.\ref{fig:36} we show the results for the case of trainings with CIFAR10, with a batch size of $200$ samples, and the changes in training speed (measured as how many batches per second are being processed) in terms of the dimension $n$ of the latent space. After $n=200$, until $n=800$, the decay is almost linear in $n$, with a decay rate in the speed of $20$ batch/s every $200$ latent dimensions.

\onecolumn

\begin{figure*}[!h]
    \centering
    \includegraphics[width=0.7\linewidth]{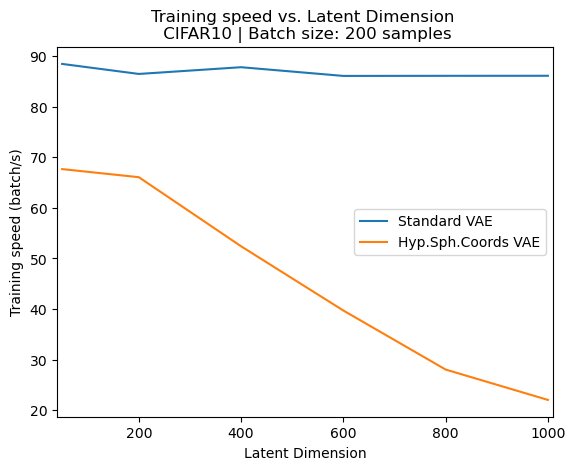}
    \caption{Differences in training speed.}
    \label{fig:36}
\end{figure*}

\twocolumn


\bibliography{references_AD}
\bibliographystyle{IEEEtran}


\newpage

\vfill

\end{document}